\pdfoutput=1
\documentclass[10pt,journal,compsoc]{IEEEtran}

\ifCLASSOPTIONcompsoc
  \usepackage[nocompress]{cite}
\else
  \usepackage{cite}
\fi

\usepackage{graphicx}
\ifCLASSINFOpdf
\else
\fi

\usepackage{amsmath,amssymb,amsfonts}

\usepackage{booktabs}
\usepackage[loop,autoplay]{animate}

\usepackage{siunitx}
\DeclareSIUnit{\nothing}{\relax}
\usepackage{algpseudocode,algorithm}
\usepackage{xcolor}
\usepackage{array}

\usepackage{subcaption}
\usepackage{makecell}
\usepackage{adjustbox} %
\usepackage{enumitem} %
\usepackage{multirow}
\usepackage{mathrsfs}
\usepackage{empheq}

\usepackage{etoolbox} %
\newif\ifhideimages
\hideimagesfalse

\newif\ifshowanimationcommentarxiv
\showanimationcommentarxivtrue

\usepackage{url}
\def\MYTITLE{Formulating Event-based Image Reconstruction as a Linear Inverse Problem with Deep Regularization using Optical Flow}

\newcommand\MYhyperrefoptions{bookmarks=true,bookmarksnumbered=true,
pdfpagemode={UseOutlines},plainpages=false,pdfpagelabels=true,
hidelinks=true, %
breaklinks=true,
pdftitle={\MYTITLE},%
pdfsubject={Computer Vision, Inverse Problems, Robotics},%
pdfauthor={Z. Zhang, A. Yezzi, G. Gallego},%
pdfkeywords={Event camera, Asynchronous sensor, Inverse Problems, Image Reconstruction}}%
\usepackage[\MYhyperrefoptions,pdftex]{hyperref}

\usepackage[capitalize]{cleveref}
\crefname{section}{Sec.}{Secs.}
\Crefname{section}{Section}{Sections}
\Crefname{table}{Table}{Tables}
\crefname{table}{Tab.}{Tabs.}

\pdfpxdimen=\dimexpr 1 in/72\relax
\usepackage{calc}

\def\Lum{L}
\def\tref{t_\text{ref}} %
\def\pol{p} %

\def\cE{\mathcal{E}} %
\def\numEvents{N_e} %

\def\bu{\mathbf{u}}
\def\bx{\mathbf{x}}

\def\pol{p}

\def\bell{\boldsymbol{\ell}} %
\def\btheta{\boldsymbol{\theta}} %

\def\vecniwe{\mathbf{b}}
\def\bn{\mathbf{n}}
\def\cN{\mathcal{N}} %
\def\cR{\mathcal{R}} %
\def\bz{\mathbf{z}}
\def\cF{\mathcal{F}} %
\def\veclap{\mathbf{c}}
\def\bzero{\mathbf{0}}
\def\mId{\mathrm{Id}}

\def\Real{\mathbb{R}} %

\robustify\bfseries
\newcommand{\bnum}[1]{\bfseries #1}
\newcommand{\unum}[1]{\underline{\num{#1}}}

\ifshowanimationcommentarxiv
\makeatletter
\long\def\@IEEEtitleabstractindextextbox#1{\parbox{0.922\textwidth}{#1}}
\makeatother

\usepackage{orcidlink}
\usepackage[absolute]{textpos}
\fi

\begin{document}
\title{\MYTITLE}

\ifshowanimationcommentarxiv
\definecolor{somegray}{gray}{0.6}
\newcommand{\darkgrayed}[1]{\textcolor{somegray}{#1}}
\begin{textblock}{11}(2.5, 0.3)
\begin{center}
\darkgrayed{This paper has been accepted for publication at the\\
IEEE Transactions on Pattern Analysis and Machine Intelligence, 2022.
\copyright IEEE}
\end{center}
\end{textblock}
\fi

\author{
Zelin Zhang$^1$, Anthony Yezzi$^{2}$, Guillermo Gallego$^{1,3}$\orcidlink{0000-0002-2672-9241}%
\IEEEcompsocitemizethanks{\IEEEcompsocthanksitem 
$^1$ Z. Zhang and G. Gallego are with the Technische Universit\"at Berlin, Berlin, Germany.
$^3$ G. Gallego is with the Einstein Center Digital Future and the Science of Intelligence Excellence Cluster, Berlin, Germany. guillermo.gallego@tu-berlin.de
\IEEEcompsocthanksitem 
$^2$ Anthony Yezzi is with the Department of Electrical and Computer Engineering, Georgia Institute of Technology, Atlanta, USA.
\IEEEcompsocthanksitem Preprint of paper accepted at IEEE T-PAMI, 2022.
}%
}

\IEEEtitleabstractindextext{%

\begin{abstract}
Event cameras are novel bio-inspired sensors that measure per-pixel brightness differences asynchronously. Recovering brightness from events is appealing since the reconstructed images inherit the high dynamic range (HDR) and high-speed properties of events; hence they can be used in many robotic vision applications and to generate slow-motion HDR videos. However, state-of-the-art methods tackle this problem by training an event-to-image Recurrent Neural Network (RNN), which lacks explainability and is difficult to tune. 
In this work we show, for the first time, how tackling the combined problem of motion and brightness estimation leads us to formulate event-based image reconstruction as a \emph{linear inverse problem} that can be solved without training an image reconstruction RNN. Instead, classical and learning-based regularizers are used to solve the problem and remove artifacts from the reconstructed images. The experiments show that the proposed approach generates images with visual quality on par with state-of-the-art methods despite only using data from a short time interval. 
State-of-the-art results are achieved using an image denoising Convolutional Neural Network (CNN) as the regularization function.
The proposed regularized formulation and solvers have a unifying character because they can be applied also to reconstruct brightness from the second derivative. Additionally, the formulation is attractive because it can be naturally combined with super-resolution, motion-segmentation and color demosaicing.
Code is available at \url{https://github.com/tub-rip/event_based_image_rec_inverse_problem}
\end{abstract}

}

\maketitle

\IEEEdisplaynontitleabstractindextext

\IEEEpeerreviewmaketitle

\section{Introduction}
\label{sec:intro}

Event cameras are novel bio-inspired sensors that offer advantages over traditional frame-based cameras (high speed, high dynamic range (HDR), low power, etc.) \cite{Lichtsteiner06isscc,Gallego20pami}. 
However they acquire visual data in the form of asynchronous per-pixel brightness \emph{changes}, called ``events'', instead of standard brightness images.
We tackle the problem of image reconstruction from event-based data, i.e., recovering the brightness signal that caused the events. 
It is an interesting problem in itself and finds multiple applications:
($i$) it enables the creation of high-speed and/or HDR videos,
($ii$) if the reconstructed images are of sufficient quality they can be used as input to a large body of algorithms developed for traditional cameras. Hence, image reconstruction makes event data compatible with mainstream computer vision.
Reconstructed images have applications from autonomous driving to smartphone applications for everyday use~\cite{Chen20msp}.

\begin{figure}[t]
  \centering
  \includegraphics[trim=2.1cm 4.1cm 5.5cm 1.5cm,clip,width=\linewidth]{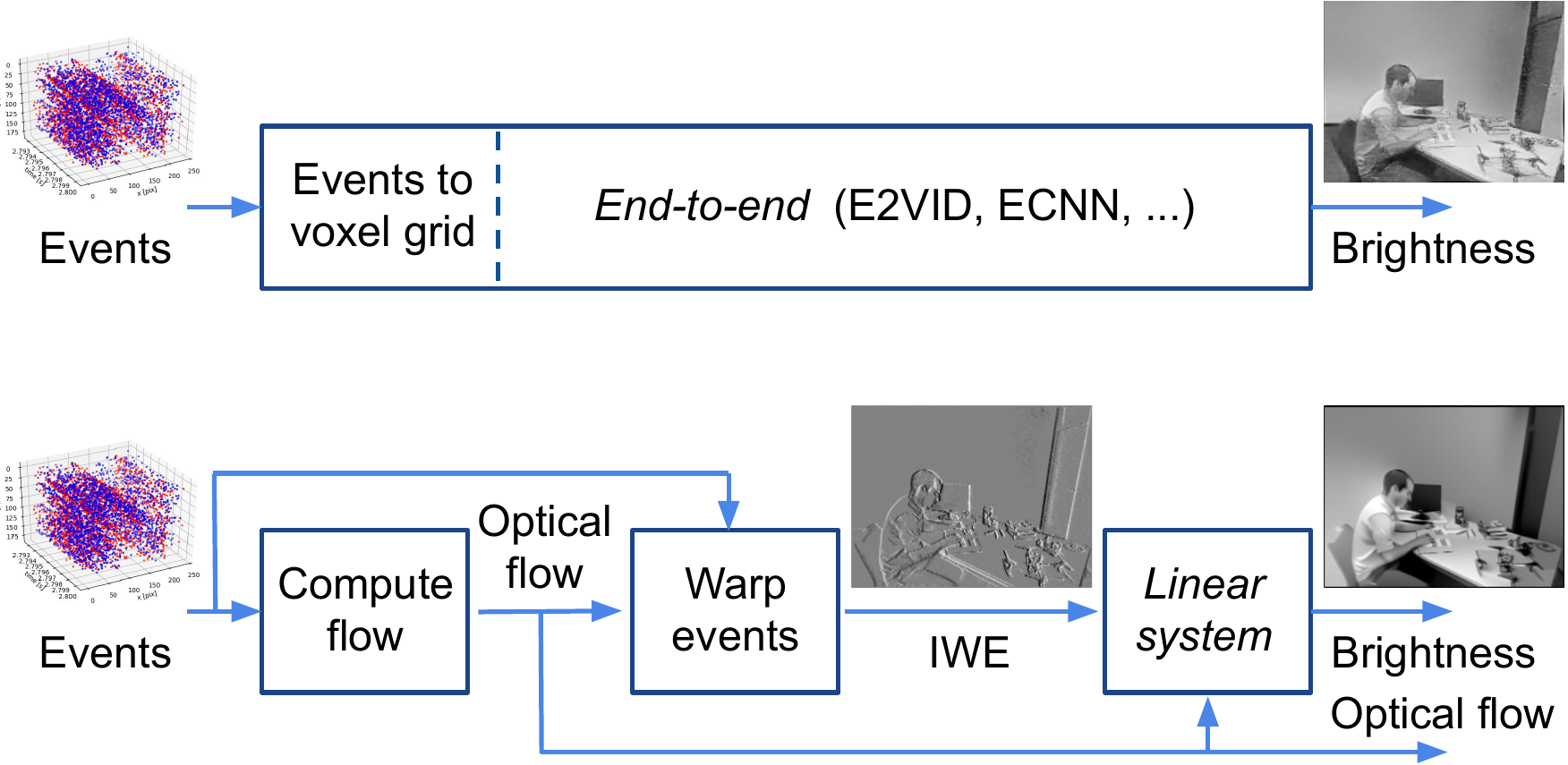}
  \caption{Overview of the proposed method (bottom) in comparison with state-of-the-art event-to-image recurrent neural networks (RNNs) (top).
  End-to-end image reconstruction methods can be replaced by an explainable system that recovers both optical flow and image brightness. 
  We show that, given optical flow, image reconstruction is a linear problem in the unknown brightness,
  and exploit this knowledge to estimate brightness by means of several classical and recent learning-based solvers.
  \label{fig:diagram}
  }
\end{figure}

Image reconstruction is challenging because events are an unfamiliar representation of visual data, they depend on motion and exhibit a considerable amount of noise and non-ideal effects (caused by pixel fabrication mismatch, the amount of incident light, sub-threshold transistor operation, etc.).
Recently, notable progress has been made. 
Early methods performed image reconstruction as a means to aid other tasks, such as ego-motion estimation~\cite{Kim14bmvc}.
They were model-based, proposing a leaky integrator or filter that would perform some form of spatio-temporal smoothing and denoising~\cite{Munda18ijcv,Scheerlinck18accv}. 
Nowadays, due to the outstanding capacity of artificial neural networks (ANNs) for pixel regression, state-of-the-art methods are deep-learning based \cite{Rebecq19pami,Mostafavi21ijcv}. 
They leverage large amounts of data in a supervised or unsupervised manner (e.g., via generative adversarial networks (GANs)) to recover surprisingly high-quality image brightness from a voxelized representation of the event data, 
at a significantly higher computational cost compared to model-based methods (\cref{fig:diagram}, top).

However, these approaches suffer from several problems: 
(\emph{i}) the resulting event-to-image ANNs are black boxes, relying on the networks to learn how to recover images from events; 
(\emph{ii}) the ANNs suffer from artifacts and there is no easy way to tune them except for training with more data variation;
(\emph{iii}) the approaches focus only on recovering image brightness, but motion is as equally important to recover and both variables are firmly interconnected %
in the event data;
(\emph{iv}) events are sparse but they are converted into a voxelized representation for compatibility with ANNs, which causes a filling-in effect that requires considerable memory while most of the voxels may be empty;
(\emph{v}) some methods use a first-order event generation model, which inevitably introduces linearization errors.

Our work provides a new point of view on image reconstruction from event data.
If one considers brightness and motion (i.e., optical flow) as equally important entangled variables in the event stream, 
the goal should be to estimate both (not just one of them) and to leverage existing \emph{asymmetries} in the process. 
Specifically, show that, given optical flow, brightness estimation is a \emph{linear problem}. 
Therefore, (\emph{i}) the difficult subproblem of the two is that of estimating accurate optical flow\footnote{This admits analogies in other domains of computer vision, such as camera self-calibration~\cite{Hartley03book}: 
given the plane at infinity, recovering the intrinsic parameters of the camera is also a linear problem. 
Hence, finding the plane at infinity is the difficult (sub)task.
}, 
(\emph{ii}) easier and more interpretable (explainable) methods than current ones shall be used for image reconstruction (\cref{fig:diagram}, bottom).
In particular, we leverage the fact that linear problems are extensively studied in science,
and the corresponding solvers keep improving, benefiting from a growing body of work that exploits large image datasets.
This naturally leads to lower dimensional, efficient intermediate representations. 
Additionally, the proposed image reconstruction method does not need ground truth $\langle$event, image$\rangle$ data pairs, which are required in state-of-the-art supervised learning approaches.

We evaluate the method on standard datasets and the experiments demonstrate that image reconstruction with the proposed technique (linear solver plus deep image priors) produces compelling results, on par with the state of the art in terms of quality, 
but without using any ground truth image and with notably better interpretability and flexibility.

In summary, our main contributions are:
\begin{itemize}%
\item A novel formulation of the problem of image reconstruction from events, aided by optical flow,
in the form of a linear system of equations directly in the unknown brightness image (\cref{sec:method:general}).
\item We propose a variety of solvers and regularizers, including recent deep denoisers, to solve the above problem, 
using efficient input event representations and without using ground truth labels (\cref{sec:method:priors}).
\item Explainability: Our method combines physics (the event generation model used to derive the system of equations) 
and machine learning (for artifact removal through image denoising) (Secs.~\ref{sec:method:general} to \ref{sec:method:priors}).
\item Controllability: We can control the amount of regularization online, without retraining; just by changing a hyperparameter.
\item A natural extension of our method to simultaneously perform image reconstruction and super-resolution (\cref{sec:method:superresolution}).
\item We show that the same technique (linear constraints, inverse problem and CNN image priors) can be used to solve the image reconstruction problem in two different variants
(\cref{sec:method:general,sec:method:LaplacianRec}).
CNN image priors are combined with Poisson solvers to provide a non-linear (akin M-estimator) norm to guide the solution.

\item We also show how our method can be combined with motion segmentation (e.g., to recover images of independently moving objects in the scene) and with color event cameras (to reconstruct color images), (\cref{sec:method:extensions}).

\item A thorough evaluation on standard datasets and comparing with three state-of-the-art event-to-image RNN methods (\cref{sec:experim}).
\end{itemize}
\begin{figure}[t]
  \centering
  \includegraphics[trim=6.5cm 3.1cm 5.5cm 2.5cm,clip,width=.9\linewidth]{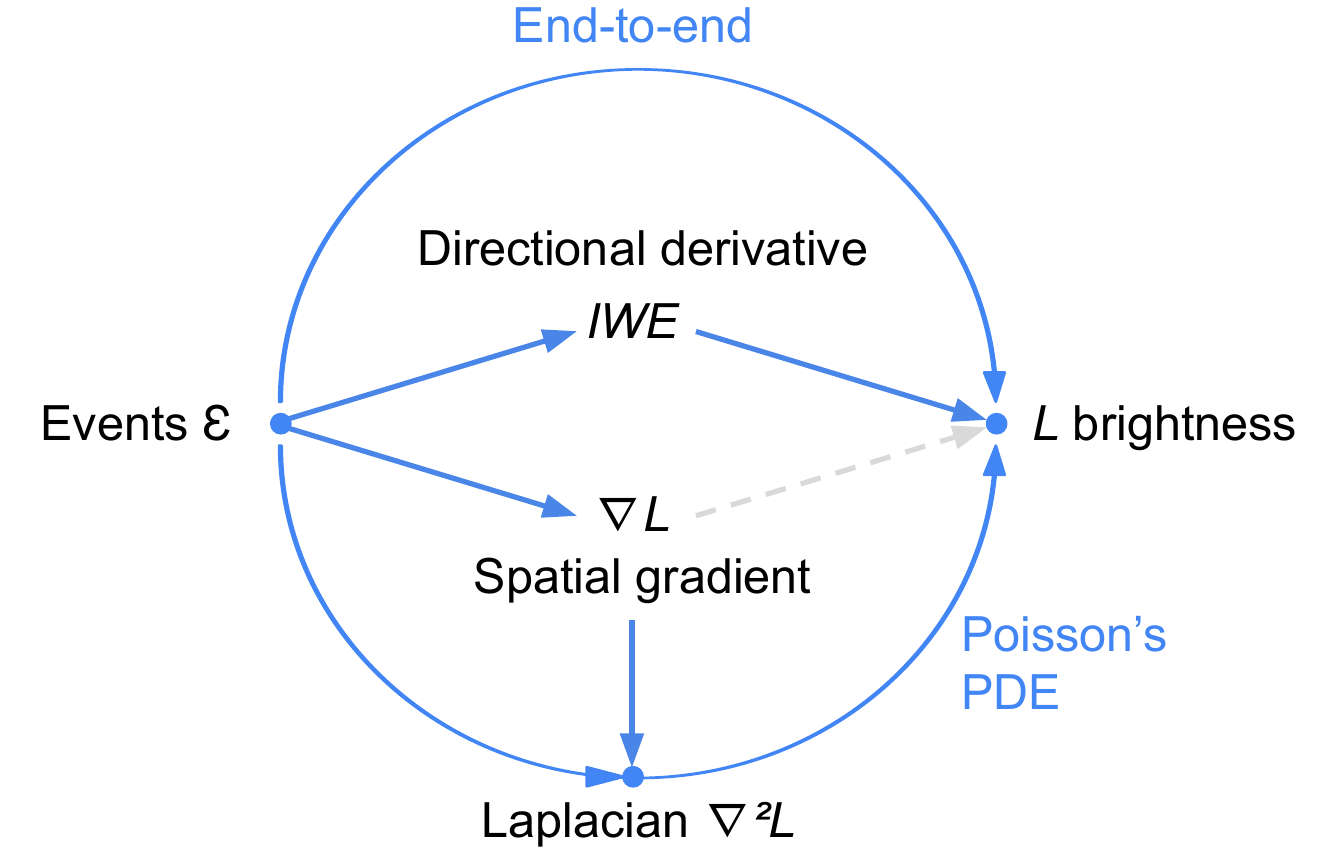}
  \caption{From events to brightness. 
  Paths followed by image reconstruction methods in terms of the visual quantities estimated.
  The two predominant categories of methods are either end-to-end or estimate the Laplacian $\nabla^2 L$ and subsequently solve Poisson's equation.
  We show ($i$) \emph{a new path} based on the image of warped events (IWE), 
  and ($ii$) how to improve Poisson's path by incorporating image priors.
  Paths are blue; visual quantities are black.
  \label{fig:diagrammethods}
  }
\end{figure}

The proposed approach is appealing not only because of the novel point of view (combined estimation of entangled variables while exploiting asymmetries) but also because: 
($i$) it opens a new route to reconstruct brightness based on the first directional derivative (\cref{fig:diagrammethods}), 
($ii$) its techniques can also be applied to the approach of reconstruction based on second derivatives of the brightness,
and ($iii$) it can be combined with a variety of topics (super-resolution, motion segmentation, demosaicing) in a natural way due to its linear inverse problem formulation.

\section{Related Work}
\label{sec:relatedwork}

\textbf{Image brightness reconstruction from events}.
Event cameras such as the DVS~\cite{Lichtsteiner08ssc,Son17isscc,Suh20iscas,Finateu20isscc} are bio-inspired sensors that capture pixelwise \emph{brightness changes}, called events, instead of brightness images. 
An event $e_k \doteq (\bx_k, t_k, \pol_{k})$ is triggered as soon as the logarithmic brightness $\Lum$ at a pixel exceeds a preset contrast sensitivity $C>0$, 
\begin{equation}
\label{eq:generativeEventCondition}
\Lum(\bx_k,t_k) - \Lum(\bx_k, t_k-\Delta t_k) = \pol_k \, C,
\end{equation} 
where $\bx_k\doteq (x_k, y_k)^{\top}$, $t_k$ (with \si{\micro\second} resolution) and polarity $\pol_{k} \in \{+1,-1\}$
are the spatio-temporal coordinates and sign of the brightness change, respectively,
and $\Delta t_k$ is the time elapsed since the last event at the same pixel $\bx_k$.

\def\imgHeight{2.6cm}

\def\imgWidth{0.2\linewidth}
\begin{figure*}
\centering
\begin{subfigure}{\imgWidth}
    \includegraphics[trim=2cm 1cm 2cm 2cm,clip,height=\imgHeight]{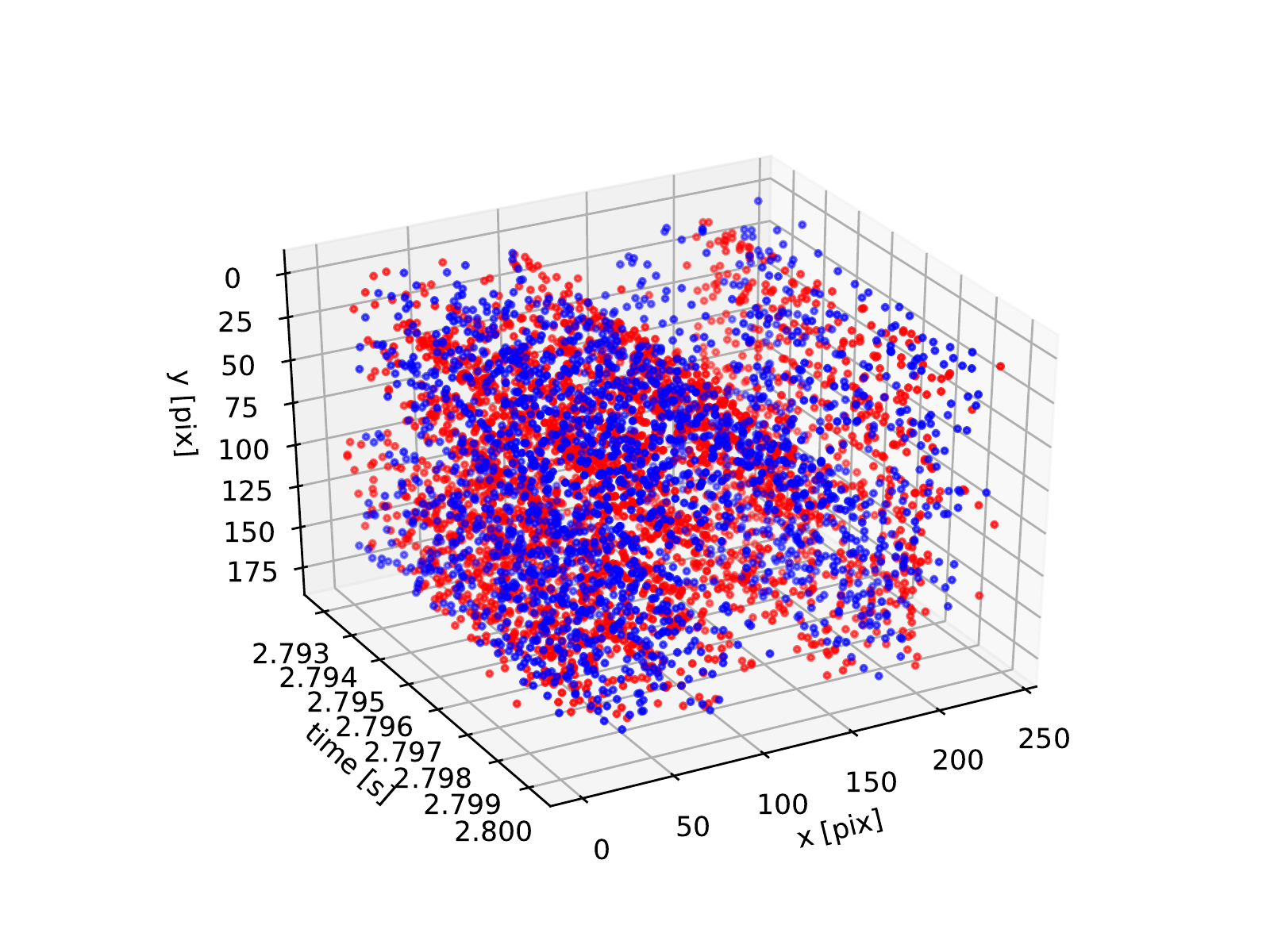}
    \caption{Input events $\cE$\label{fig:steps:events}}
\end{subfigure}%
\begin{subfigure}{\imgWidth}
    \frame{\includegraphics[height=\imgHeight]{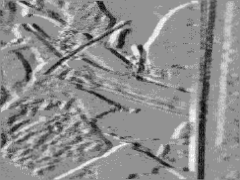}}
    \caption{Pixel-sum of polarities\label{fig:steps:incremimage}}
\end{subfigure}%
\begin{subfigure}{\imgWidth}
    \frame{\includegraphics[height=\imgHeight]{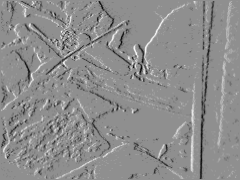}}
    \caption{IWE (sharp)\label{fig:steps:iwe}}
\end{subfigure}%
\begin{subfigure}{\imgWidth}
    \includegraphics[height=\imgHeight]{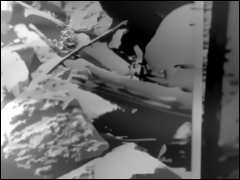}
    \caption{Reconstructed image \label{fig:steps:e2vid}}
\end{subfigure}%
\begin{subfigure}{\imgWidth}
    \includegraphics[height=\imgHeight]{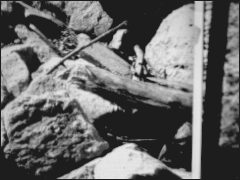}
    \caption{Ground truth image \label{fig:steps:frame}}
\end{subfigure}
\caption{The events $\cE$ in (a) %
are motion-compensated using optical flow to create the image of warped events (IWE) (c). 
The sharp IWE approximates the $x$-derivative of the ground truth frame better than the uncompensated image (b).
Our method reconstructs brightness image %
(d) from the IWE (c) by solving a linear system of equations with regularization (CNN-based in this example). 
Texture details smaller than the contrast sensitivity $C$ in~\eqref{eq:generativeEventCondition} cannot be recovered since they do not trigger events. %
The event data in the figure is from the \emph{slider\_far} sequence in~\cite{Mueggler17ijrr}, consisting of rocks and tree-like textures.
}
\label{fig:steps}
\end{figure*}

Image brightness $\Lum$ can be estimated from a stream of asynchronous events in several ways, as depicted in \cref{fig:diagrammethods}.
Methods can be categorized into model-based~\cite{Cook11ijcnn,Kim14bmvc,Kim16eccv,Rebecq17ral,Bardow16cvpr,Munda18ijcv,Scheerlinck18accv} 
or learning-based~\cite{Rebecq19pami,Scheerlinck20wacv,Stoffregen20eccv,Mostafavi21ijcv,Bardow18thesis}; 
sometimes they are both~\cite{Barua16wacv,Duwek21cvprw}, with submodules in each category.
Some methods estimate the spatial brightness gradient ($\nabla \Lum$) first,
as an intermediate quantity \cite{Cook11ijcnn,Kim14bmvc,Barua16wacv,Kim16eccv,Rebecq17ral};
then brightness is recovered by computing the Laplacian image and solving Poisson's PDE~\cite{Kim14bmvc,Barua16wacv,Rebecq17ral} or its more robust version~\cite{Kim16eccv}.
The first step may be bypassed, estimating the Laplacian image directly from events~\cite{Duwek21cvprw}.

\textbf{Early works}. 
Cook et al. \cite{Cook11ijcnn} defined a network where messages with local update rules were passed to estimate several visual quantities simultaneously. 
Image brightness was computed based on its relation to the spatial gradient, which is connected to other network quantities. 
Kim et al. \cite{Kim14bmvc,Kim16eccv} used parallel Bayesian filters to simultaneously estimate camera motion and image gradients. 
Image brightness was recovered via Poisson integration and passed to the camera tracker. 
In \cite{Bardow16cvpr} image brightness was obtained as the supporting variable of a costly, space-time variational functional optimization problem whose primary goal was optical flow estimation. 
The function comprised data-fidelity and regularization terms for the flow and the brightness, and the solver proceeded in an alternating fashion via a parallelized primal-dual algorithm \cite{Pock11iccv}.
Barua et al. \cite{Barua16wacv} used dictionary learning on space-time patches, learning atoms via K-SVD \cite{Aharon06tsp}.
Events were used to estimate the spatial brightness gradient via patch-wise sparse coding, and subsequently the Laplacian was computed and Poisson-integrated.

Many of the above methods require knowledge of the camera motion (pure rotation \cite{Cook11ijcnn,Kim14bmvc}), and/or depth~\cite{Kim16eccv,Rebecq17ral}. 
Only recently this requirement has been relaxed~\cite{Bardow16cvpr,Munda18ijcv,Scheerlinck18accv,Barua16wacv}, 
achieving  brightness reconstruction via temporal filtering~\cite{Scheerlinck18accv} 
or temporal integration with manifold denoising~\cite{Munda18ijcv}.
These model-based approaches suffer from artifacts such as ghosting effects and bleeding edges.
Compared to \cite{Bardow16cvpr}, our method assumes optical flow is obtained first, for example via contrast maximization \cite{Gallego18cvpr,Shiba22eccv} or E-RAFT \cite{Gehrig21threedv}, and then it recovers brightness. 
The goal is to highlight and exploit the asymmetries in the combined problem of recovering flow and brightness, which are entangled variables in the event stream.

\textbf{Deep learning event-to-image methods}.
State-of-the-art methods outperform the above prior methods and operate by letting ANNs learn the mapping from events to brightness images. 
Events are typically converted into a dense voxelgrid representation and fed to modern ANNs.
High quality results are obtained using supervised learning on recurrent neural networks (RNNs) trained on large amounts of synthetic data \cite{Rebecq19pami,Scheerlinck20wacv,Stoffregen20eccv}.
A jump in image reconstruction quality was demonstrated by E2VID \cite{Rebecq19pami} and its improved trained version \cite{Stoffregen20eccv}.
Unsupervised generative adversarial network (GAN) \cite{Bardow18thesis,Mostafavi21ijcv,WangLin20cvpr} and domain adaptation \cite{Zhang20eccv} approaches have been also investigated, but results are not as good.
Recently, a self-supervised learning approach based on a linearized version of the event generation model~\eqref{eq:generativeEventCondition} has been proposed in \cite{Paredes21cvpr} to try to leverage large amounts of unlabeled event data for image reconstruction.
It consists of two ANNs, one for optical flow estimation\cite{Zhu19cvpr} and another one for image reconstruction, 
and the latter is trained using brightness constancy as a loss.

The above event-to-image deep learning methods are to a large extent black boxes, with a lot of effort spent not only on loss design and architecture search but also on dataset preparation \cite{Stoffregen20eccv,WangLin20cvpr,Zou21cvpr} to train the ANNs and make them learn the desired transformation. 
Different from these methods, our approach does not train any event-to-image ANN.
We pose the problem as image reconstruction aided by optical flow \cite{Paredes21cvpr} 
and show that \eqref{eq:generativeEventCondition} leads to a \emph{linear} system of equations. 
Several explainable methods are used to solve such a system.
To achieve best results, we reutilize state-of-the-art image denoising ANNs, without retraining.

\section{Method}
\label{sec:method}

\subsection{Intuitive 1D Example}
\label{sec:method:onedim}

To convey the main idea of our method let us consider the simple case of 1D motion (horizontally, along the $x$ axis).
For further simplicity, consider that the scene is planar and front-to-parallel with respect to the camera. 
Hence, all pixels on the image plane move with the same velocity $\bu$.

\Cref{fig:steps} summarizes the main idea and steps of the method.
Consider the set of events $\cE \doteq \{e_k\}_{k=1}^{\numEvents}$ in a volume of the image plane (\cref{fig:steps:events}).
Since events are temporal brightness changes, pixel-wise accumulation of the event polarities gives the brightness increment image in \cref{fig:steps:incremimage}.
If the accumulation interval $\Delta t = [t_1, t_{\numEvents}]$ is large, then this image is blurred.
Instead, assuming the optical flow is known we may motion-compensate the events, 
$\cE \mapsto \cE'$,
by transporting them to new pixel locations 
\begin{equation}
\label{eq:warpeventsbyflow}
\bx'_k = \bx_k - (t_k - \tref)\,\bu
\end{equation}
(horizontally, in this case), 
where they will be aligned and produce a sharp Image of Warped Events (IWE) (\cref{fig:steps:iwe}).
This sharp image represents the strength of the moving edges in the scene~\cite{Gallego18cvpr} and is entangled with the motion $\bu$.
The IWE resembles the $x$-derivative of the ground truth brightness frame.
Such a spatial derivative is obtained from the brightness image using a finite difference operator $D_x$.
In the simple case considered, we have:
\begin{equation}
\label{eq:onedim:iwe-and-Dx}
D_x \Lum  \approx  \text{IWE}(x,y)
\end{equation}
where $D_x \Lum \doteq \Lum(x+1,y) - \Lum(x,y)$ is a 2-point finite difference approximation to the $x$-derivative over the pixel grid.
More elaborate operators, like Sobel's, are applicable.
The key ideas to notice are that each pixel of the IWE provides one equation \eqref{eq:onedim:iwe-and-Dx} to recover the unknown brightness $\Lum$, and that equations are \emph{linear} in $\Lum$.
Stacking \eqref{eq:onedim:iwe-and-Dx} for all pixels, produces a \emph{linear system of equations} 
$D_x\bell = \text{vec}(\text{IWE})$ in the unknown $\bell \doteq \text{vec}(\Lum)$.
Hence, we reformulate the problem of image reconstruction %
as that of solving a linear system of equations.
In the example, it is the problem of finding $\Lum$ whose $x$-derivative is the IWE.
Thus we may study how well-posed the problem is, and test the large collection of linear solvers (GMRES \cite{GMRES86}, BiCGStab \cite{BiCGSTAB92}, etc.) to estimate $\Lum$.
If there are measurement errors in the equations, 
we may reformulate the problem as a least-squares one:
\begin{equation}
\label{eq:onedim:leastsquares}
    \min_{\bell} \|D_x\bell - \text{vec}(\text{IWE})\|_2.
\end{equation}
\Cref{fig:steps:e2vid} shows an example of a reconstructed brightness image $\Lum$ by solving the above system of equations 
with an additional regularization term.
This is discussed next.

\subsection{Generalization to arbitrary Motions and Scenes}
\label{sec:method:general}

In the general case of arbitrary scene and motion 
we still follow the reasoning above, but with minor modifications.
Events $\cE\mapsto\cE'$ are warped~\eqref{eq:warpeventsbyflow} according to an image velocity field that may change at each pixel location, $\bu(\bx)$.
The resulting IWE is still sharp~\cite{Gallego18cvpr}, 
\begin{equation}
\label{eq:IWEfromevents}
\text{IWE}(\bx) \equiv \Delta \Lum(\bx) \doteq \sum_{k=1}^{\numEvents} C\, \pol_{k}\, \delta (\bx - \bx'_k),
\end{equation}
but to account for different optical flow magnitudes 
$\left \| \bu(\bx) \right \| > 0$
(e.g., objects at different depths trigger a different number of events), 
we divide by the magnitude of the motion.
Warped events $\bx'_k$ have floating point precision. 
The delta $\delta$ in \eqref{eq:IWEfromevents} is an idealized model amenable for integration. 
In practice it is replaced by a kernel, such as a Gaussian $\delta (\bx) \approx \cN(\bx; \bzero, \epsilon^2 \mId)$, where $\epsilon\approx 1$ pixel and $\mId$ is the identity matrix \cite{Gallego18cvpr,Ng22ral}.
For speed-up, it is implemented using bilinear voting and Gaussian convolution \cite{Gallego17ral}. 
This produces the normalized IWE (NIWE):
\begin{equation}
\label{eq:NIWE}
\Delta \Lum'(\bx) \doteq \Delta L(\bx) \,/\, \| \bu(\bx)\| \Delta t.
\end{equation}
If $\left \| \bu(\bx) \right \| = 0$, no events are triggered at pixel location $\bx$ and we set $\Delta \Lum'(\bx)  = 0$.
Assuming brightness constancy, the NIWE corresponds to the edge strength of the scene, i.e., instantaneous rate of brightness change in the optical flow direction at pixel $\bx$:
\begin{equation}
\label{eq:niwedotproduct}
\Delta \Lum'(\bx) \approx -\nabla \Lum \cdot \hat{\bu}(\bx),
\end{equation}
where $\nabla L=(\partial_x \Lum, \partial_y \Lum)^\top$ is the spatial gradient 
and $\hat{\bu} = \bu/\| \bu\|$ is the unit vector in the direction of $\bu$. 
The 1D example in \cref{sec:method:onedim} corresponds to the case $\hat{\bu}(\bx) = -(1,0)^\top$, 
so that $-\nabla \Lum \cdot \hat{\bu}(\bx)=\partial_x \Lum$,
and just like $\partial_x \Lum$ was discretized using a finite difference formula with a 1~pixel step over the pixel grid, 
we may also discretize $\nabla \Lum \cdot \hat{\bu}(\bx)$, leading to:
\begin{equation}
\label{eq:linearitypixel}
\Delta \Lum'(\bx) \approx \Lum(\bx) - \Lum(\mathbf{x + \hat{\bu}(\bx)}),
\end{equation}
where $\Lum(\mathbf{x + \hat{\bu}(\bx)})$ can be computed by interpolation (e.g., bilinear interpolation). %
Stacking all equations \eqref{eq:linearitypixel}, 
with the NIWE $\vecniwe \doteq \Delta \Lum'(\bx)$ computed from the input data~\eqref{eq:NIWE},
produces the anticipated \emph{linear system of equations} 
\begin{equation}
\label{eq:general:linearsystem}
D \bell = \vecniwe,
\end{equation}
where $D \bell$ is a compact notation for the (directional derivative approximation given by the) right hand side of~\eqref{eq:linearitypixel}.

\emph{Notation}: we use $D$ to represent the operator acting on $\Lum$ or on its stacked version $\bell$.
If $\Lum \in \Real^{m\times n}$, $\bell\in \Real^{mn}$, and $D$ in $D\bell$ is a sparse matrix of size ${mn \times mn}$, 
which can be constructed given the pixel-wise optical flow $\hat{\bu}(\bx)$.

\subsection{The Operator in the Linear System}
\label{sec:method:operator}

One main challenge in solving the linear system~\eqref{eq:general:linearsystem} is that the operator $D$ is not full rank. 
Hence, besides choosing a numerically good discretization for $D$, 
additional equations are needed to ensure that the problem is well posed.

Intuitively, in the 1D example of \cref{sec:method:onedim} we look for the image $\Lum$ whose $x$-derivative agrees with the data (the NIWE),
and we know that inverting a derivative may be ill-posed.
If $D$ is discretized using the 2-point finite difference formula, $D\Lum(x,y) \approx \Lum(x+1,y) - \Lum(x,y)$,
each pixel of $\Lum$ is only strongly connected to one of its side neighbors,
hence information in the linear solvers %
only propagates along the rows of $\Lum$.
This causes streamline-like \emph{artifacts} in the reconstructed image (\cref{fig:compare:reg:no}).

A simple way to mitigate these artifacts is to choose a better numerical discretization of $D$ 
that encourages cross-talking (i.e., coupling) among more pixel neighbors. 
\Cref{fig:compare:reg:gaussian} shows the effect of replacing the 2-point finite difference formula with Sobel's 9-point formula, 
which couples each pixel to its 8 neighbors (matrix $D$ is sparse and has at most 9 non-zero diagonals), 
thus encouraging smoothness of the solution in the direction perpendicular to the flow.

More advanced regularization techniques exist in the literature, such as Tikhonov regularization, total variation (TV), Beltrami, etc., 
as we discuss in the next section.

\emph{Remark}: At border pixels, some optical flow vectors may point out of the image boundary. 
To tackle this issue, we set the optical flow and NIWE at border pixels to zero so that both sides of \eqref{eq:linearitypixel} automatically agree. 
This operation, however, does not improve the rank-deficiency of matrix $D$. %

\def\figWidth{0.32\linewidth}
\begin{figure}[t]
\centering
\begin{subfigure}{\figWidth}
    \includegraphics[width=\linewidth]{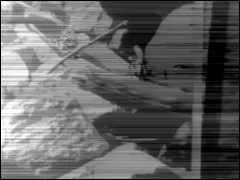}
    \caption{\label{fig:compare:reg:no}Without reg.}
\end{subfigure} 
\begin{subfigure}{\figWidth}
    \includegraphics[width=\linewidth]{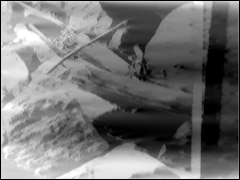}
    \caption{\label{fig:compare:reg:gaussian}Gaussian}
\end{subfigure} 
\begin{subfigure}{\figWidth}
    \includegraphics[width=\linewidth]{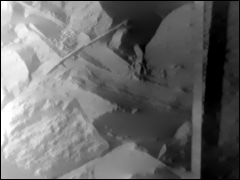}
    \caption{\label{fig:compare:reg:ltwo}Tikhonov}
\end{subfigure}\\[0.8ex]
\begin{subfigure}{\figWidth}
    \includegraphics[width=\linewidth]{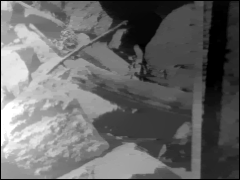}
    \caption{\label{fig:compare:reg:lone}Total variation}
\end{subfigure} 
\begin{subfigure}{\figWidth}
    \includegraphics[width=\linewidth]{images/example2/denoiser.png}
    \caption{\label{fig:compare:reg:cnn}CNN prior}
\end{subfigure} 
\begin{subfigure}{\figWidth}
    \includegraphics[width=\linewidth]{images/example2/gt_normalized.png}
    \caption{\label{fig:compare:reg:gt}Ground truth}
\end{subfigure}\\[0.8ex]
\begin{subfigure}{\figWidth}
    \frame{\includegraphics[width=\linewidth]{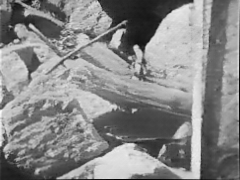}}
    \caption{\label{fig:compare:reg:e2vid}E2VID~\cite{Rebecq19pami}}
\end{subfigure} 
\begin{subfigure}{\figWidth}
    \frame{\includegraphics[width=\linewidth]{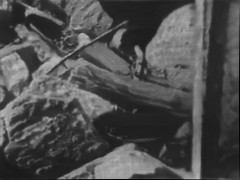}}
    \caption{\label{fig:compare:reg:ecnn}ECNN~\cite{Stoffregen20eccv}}
\end{subfigure} 
\begin{subfigure}{\figWidth}
    \frame{\includegraphics[width=\linewidth]{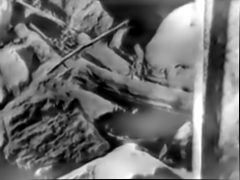}}
    \caption{\label{fig:compare:reg:bteb}BTEB~\cite{Paredes21cvpr}}
\end{subfigure}
	\caption{Image reconstruction using different regularization terms with increasing complexity 
	(continued from \cref{fig:steps}).
	Solving the linear system~\eqref{eq:general:linearsystem} without regularization already captures most of the scene content (a), 
	but the solution suffers from discretization artifacts along the optical flow streamlines.
	The regularization terms (b)-(e) mitigate the artifacts, thus improving image quality, by enforcing structural information in different ways.
	The last row shows results from state-of-the-art baselines (\cref{sec:experim}).
	\label{fig:compare:regularizers}
	}
\end{figure}

\subsection{Probabilistic Formulation and Priors}
\label{sec:method:priors}

If we assume the measurement vector $\vecniwe$ is corrupted by zero-mean additive white 
Gaussian noise $\bn$ with variance $\sigma^2$, 
the observation process may be written as $\vecniwe = D\bell + \bn$, 
yielding the conditional PDF %
$p(\vecniwe \mid \bell) = \cN(\vecniwe; D\bell, \sigma^2 \mId).$

The Maximum A Posteriori (MAP) estimate of the reconstructed brightness image $\bell$ is given by
$\hat{\bell} = \arg\max_{\bell} \,\log p(\vecniwe\mid\bell) + \log p(\bell),$
where $\log p(\bell)$ represents the prior on $\bell$. 
The estimation problem can be rewritten as
\begin{equation}
\label{eq:map:minproblem}
\hat{\bell} = \arg\min_{\bell} \, \frac1{2} \left \| \vecniwe - D\bell\right\|^2 + \lambda \cR(\bell),
\end{equation}
where the first term is the data fidelity term, and the second term is the regularization term. 
$\lambda\geq 0$ is introduced to control the degree of regularization on the reconstructed image. 
The data term makes the estimation satisfy the observation process (i.e., the events), 
while the regularization term models structural information about the desired solution.
Essentially we formulate the problem in a form that opens the door for the application of many regularizers, leveraging years of development in computer vision.
Moreover, we can benefit from learning-based regularizers (i.e., image priors).

\textbf{Priors}.
Generally, the regularizer in \eqref{eq:map:minproblem} can be classified into two categories: 
model-based priors and learning-based priors. 
In model-based regularization, the image is assumed to be smooth. 
Tikhonov regularization~\cite{Tikhonov95book} penalizes the squared norm of the gradient of the solution, 
i.e., $\cR(\bell) = \|\nabla \bell\|^2$, where the symbol $\nabla$ represents the spatial gradient of $\Lum$.
Total variation (TV) regularization~\cite{Rudin92physica} penalizes the norm, $\cR(\bell) = \|\nabla \bell\|$.
Alternatively, the Beltrami regularization~\cite{Yezzi20jmiv}, which interpolates between the TV and Tikhonov, can be used as regularizer.
The results of some of these regularizers are shown in \cref{fig:compare:regularizers}.

Among the learning-based priors, convolutional neural network (CNN) image denoisers are a popular choice in recent research. 
To use a CNN image denoiser in the process of solving \eqref{eq:map:minproblem}, we adopt the half quadratic splitting method (HQS) proposed in \cite{zhang2021plug,zhang2017learning}. 
To this end, we introduce an auxiliary variable $\bz$ that allows us to split the problem into two decoupled ones, 
and we couple them by increasing a penalty term linking $\bz$ to $\bell$ as the iterations proceed.
We replace \eqref{eq:map:minproblem} with
\begin{equation}
\label{eq:Lagrangian}
\hat{\bell} = \arg\min_{\bell, \bz} \, \frac1{2}\left \| \vecniwe - D\bell\right \|^2 + \lambda\cR(\bz) + \frac{\mu}{2}\left \| \bz - \bell\right \|^2
\end{equation}
as $\mu \to \infty$.
The problem is iteratively solved, alternating:
\begin{subequations}
\label{eq:iterative}
\begin{empheq}[left=\empheqlbrace]{align}
\label{eq:iterative:a} \bell_{k} & = \arg\min_{\bell} \, \left \| \vecniwe - D\bell\right \|^2 + \mu \left \| \bell - \bz_{k-1}\right \|^2, \\
\label{eq:iterative:b} \bz_{k} & = \arg\min_{\bz} \, \frac1{2(\sqrt{\lambda \slash \mu})^2}\left \| \bz - \bell_{k}\right \|^2 + \cR(\bz),
\end{empheq}
\end{subequations}
where the data fidelity and regularization terms are decoupled into two separate optimization problems. 
Notably, \eqref{eq:iterative:a} admits a closed-form solution:
\begin{equation}
\label{eq:closedform}
\bell_{k} = (D^{\top}D + \mu \mathrm{Id})^{-1}(D^{\top}\vecniwe + \mu \bz_{k-1}).
\end{equation}
From a Bayesian estimation point of view, \eqref{eq:iterative:b} is an image denoising problem with additive Gaussian noise (of standard deviation $\sqrt{\lambda \slash \mu}$). 
Theoretically, any Gaussian denoiser can be applied to \eqref{eq:iterative:b} to obtain a solution:
\begin{equation}
\label{eq:denoiser}
\bz_{k} = \text{Denoiser}(\bell_{k}, \sqrt{\lambda \slash \mu}).
\end{equation}
Since CNN image denoisers are shown to have a superior performance than other image denoisers, such as BM3D\cite{dabov2007color}, we adopt CNN image denoisers to solve \eqref{eq:denoiser}.
Hence, the algorithm is not strictly linear in its competitive form.

The effect of the alternating solver is that \eqref{eq:iterative:a} enforces the data fidelity equations 
while the denoiser \eqref{eq:iterative:b} pulls the solution towards the space of natural-looking images.

\subsection{Image Reconstruction and Super-resolution}
\label{sec:method:superresolution}
The proposed image reconstruction method can be naturally extended to handle super-resolution.
This is a consequence of the fact that the linear system \eqref{eq:general:linearsystem} holds at any resolution.
To achieve super-resolution we just need to provide the input NIWE $\vecniwe$ and the operator $D$ at the desired resolution.
This completely departs from end-to-end super-resolution image reconstruction approaches, 
which use supervised learning with four residual networks to realize a large encoder-decoder super-resolving network \cite{Mostafavi21pami}, 
or a 3-phase GAN with tailored and complex training datasets \cite{Wang20cvpr}.
The operator $D$ can be computed by optical flow interpolation.
The NIWE is computed by warping events~\eqref{eq:warpeventsbyflow}, \eqref{eq:IWEfromevents}, which is done in floating point precision~\cite{Gallego17ral}.
With enough warped events it is possible to provide a suitable NIWE at moderate upscaling factors, e.g., $2\times$--$4\times$.

\subsection{Image Reconstruction from its Laplacian}
\label{sec:method:LaplacianRec}

An alternative approach to reconstruct brightness from events is based on Poisson reconstruction~\cite{Kim14bmvc,Barua16wacv,Rebecq17ral,Duwek21cvprw}.
In the first step, events are used to estimate the Laplacian image, $\cE \mapsto \nabla^2 \Lum$ (\cref{fig:diagrammethods}), e.g., by means of an ANN~\cite{Duwek21cvprw}. 
In the second step, the Laplacian image is fed to a Poisson PDE solver to recover image brightness.
The key idea is that the first step is local whereas the second step is global.
That is, events only need to affect nearby pixels to create the edgemap-like image $\nabla^2 \Lum$,
which can be achieved with an ANN (few layers, small receptive fields).
The global step integrates the edgemap, filling in the brightness in regions with no events.

The relation between an image and its Laplacian can be written as a convolution $\nabla^2 \Lum(\bx) = \kappa(\bx) * \Lum(\bx)$, 
where $\kappa$ is the Laplacian kernel.
In vectorized form, it is $\veclap = \mathbf{k} \otimes \bell$, 
where $\veclap \doteq \text{vec}(\nabla^2 \Lum)$, $\mathbf{k}$ denotes the Laplacian kernel and $\otimes$ denotes convolution.
It is key to notice that this is, like~\eqref{eq:general:linearsystem}, a \emph{linear system of equations} in $\bell$.
Hence, one may combine fast Poisson solvers with image priors to recover more perceptually appealing images.
The optimization problem to recover $\bell$ from the Laplacian $\veclap$ becomes:
\begin{equation}
\label{eq:recLaplacian}
\hat{\bell} = \arg\min_{\bell}\frac1{2}\left \| \veclap - \mathbf{k} \otimes \bell\right \|^2 + \lambda\cR(\bell).
\end{equation}

\global\long\def\figWidth{0.155\linewidth}
\begin{figure*}[t]
    \ifhideimages
    \else
	\centering
    {\small
    \setlength{\tabcolsep}{2pt}
	\begin{tabular}{
	>{\centering\arraybackslash}m{0.3cm}
	>{\centering\arraybackslash}m{\figWidth} 
	>{\centering\arraybackslash}m{\figWidth}
	>{\centering\arraybackslash}m{\figWidth}
	>{\centering\arraybackslash}m{\figWidth}
	>{\centering\arraybackslash}m{\figWidth}
	>{\centering\arraybackslash}m{\figWidth}}

		\rotatebox{90}{\makecell{boxes\_rot}}
		&\frame{\includegraphics[width=\linewidth]{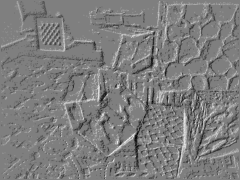}}
		&\frame{\includegraphics[width=\linewidth]{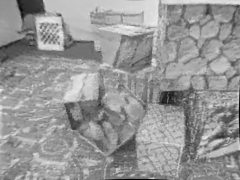}}
		&\frame{\includegraphics[width=\linewidth]{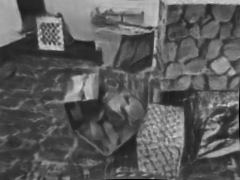}}
		&\frame{\includegraphics[width=\linewidth]{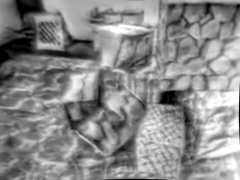}}
		&\frame{\includegraphics[width=\linewidth]{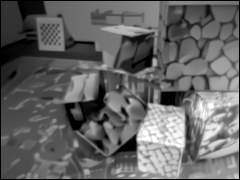}}
		&\frame{\includegraphics[width=\linewidth]{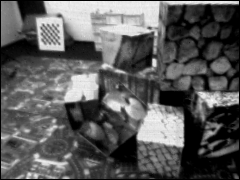}}
		\\

		\rotatebox{90}{\makecell{poster\_rot}}
		&\frame{\includegraphics[width=\linewidth]{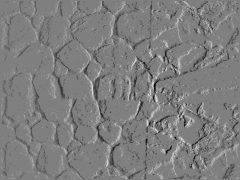}}
		&\frame{\includegraphics[width=\linewidth]{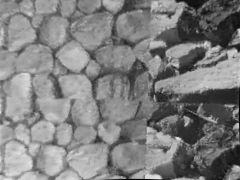}}
		&\frame{\includegraphics[width=\linewidth]{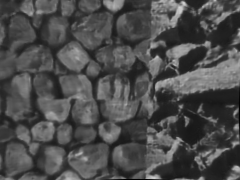}}
		&\frame{\includegraphics[width=\linewidth]{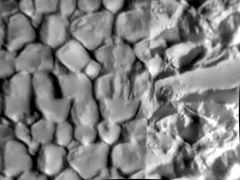}}
		&\frame{\includegraphics[width=\linewidth]{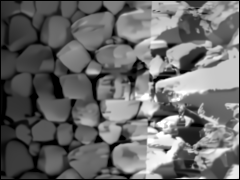}}
		&\frame{\includegraphics[width=\linewidth]{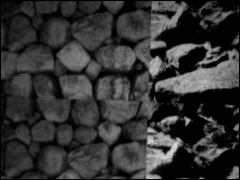}}
		\\
		
		\rotatebox{90}{\makecell{dynamic\_rot}}
		&\frame{\includegraphics[width=\linewidth]{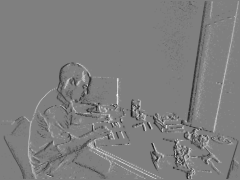}}
		&\frame{\includegraphics[width=\linewidth]{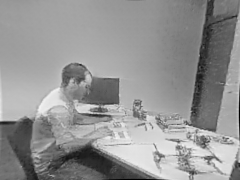}}
		&\frame{\includegraphics[width=\linewidth]{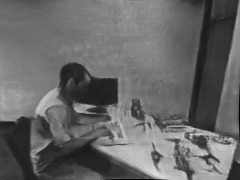}}
		&\frame{\includegraphics[width=\linewidth]{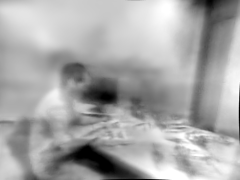}}
		&\frame{\includegraphics[width=\linewidth]{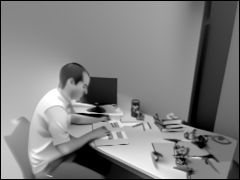}}
		&\frame{\includegraphics[width=\linewidth]{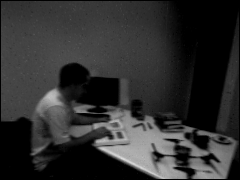}}
		\\
		
		\rotatebox{90}{\makecell{shapes\_rot}}
		&\frame{\includegraphics[width=\linewidth]{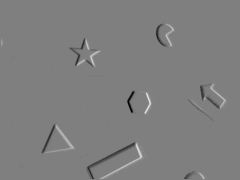}}
        &\frame{\includegraphics[width=\linewidth]{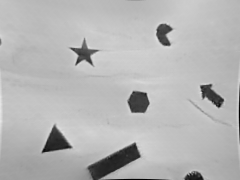}}
		&\frame{\includegraphics[width=\linewidth]{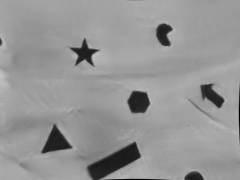}}
		&\frame{\includegraphics[width=\linewidth]{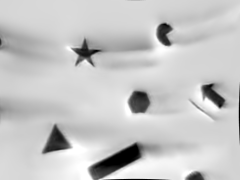}}
		&\frame{\includegraphics[width=\linewidth]{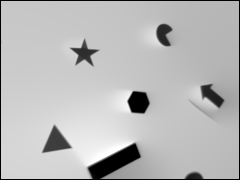}}
		&\frame{\includegraphics[width=\linewidth]{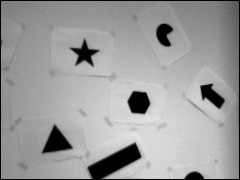}}
		\\

		\rotatebox{90}{\makecell{slider\_hdr\_far}}
		&\frame{\includegraphics[width=\linewidth]{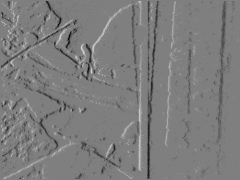}}
		&\frame{\includegraphics[width=\linewidth]{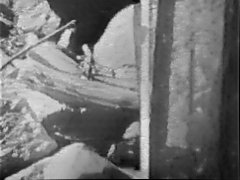}}
		&\frame{\includegraphics[width=\linewidth]{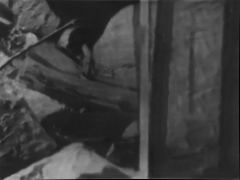}}
		&\frame{\includegraphics[width=\linewidth]{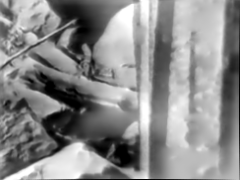}}
		&\frame{\includegraphics[width=\linewidth]{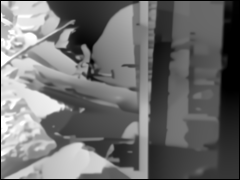}}
		&\frame{\includegraphics[width=\linewidth]{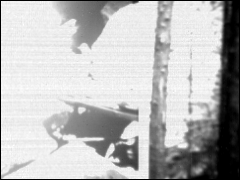}}
		\\
		
		\rotatebox{90}{\makecell{slider\_hdr\_close}}
		&\frame{\includegraphics[width=\linewidth]{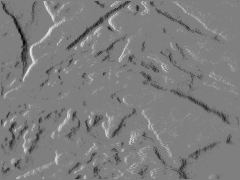}}
		&\frame{\includegraphics[width=\linewidth]{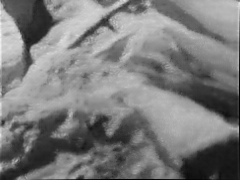}}
		&\frame{\includegraphics[width=\linewidth]{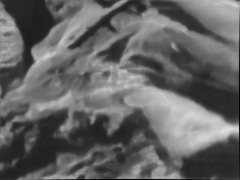}}
		&\frame{\includegraphics[width=\linewidth]{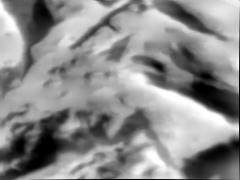}}
		&\frame{\includegraphics[width=\linewidth]{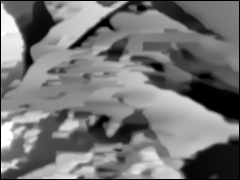}}
		&\frame{\includegraphics[width=\linewidth]{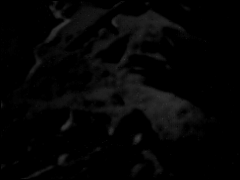}}
		\\

		& (a) NIWE (events)
		& (b) E2VID~\cite{Rebecq19pami}
		& (c) ECNN~\cite{Stoffregen20eccv}
		& (d) BTEB~\cite{Paredes21cvpr}
		& (e) Ours~\eqref{eq:iterative}, CNN
		& (f) DAVIS frames
	\end{tabular}
	}
	\fi
	\caption{Qualitative comparison with the state of the art on sequences from~\cite{Mueggler17ijrr}. 
	Histogram equalization is not used. %
	}
	\label{fig:compare:ijrr}
\end{figure*}

This problem can be solved by the HQS method, with similar steps as those in \cref{sec:method:priors}. 
Equation \eqref{eq:denoiser} still applies, 
and if we assume the reconstructed image has periodic boundary conditions, the subproblem in the data fidelity term has closed-form solution (the analogue of~\eqref{eq:closedform}):
\begin{equation}
\label{eq:closedformFourier}
\bell_{k} = \cF^{-1}\left ( \frac{\overline{\cF(\mathbf{k})}\cF(\veclap) + \mu \cF(\bz_{k-1})}
{\overline{\cF(\mathbf{k})}\cF(\mathbf{k}) + \mu} \right),
\end{equation}
where $\cF(\cdot)$ represents the Discrete Fourier Transform (DFT), 
$\cF^{-1}(\cdot)$ is the inverse DFT, 
and $\overline{\cF(\cdot)}$ denotes the complex conjugate of $\cF(\cdot)$. 
Without periodic boundary conditions, one may replace the DFT with the Discrete Cosine Transform (DCT) or the Discrete Sine Transform (DST) based on different symmetry assumptions.

In summary, estimation problems \eqref{eq:map:minproblem} and \eqref{eq:recLaplacian} share many similarities:
both arise from linear systems of equations on the derivatives of the brightness, encoded by events,
and both can be solved using classical techniques (TV regularizer, etc.) as well as more modern ones (image priors),
without event-to-image ANNs.
There are, of course, differences, since \eqref{eq:map:minproblem} is based on the first derivative and \eqref{eq:recLaplacian} is based on the second derivative:
($i$) the Laplacian $\veclap$ is a motion-invariant representation, whereas the NIWE is not (it is motion dependent); 
hence ($ii$) the Laplacian operator $\mathbf{k} \otimes$ in \eqref{eq:recLaplacian} is isotropic, whereas $D$ in \eqref{eq:map:minproblem} is not.

\subsection{Extensions}
\label{sec:method:extensions}

The proposed method is versatile: it can not only handle super-resolution, but also be combined with motion segmentation approaches and used to recover color images.

\subsubsection{Motion Segmentation}
\label{sec:method:motseg}

Our method can be combined with event-based motion segmentation methods \cite{Stoffregen19iccv,Zhou21tnnls}.
Such methods simultaneously classify the input events $\cE$ into $N_c$ clusters corresponding to different moving parts of the scene and estimate the motion of the clusters:
\begin{equation}
    \label{eq:motseg:io}
    \cE \mapsto \{\cE_j,\btheta_j\}_{j=1}^{N_c}.
\end{equation}
Each cluster has an associated IWE, whose contrast is maximized by the events $\cE_j$ that belong to the cluster.
The motion parameters $\btheta_j$ provide accurate information about optical flow.
Hence, the IWEs and motion parameters obtained via motion segmentation can be used as input to our method. 
As a result, the brightness of each moving part of the scene is recovered.
Event-based motion segmentation with subsequent image reconstruction has the effect of splitting a reconstructed image into its moving parts.

\begin{figure*}[t!] %
\centering
\captionof{table}{
Quantitative evaluation of our method and the state of the art on sequences from~\cite{Mueggler17ijrr}.
We report median values (since they are more robust to outliers than the mean) of MSE, SSIM and LPIPS quality metrics over all reconstructed images.
Images are equalized before computing the metrics.
\label{tab:imgrec:ijrr:equalized}
}
\begin{adjustbox}{max width=\linewidth}
\setlength{\tabcolsep}{2pt}
\begin{tabular}{@{}l *{18}{S[table-format=1.4]}@{}}
             & \multicolumn{6}{c}{MSE $\downarrow$} & \multicolumn{6}{c}{SSIM $\uparrow$} & \multicolumn{6}{c}{LPIPS $\downarrow$}\\
             \cmidrule(l{2mm}r{2mm}){2-7} \cmidrule(l{2mm}r{2mm}){8-13} \cmidrule(l{2mm}r{2mm}){14-19} %
               & & & & \multicolumn{3}{c}{Ours}
               & & & & \multicolumn{3}{c}{Ours}
               & & & & \multicolumn{3}{c}{Ours}\\
             \cmidrule(l{2mm}r{2mm}){5-7} \cmidrule(l{2mm}r{2mm}){11-13} \cmidrule(l{2mm}r{2mm}){17-19} %
Sequence name  & $\text{E2VID}$ & $\text{ECNN}$ & $\text{BTEB}$ & $\text{Tikh.}$ & $\text{TV}$ & $\text{CNN}$
               & $\text{E2VID}$ & $\text{ECNN}$ & $\text{BTEB}$ & $\text{Tikh.}$ & $\text{TV}$ & $\text{CNN}$
               & $\text{E2VID}$ & $\text{ECNN}$ & $\text{BTEB}$ & $\text{Tikh.}$ & $\text{TV}$ & $\text{CNN}$\\
	
\midrule
boxes\_rotation & \bnum{0.04578163} & \unum{0.04908111} & 0.06580944 & 0.11448207 & 0.10953972 & 0.09000321 
                & \unum{0.48920843} & \bnum{0.5353875} & 0.45597288 & 0.4306791 & 0.43688878 & 0.48708642
                & \unum{0.3893156} & \bnum{0.3702461} & 0.4453549 & 0.41472387 & 0.42406112 & 0.44492048\\
poster\_rotation  & \unum{0.04182485} & \bnum{0.03622744} & 0.06340547 & 0.13272578 & 0.12765434 & 0.07532048 
                & \unum{0.5156613886356354} & \bnum{0.5559019148349762} & 0.4560096263885498 & 0.38392913341522217 & 0.3957895040512085 & 0.5066708326339722 
                & \bnum{0.33636343} & \unum{0.3689409} & 0.45860404 & 0.43664682 & 0.44562685 & 0.43891567 \\
dynamic\_rotation  & \unum{0.07379299} & \bnum{0.07286617} & 0.11927439 & 0.1354472 & 0.123047434 & 0.104789436 & 0.2920428514480591 & 0.31260573863983154 & 0.26437908411026 & 0.3123185634613037 & \unum{0.3337557911872864} & \bnum{0.3551797866821289} & \unum{0.48696864} & \bnum{0.4777416} & 0.5659355 & 0.5187025 & 0.5363434 & 0.49947798 \\
shapes\_rotation  & 0.17417653 & \unum{0.10286203} & 0.14803968 & 0.13512397 & 0.1436538 & \bnum{0.09018336} & 0.2564120292663574 & 0.22544604539871216 & 0.23969781398773193 & 0.340598464012146 & \unum{0.3979228734970093} & \bnum{0.4189654588699341} & \bnum{0.54103523} & 0.55461067 & 0.64298844 & 0.5536429 & 0.6148522 & \unum{0.5458323} \\
slider\_far  & \unum{0.05327062} & \bnum{0.04910804} & 0.0857418 & 0.12695538 & 0.10089316 & 0.06945322 & 0.40828532 & \unum{0.4155444} & 0.37443316 & 0.3183536 & 0.37908185 & \bnum{0.443735} & \bnum{0.3755945} & \unum{0.3824887} & 0.44557685 & 0.44254404 & 0.4443134 & 0.450082 \\
slider\_close  & \bnum{0.02628192} & \unum{0.02639169} & 0.05473252 & 0.08220527 & 0.07066046 & 0.04977795 & \unum{0.41080597} & \bnum{0.45148358} & 0.35230824 & 0.34755015 & 0.37251177 & 0.4075567 & \unum{0.49994627} & \bnum{0.49847627} & 0.56060934 & 0.5424131 & 0.54484665 & 0.53221226 \\
\midrule
Average      & \unum{0.06918809} & \bnum{0.05608941} & 0.08950055 & 0.12115661 & 0.11257482 & 0.07992128 & 0.39540266 & \unum{0.41606153} & 0.35713347 & 0.3555715 & 0.38599176 & \bnum{0.43653237} & \bnum{0.43820394} & \unum{0.44208404} & 0.51984485 & 0.48477887 & 0.50167394 & 0.48524012 \\ 
\end{tabular}
\end{adjustbox}  

\ifhideimages
\else
    \def\figWidth{0.24\linewidth}
    {\small
    \setlength{\tabcolsep}{1pt}
	\begin{tabular}{
	>{\centering\arraybackslash}m{\figWidth}
	>{\centering\arraybackslash}m{\figWidth}
	>{\centering\arraybackslash}m{\figWidth}
	>{\centering\arraybackslash}m{\figWidth}}
		\\

		 {\includegraphics[trim=1.0cm 0.15cm 3.5cm 0.6cm,clip, width=\linewidth]{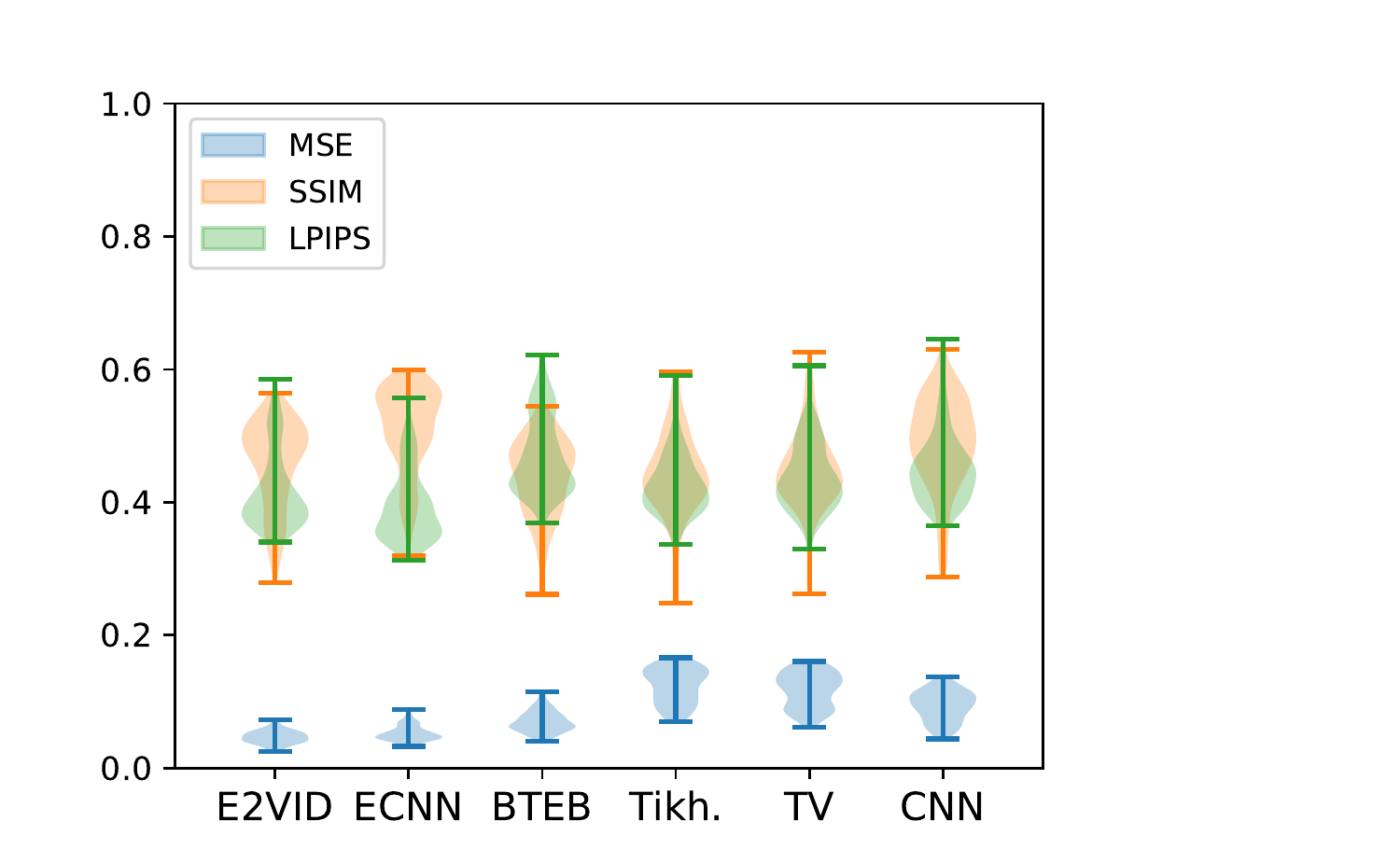}}
		&{\includegraphics[trim=1.0cm 0.15cm 3.5cm 0.6cm,clip, width=\linewidth]{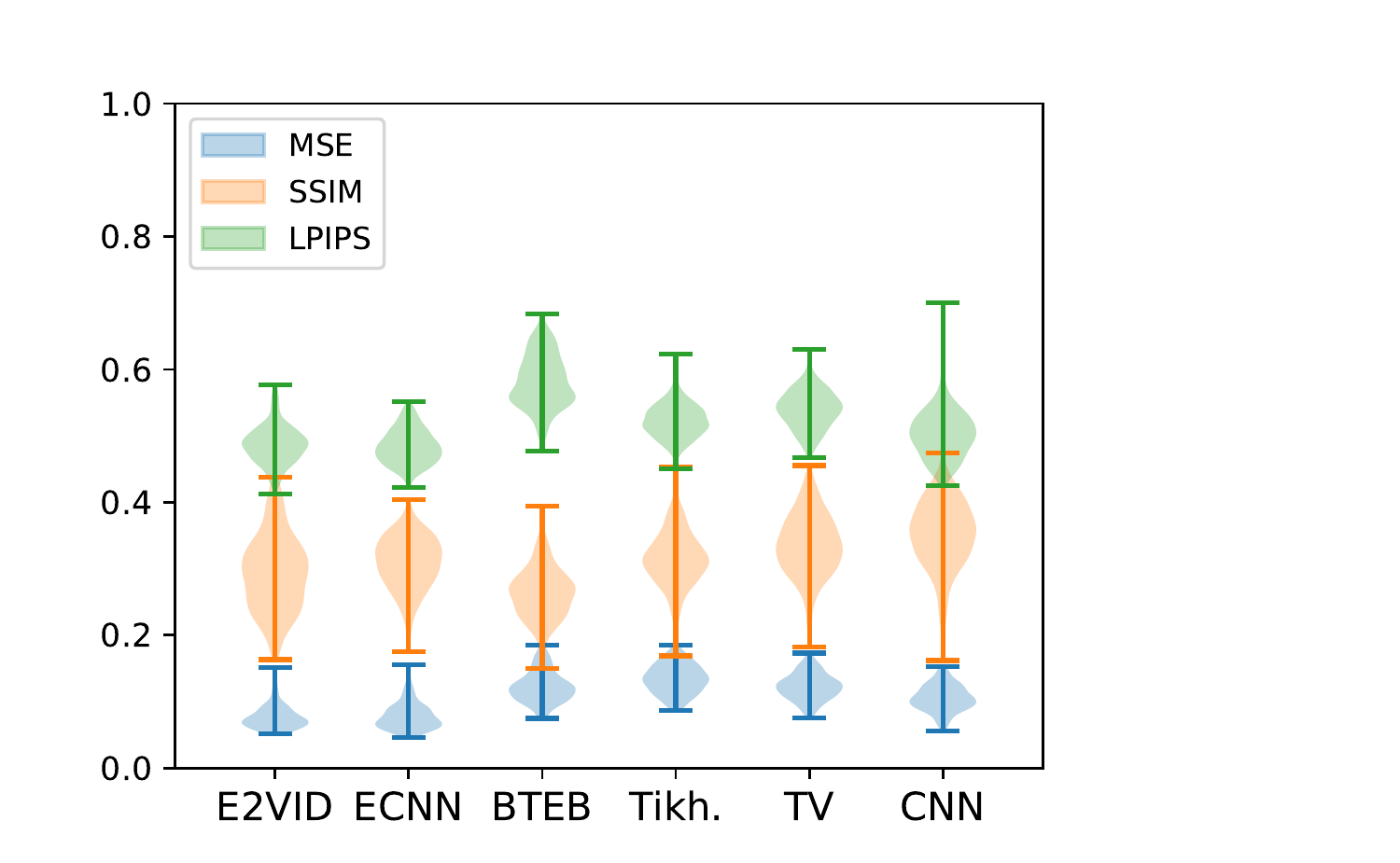}}
		&{\includegraphics[trim=1.0cm 0.15cm 3.5cm 0.6cm,clip, width=\linewidth]{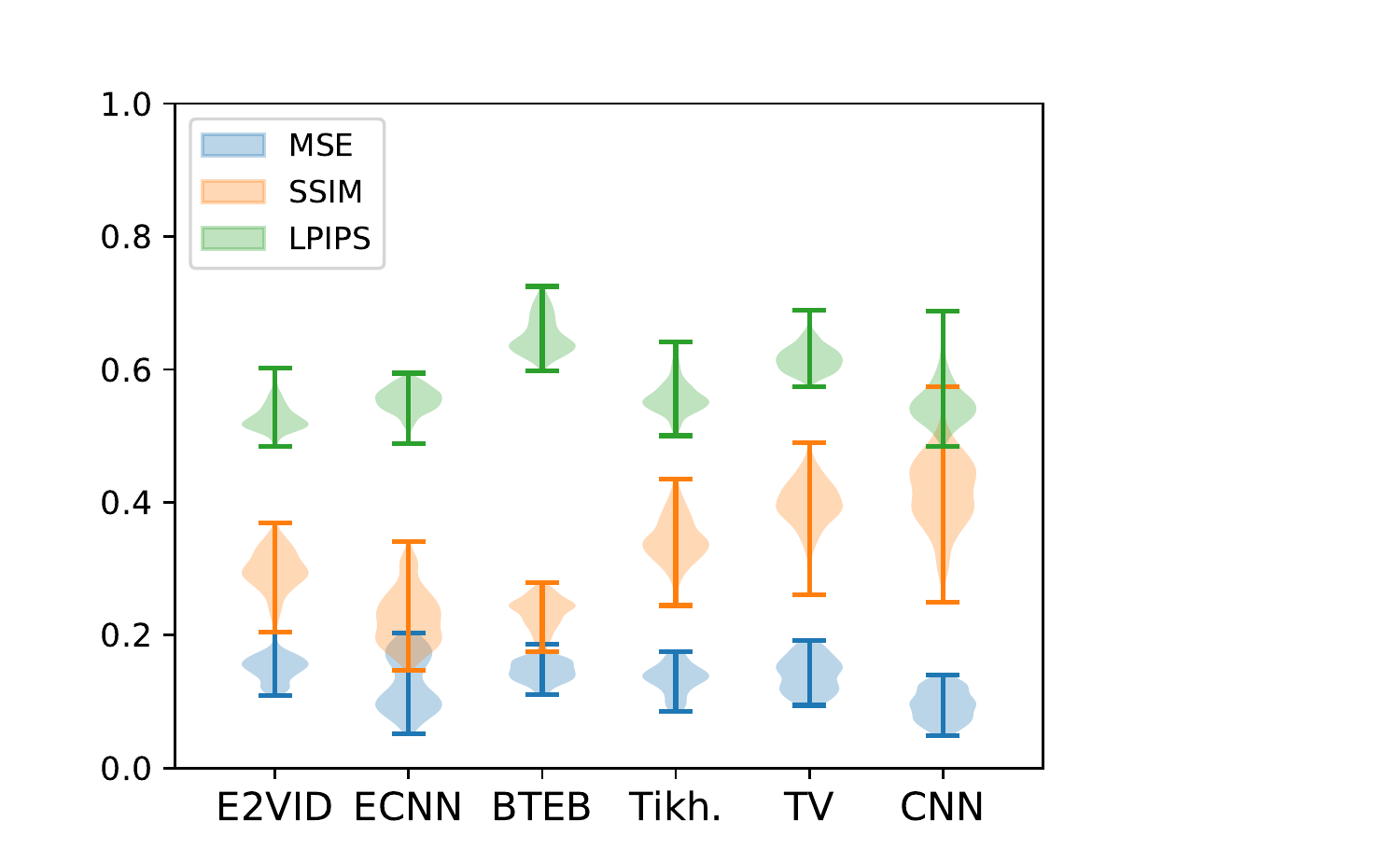}}
		&{\includegraphics[trim=1.0cm 0.15cm 3.5cm 0.6cm,clip, width=\linewidth]{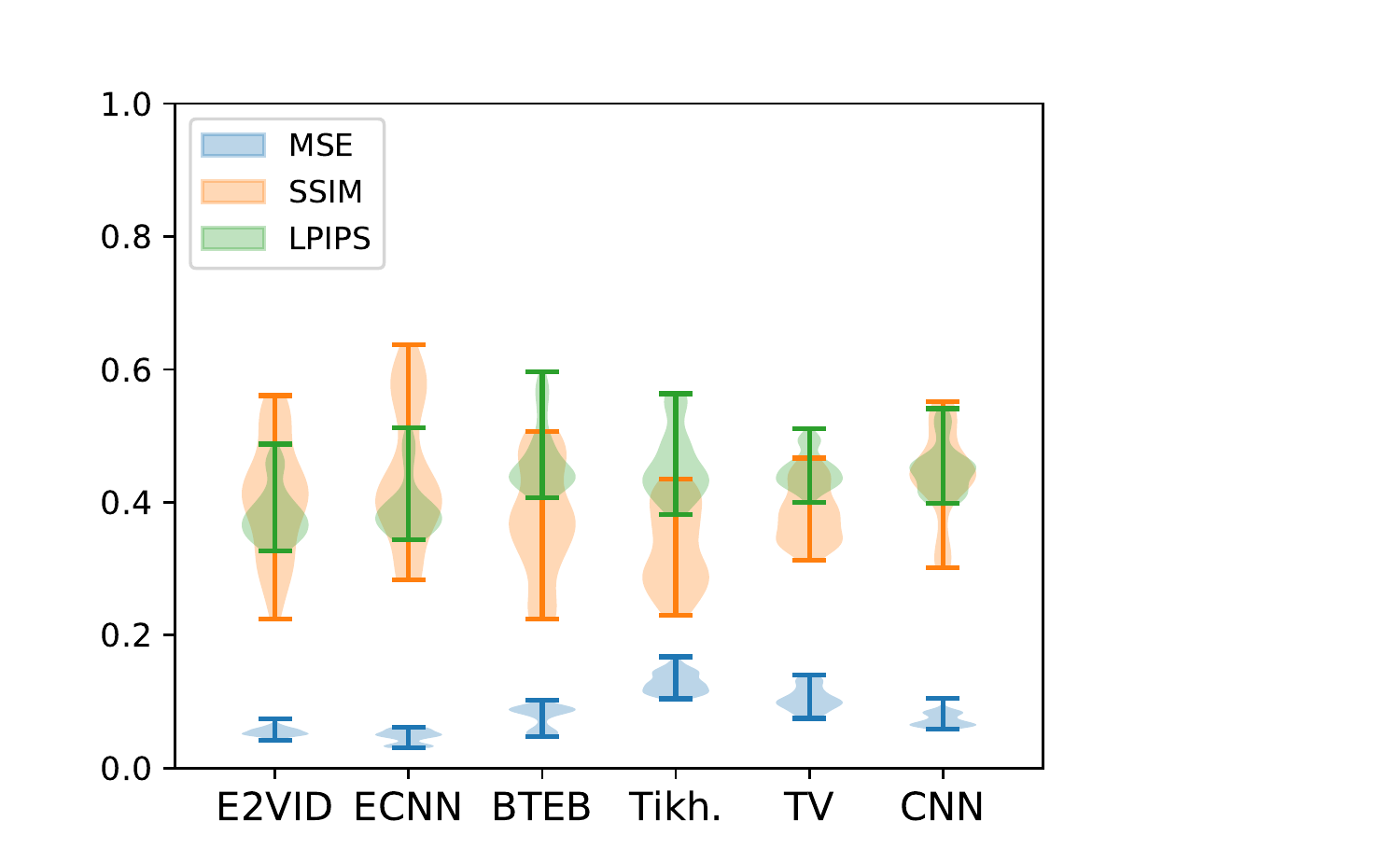}}
		\\
		
		(a) Boxes
		& (b) Dynamic
		& (c) Shapes
		& (d) Slider\_far
		\\
	\end{tabular}
	}

\fi
\vspace{0.8ex}
\captionof{figure}{
Corresponding distribution plots of the six methods in \cref{tab:imgrec:ijrr:equalized}. 
Number of reconstructed images per sequence:
(a)-(c) 300, (d) 136 (the whole sequence).
Histogram-equalized images.
\label{fig:violin:eq}
}
\end{figure*}

\subsubsection{Color Image Reconstruction}
\label{sec:method:color}

Color image reconstruction from events is possible if events are produced by a color event camera such as the color-DAVIS346 \cite{Taverni18tcsii}, which has a Bayer filter mosaic that makes each pixel sensitive to red, green or blue light.
A simple strategy to perform demosaicing from the RGB color channels is to reconstruct each channel separately at their undersampled resolution and then upsample to the original resolution using bicubic interpolation \cite{Scheerlinck19cvprw}.
However, since our method can naturally handle super-resolution, we perform demosaicing by combining the reconstruction and interpolation steps, as follows:
($i$) reconstruct and super-resolve (by a factor of $2\times$) each color channel (R,G,B) from their corresponding events (\cref{sec:method:superresolution}), 
($ii$) concatenate the resulting (R,G,B) channels to generate a full-size color image.
Color may be useful in applications such as object recognition and microscopy imaging.

\section{Experiments}
\label{sec:experim}

This section presents the evaluation of our proposal.
Mimicking prior work, a large portion of the experiments (\cref{sec:experim:imgquality}) focus on assessing the quality of the reconstructed images on standard datasets and comparing to competing baselines in the state of the art.
\Cref{sec:experim:sr} shows super-resolution experiments.
\Cref{sec:experim:laplacian} demonstrates image reconstruction via the brightness Laplacian.
\Cref{sec:experim:analysis} analyzes: the evolution of the three solvers (CNN, TV, Tikhonov) as iterations proceed, 
the effect of varying the amount of regularization, and a serendipitous application to monitoring.
\Cref{sec:experim:extensions} shows the capabilities of the IWE method to be combined with motion segmentation and color events.
\Cref{sec:experim:denseflow} explores the combination of the IWE method with recent dense optical flow estimation methods 
and evaluates the influence of erroneous optical flow.
Finally, \cref{sec:experim:runtime} reports computational performance and \cref{sec:limitations} points outs the limitations of the method.

\subsection{Image Quality Assessment}
\label{sec:experim:imgquality}

\subsubsection{Datasets, Metrics and Baselines}
\label{sec:experim:imgquality:datasets}

We evaluate the performance of our approaches on sequences from standard datasets~\cite{Mueggler17ijrr,Orchard15fns}.
We compare with state-of-the-art image reconstruction methods \cite{Rebecq19pami,Stoffregen20eccv,Paredes21cvpr}.
E2VID \cite{Rebecq19pami} and ECNN \cite{Stoffregen20eccv} are event-to-image RNNs, trained in a supervised manner, with ground truth images. %
ECNN is an improved version of \cite{Rebecq19pami} trained on augmented data to reduce the sim-to-real gap.
BTEB \cite{Paredes21cvpr} comprises both an optical flow estimation CNN \cite{Zhu19cvpr} and an image reconstruction RNN (same architecture as \cite{Rebecq19pami}), where the latter is trained in a self-supervised manner, without ground-truth images.

As is standard, we reconstruct images at the timestamps of the ground truth images (e.g., DAVIS frames) and compare in terms of mean square error (MSE), structural similarity (SSIM)~\cite{Wang04tip} and perceptual similarity (LPIPS)~\cite{Zhang18cvprLPIPS}.
The sequences from \cite{Mueggler17ijrr} have 60s duration and an increasing motion speed. 
Following the sequence cuts proposed by \cite{Stoffregen20eccv}, we use the timestamps of the 300 frames in $[5,20]$s before motion blur starts to corrupt the ground truth.
For the slider sequences we reconstructed images at all $>135$ timestamps of the frames because there is little motion blur.

We use the last 20k--50k\,events per IWE, depending on the amount of texture in the scene.
Motion was estimated by means of contrast maximization \cite{Gallego18cvpr,Gu21iccv} on the same set of events, from which optical flow was computed and fed to our method.
The range of values of the regularizer weight is: 
$\lambda \in [0.03,0.05]$ for Tikh.~and TV, and $\lambda \in [0.19, 0.55]$ (see \cite{zhang2021plug}) for the image prior denoiser.

\begin{figure*}[t!] %
\centering
\captionof{table}{
Quantitative evaluation like \cref{tab:imgrec:ijrr:equalized}, but without histogram equalization.
\label{tab:imgrec:ijrr:noneq}
}
\begin{adjustbox}{max width=\linewidth}
\setlength{\tabcolsep}{2pt}
\begin{tabular}{@{}l *{18}{S[table-format=1.4]}@{}}
             & \multicolumn{6}{c}{MSE $\downarrow$} & \multicolumn{6}{c}{SSIM $\uparrow$} & \multicolumn{6}{c}{LPIPS $\downarrow$}\\
             \cmidrule(l{2mm}r{2mm}){2-7} \cmidrule(l{2mm}r{2mm}){8-13} \cmidrule(l{2mm}r{2mm}){14-19} %
               & & & & \multicolumn{3}{c}{Ours}
               & & & & \multicolumn{3}{c}{Ours}
               & & & & \multicolumn{3}{c}{Ours}\\
             \cmidrule(l{2mm}r{2mm}){5-7} \cmidrule(l{2mm}r{2mm}){11-13} \cmidrule(l{2mm}r{2mm}){17-19} %
Sequence name  & $\text{E2VID}$ & $\text{ECNN}$ & $\text{BTEB}$ & $\text{Tikh.}$ & $\text{TV}$ & $\text{CNN}$
               & $\text{E2VID}$ & $\text{ECNN}$ & $\text{BTEB}$ & $\text{Tikh.}$ & $\text{TV}$ & $\text{CNN}$
               & $\text{E2VID}$ & $\text{ECNN}$ & $\text{BTEB}$ & $\text{Tikh.}$ & $\text{TV}$ & $\text{CNN}$\\
	
\midrule
boxes\_rotation & 0.0753691 & \bnum{0.03918463} & 0.09105358 & 0.10148385 & 0.1022459 & \unum{0.07256413}
                & 0.44887006 & \bnum{0.5103005} & 0.4324064 & 0.4546076 & 0.4628192 & \unum{0.50858533} & \unum{0.39975435} & \bnum{0.37712198} & 0.44515276 & 0.4051572 & 0.41118973 & 0.44313562\\
poster\_rotation  & 0.11537936 & \bnum{0.02600935} & 0.12406802 & 0.11726413 & 0.11895417 & \unum{0.09122612}
                & 0.39130044 & \bnum{0.53031814} & 0.36952007 & 0.4551787 & 0.45658645 & \unum{0.4873173}
                & \unum{0.37512076} & \bnum{0.360834} & 0.46317947 & 0.41909668 & 0.42878774 & 0.43823177 \\
dynamic\_rotation  & 0.17411229 & \bnum{0.07404337} & 0.16692267 & 0.13842851 & 0.15823445 & \unum{0.1319541} & 0.31322494 & \unum{0.39662355} & 0.30803385 & \bnum{0.39698476} & 0.3791154 & 0.3962708 & 0.5176058 & \bnum{0.47572225} & 0.5822724 & 0.5001141 & 0.49767628 & \unum{0.49043232} \\
shapes\_rotation  & 0.02710731 & \bnum{0.014268} & 0.05093651 & 0.02252861 & 0.04354482 & \unum{0.02228115} & 0.7578524 & 0.75627595 & 0.648091 & 0.7565873 & \unum{0.80140007} & \bnum{0.80746186} & 0.443548 & 0.45528013 & 0.59952724 & \unum{0.43816474} & 0.4412835 & \bnum{0.4134661} \\
slider\_far  & 0.0965981 & \bnum{0.02945322} & 0.09567116 & 0.12013268 & 0.09357519 & \unum{0.06791549} & 0.32636857 & \unum{0.3813874} & 0.32658064 & 0.2826559 & 0.32985765 & \bnum{0.39241976} & \unum{0.40688032} & \bnum{0.37264472} & 0.46869162 & 0.45428014 & 0.4469846 & 0.4587828 \\
slider\_close  & \bnum{0.04327465} & 0.08400747 & 0.08085614 & 0.09870792 & 0.08306082 & \unum{0.07303864} & \unum{0.48837513} & 0.424591 & 0.42159835 & 0.44342873 & 0.46751752 & \bnum{0.52178806} & 0.50084925 & \bnum{0.48870212} & 0.5229534 & 0.51260495 & 0.5025073 & \unum{0.49568242} \\
\midrule
Average      & 0.08864014 & \bnum{0.04449434} & 0.10158468 & 0.09975762 & 0.09993589 & \unum{0.0764966} & 0.45433192 & \unum{0.49991609} & 0.41770505 & 0.46490717 & 0.48288271 & \bnum{0.51897385} & \unum{0.44062641} & \bnum{0.42171753} & 0.51362948 & 0.45490297 & 0.45473819 & 0.45662184 \\ 
\end{tabular}
\end{adjustbox}  

\ifhideimages
\else
    \def\figWidth{0.24\linewidth}
    {\small
    \setlength{\tabcolsep}{1pt}
	\begin{tabular}{
	>{\centering\arraybackslash}m{\figWidth}
	>{\centering\arraybackslash}m{\figWidth}
	>{\centering\arraybackslash}m{\figWidth}
	>{\centering\arraybackslash}m{\figWidth}}
		\\

		 {\includegraphics[trim=1.0cm 0.15cm 3.5cm 0.6cm,clip, width=\linewidth]{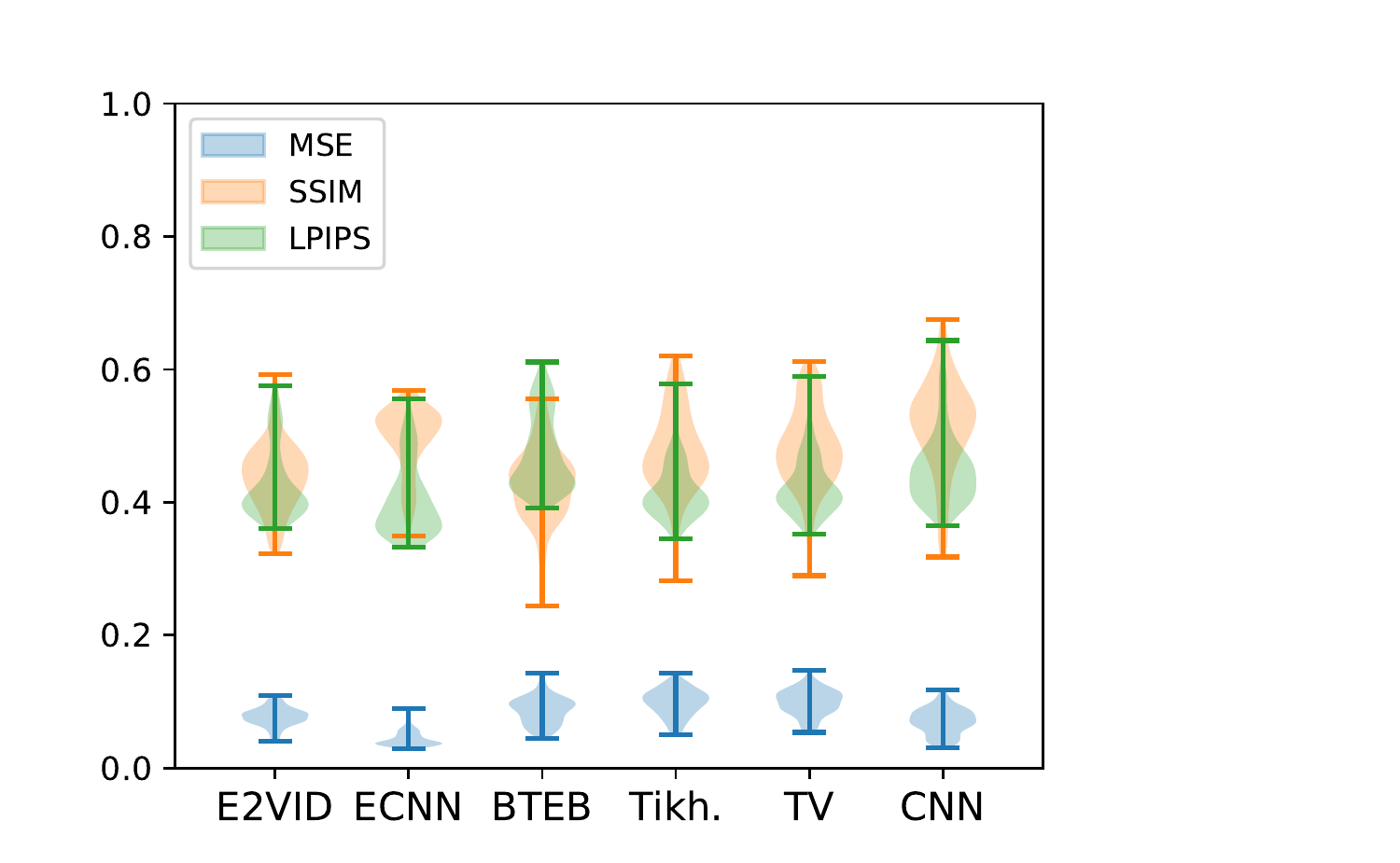}}
		&{\includegraphics[trim=1.0cm 0.15cm 3.5cm 0.6cm,clip, width=\linewidth]{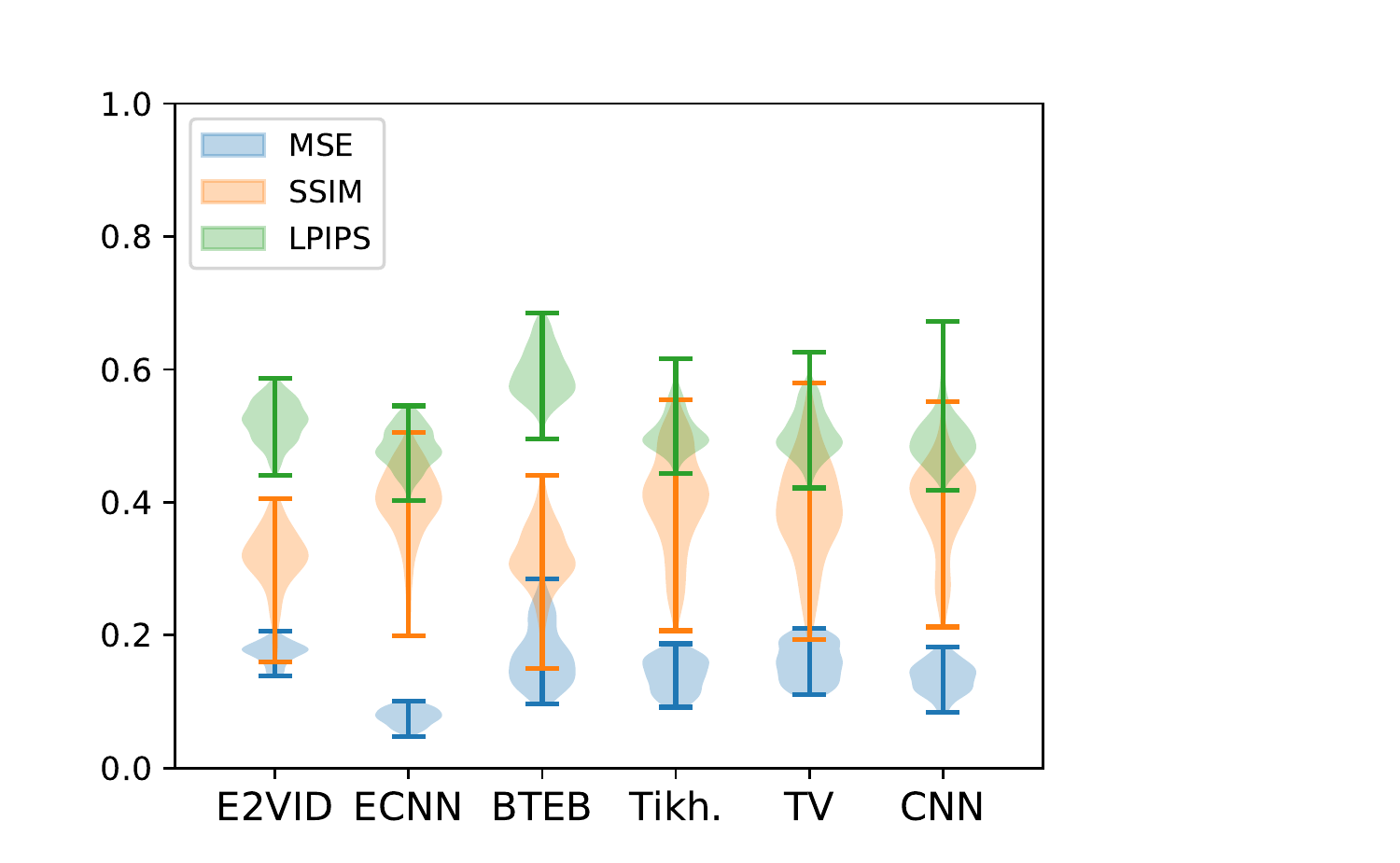}}
		&{\includegraphics[trim=1.0cm 0.15cm 3.5cm 0.6cm,clip, width=\linewidth]{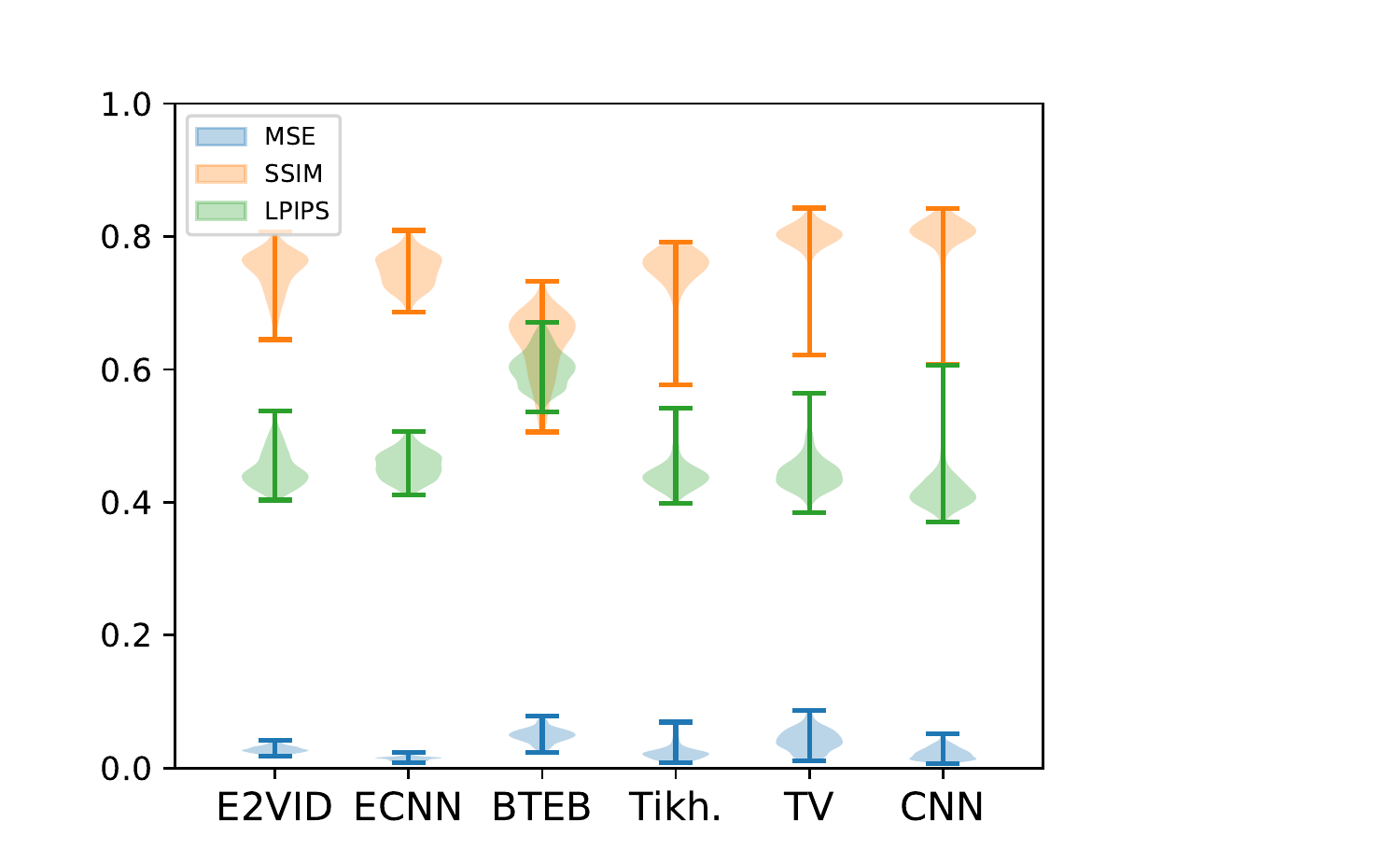}}
		&{\includegraphics[trim=1.0cm 0.15cm 3.5cm 0.6cm,clip, width=\linewidth]{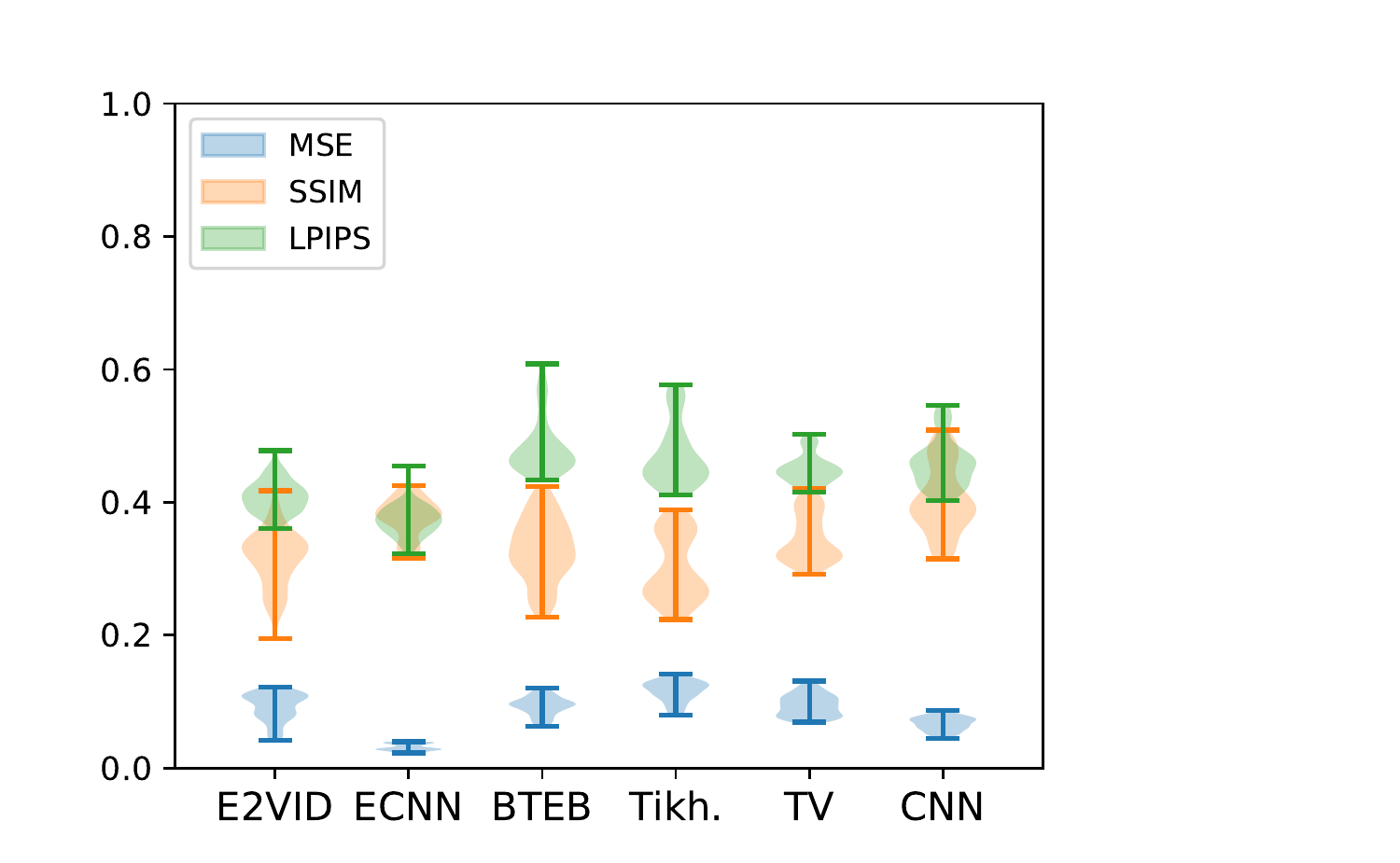}}
		\\
		
		(a) Boxes
		& (b) Dynamic
		& (c) Shapes
		& (d) Slider\_far
		\\
	\end{tabular}
	}

\fi
\vspace{0.8ex}
\captionof{figure}{
Corresponding distribution plots of the six methods in \cref{tab:imgrec:ijrr:noneq}. \emph{Non}-equalized images.
\label{fig:violin:noeq}
}
\end{figure*}

\subsubsection{Results on the Event Camera Dataset}

\Cref{fig:compare:ijrr} shows qualitative results on the Event Camera Dataset \cite{Mueggler17ijrr}.
As observed, our method~\eqref{eq:iterative} produces results on par with the state of the art. 
E2VID and ECNN produce high quality results, partly due to the large capacity of these RNNs to predict pixel intensities and because they are supervised methods.
BTEB suffers from motion blur and ``ghosting'' artifacts (most noticeably in textureless regions) \cite{Paredes21cvpr}.
Our method also suffers from some visual artifacts, but they are less pronounced than those of BTEB. 
The reconstructed images have a clean, smooth appearance, which is conferred by the image denoiser within the solver~\eqref{eq:iterative:b}.
Our method fills with a smooth clean look the textureless regions (we also observe this in the different variants presented in \cref{fig:compare:regularizers}), 
whereas BTEB is reported to have limited extrapolation capacity of edge information.
This is specially noticeable in the shapes and dynamic sequences.
However, our regularizer tends to smooth fine-grained details (e.g., within-rock texture) as they are considered noise.

The last two rows of \cref{fig:compare:ijrr} show HDR results. 
In these sequences comparison is only qualitatively since ground truth images are not HDR.
Our method is able to reconstruct high quality HDR images, on par with learning-based methods, 
and showing a smooth look, which can be controlled %
via the regularizer weight.

\def\figWidth{0.135\linewidth}
\newlength{\waccordion} \setlength{\waccordion}{928px} %
\newlength{\haccordion} \setlength{\haccordion}{344px} %
\newlength{\wairplanes} \setlength{\wairplanes}{928px} %
\newlength{\hairplanes} \setlength{\hairplanes}{220px} %
\newlength{\wanchor} \setlength{\wanchor}{816px} %
\newlength{\hanchor} \setlength{\hanchor}{344px} %
\newlength{\want} \setlength{\want}{928px} %
\newlength{\hant} \setlength{\hant}{328px} %
\newlength{\wbarrel} \setlength{\wbarrel}{624px} %
\newlength{\hbarrel} \setlength{\hbarrel}{344px} %
\newlength{\wbass} \setlength{\wbass}{688px} %
\newlength{\hbass} \setlength{\hbass}{344px} %
\newlength{\wbeaver} \setlength{\wbeaver}{928px} %
\newlength{\hbeaver} \setlength{\hbeaver}{304px} %
\newlength{\wtwo} \setlength{\wtwo}{768px} %
\newlength{\htwo} \setlength{\htwo}{344px} %
\newlength{\wbinocular} \setlength{\wbinocular}{544px} %
\newlength{\hbinocular} \setlength{\hbinocular}{344px} %
\newlength{\wbrontosaurus} \setlength{\wbrontosaurus}{928px} %
\newlength{\hbrontosaurus} \setlength{\hbrontosaurus}{344px} %
\newlength{\wbuddha} \setlength{\wbuddha}{928px} %
\newlength{\hbuddha} \setlength{\hbuddha}{344px} %
\newlength{\wcamera} \setlength{\wcamera}{688px} %
\newlength{\hcamera} \setlength{\hcamera}{344px} %
\newlength{\wcellphone} \setlength{\wcellphone}{720px} %
\newlength{\hcellphone} \setlength{\hcellphone}{344px} %
\newlength{\wcup} \setlength{\wcup}{784px} %
\newlength{\hcup} \setlength{\hcup}{344px} %

\begin{figure*}[t]
    \ifhideimages
    \else
	\centering
    {\small
    \setlength{\tabcolsep}{1pt}
	\begin{tabular}{
	>{\centering\arraybackslash}m{0.3cm}
	>{\centering\arraybackslash}m{\figWidth}
	>{\centering\arraybackslash}m{\figWidth}
	>{\centering\arraybackslash}m{\figWidth}
	>{\centering\arraybackslash}m{\figWidth}
	>{\centering\arraybackslash}m{\figWidth}
	>{\centering\arraybackslash}m{\figWidth}
	>{\centering\arraybackslash}m{\figWidth}}
        	
		\rotatebox{90}{\makecell{ant}}
		&{\includegraphics[trim=0px 0px {0.75\want} {0.5\hant},clip, width=\linewidth]{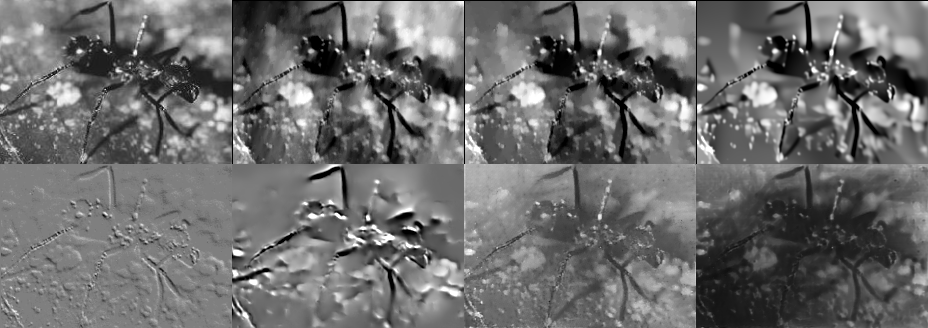}}%
		&{\includegraphics[trim={0.5\want} 0px {0.25\want} {0.5\hant},clip, width=\linewidth]{images/caltech/ant.png}}%
		&{\includegraphics[trim={0.75\want} 0px 0px {0.5\hant},clip, width=\linewidth]{images/caltech/ant.png}}%
		&{\includegraphics[trim={0.25\want} 0px {0.5\want} {0.5\hant},clip, width=\linewidth]{images/caltech/ant.png}}%
		&{\includegraphics[trim={0.5\want} {0.5\hant} {0.25\want} 0px,clip, width=\linewidth]{images/caltech/ant.png}}%
		&{\includegraphics[trim={0.75\want} {0.5\hant} 0px 0px,clip, width=\linewidth]{images/caltech/ant.png}}%
		&{\includegraphics[trim=0px {0.5\hant} {0.75\want} 0px,clip, width=\linewidth]{images/caltech/ant.png}}%
		\\
		
		\rotatebox{90}{\makecell{barrel}}
		&{\includegraphics[trim=0px 0px {0.75\wbarrel} {0.5\hbarrel},clip, width=\linewidth]{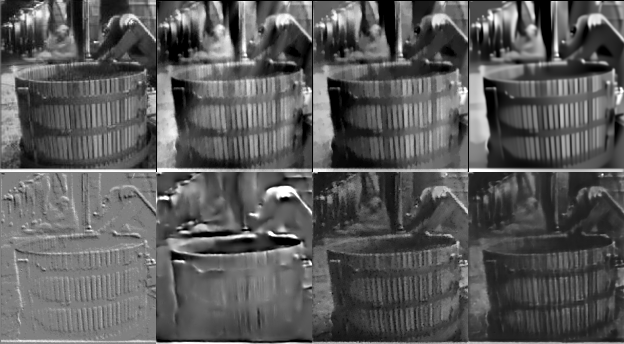}}%
		&{\includegraphics[trim={0.5\wbarrel} 0px {0.25\wbarrel} {0.5\hbarrel},clip, width=\linewidth]{images/caltech/barrel.png}}%
		&{\includegraphics[trim={0.75\wbarrel} 0px 0px {0.5\hbarrel},clip, width=\linewidth]{images/caltech/barrel.png}}%
		&{\includegraphics[trim={0.25\wbarrel} 0px {0.5\wbarrel} {0.5\hbarrel},clip, width=\linewidth]{images/caltech/barrel.png}}%
		&{\includegraphics[trim={0.5\wbarrel} {0.5\hbarrel} {0.25\wbarrel} 0px,clip, width=\linewidth]{images/caltech/barrel.png}}%
		&{\includegraphics[trim={0.75\wbarrel} {0.5\hbarrel} 0px 0px,clip, width=\linewidth]{images/caltech/barrel.png}}%
		&{\includegraphics[trim=0px {0.5\hbarrel} {0.75\wbarrel} 0px,clip, width=\linewidth]{images/caltech/barrel.png}}%
		\\
			
		\rotatebox{90}{\makecell{buddha}}
		&{\includegraphics[trim=0px 0px {0.75\wbuddha} {0.5\hbuddha},clip, width=\linewidth]{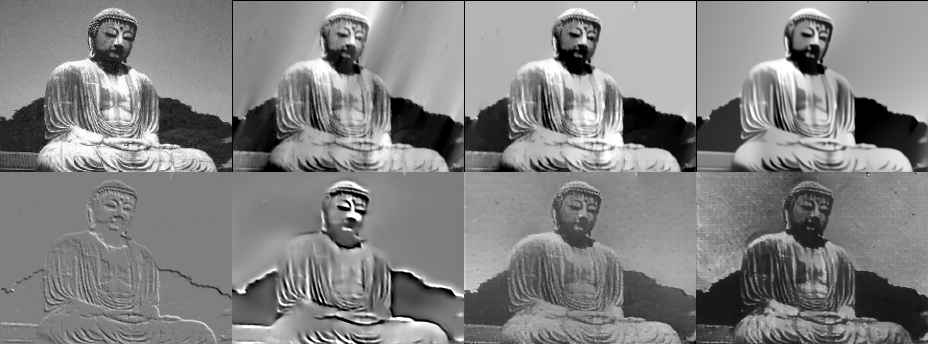}}%
		&{\includegraphics[trim={0.5\wbuddha} 0px {0.25\wbuddha} {0.5\hbuddha},clip, width=\linewidth]{images/caltech/buddha.png}}%
		&{\includegraphics[trim={0.75\wbuddha} 0px 0px {0.5\hbuddha},clip, width=\linewidth]{images/caltech/buddha.png}}%
		&{\includegraphics[trim={0.25\wbuddha} 0px {0.5\wbuddha} {0.5\hbuddha},clip, width=\linewidth]{images/caltech/buddha.png}}%
		&{\includegraphics[trim={0.5\wbuddha} {0.5\hbuddha} {0.25\wbuddha} 0px,clip, width=\linewidth]{images/caltech/buddha.png}}%
		&{\includegraphics[trim={0.75\wbuddha} {0.5\hbuddha} 0px 0px,clip, width=\linewidth]{images/caltech/buddha.png}}%
		&{\includegraphics[trim=0px {0.5\hbuddha} {0.75\wbuddha} 0px,clip, width=\linewidth]{images/caltech/buddha.png}}%
		\\

		\rotatebox{90}{\makecell{camera}}
		&{\includegraphics[trim=0px 0px {0.75\wcamera} {0.5\hcamera},clip, width=\linewidth]{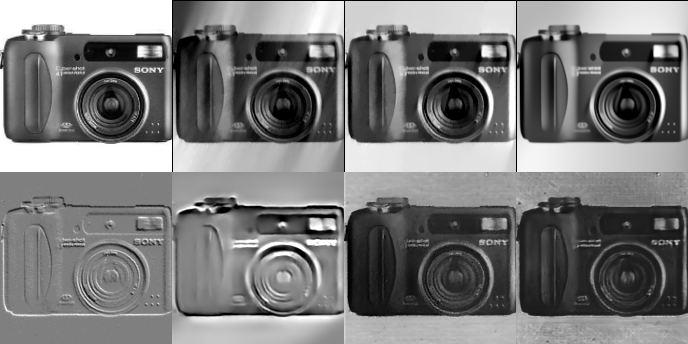}}%
		&{\includegraphics[trim={0.5\wcamera} 0px {0.25\wcamera} {0.5\hcamera},clip, width=\linewidth]{images/caltech/camera.png}}%
		&{\includegraphics[trim={0.75\wcamera} 0px 0px {0.5\hcamera},clip, width=\linewidth]{images/caltech/camera.png}}%
		&{\includegraphics[trim={0.25\wcamera} 0px {0.5\wcamera} {0.5\hcamera},clip, width=\linewidth]{images/caltech/camera.png}}%
		&{\includegraphics[trim={0.5\wcamera} {0.5\hcamera} {0.25\wcamera} 0px,clip, width=\linewidth]{images/caltech/camera.png}}%
		&{\includegraphics[trim={0.75\wcamera} {0.5\hcamera} 0px 0px,clip, width=\linewidth]{images/caltech/camera.png}}%
		&{\includegraphics[trim=0px {0.5\hcamera} {0.75\wcamera} 0px,clip, width=\linewidth]{images/caltech/camera.png}}%
		\\
		
		\rotatebox{90}{\makecell{cellphone}}
		&{\includegraphics[trim=0px 0px {0.75\wcellphone} {0.5\hcellphone},clip, width=\linewidth]{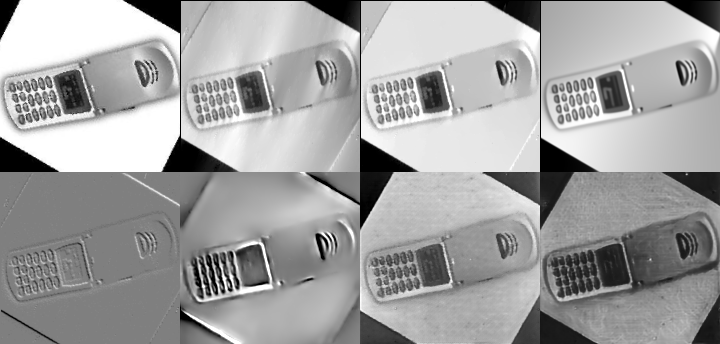}}%
		&{\includegraphics[trim={0.5\wcellphone} 0px {0.25\wcellphone} {0.5\hcellphone},clip, width=\linewidth]{images/caltech/cellphone.png}}%
		&{\includegraphics[trim={0.75\wcellphone} 0px 0px {0.5\hcellphone},clip, width=\linewidth]{images/caltech/cellphone.png}}%
		&{\includegraphics[trim={0.25\wcellphone} 0px {0.5\wcellphone} {0.5\hcellphone},clip, width=\linewidth]{images/caltech/cellphone.png}}%
		&{\includegraphics[trim={0.5\wcellphone} {0.5\hcellphone} {0.25\wcellphone} 0px,clip, width=\linewidth]{images/caltech/cellphone.png}}%
		&{\includegraphics[trim={0.75\wcellphone} {0.5\hcellphone} 0px 0px,clip, width=\linewidth]{images/caltech/cellphone.png}}%
		&{\includegraphics[trim=0px {0.5\hcellphone} {0.75\wcellphone} 0px,clip, width=\linewidth]{images/caltech/cellphone.png}}%
		\\

		\rotatebox{90}{\makecell{cup}}
		&{\includegraphics[trim=0px 0px {0.75\wcup} {0.5\hcup},clip, width=\linewidth]{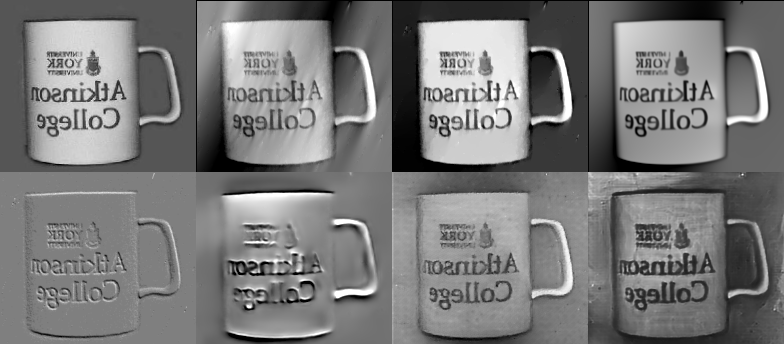}}%
		&{\includegraphics[trim={0.5\wcup} 0px {0.25\wcup} {0.5\hcup},clip, width=\linewidth]{images/caltech/cup.png}}%
		&{\includegraphics[trim={0.75\wcup} 0px 0px {0.5\hcup},clip, width=\linewidth]{images/caltech/cup.png}}%
		&{\includegraphics[trim={0.25\wcup} 0px {0.5\wcup} {0.5\hcup},clip, width=\linewidth]{images/caltech/cup.png}}%
		&{\includegraphics[trim={0.5\wcup} {0.5\hcup} {0.25\wcup} 0px,clip, width=\linewidth]{images/caltech/cup.png}}%
		&{\includegraphics[trim={0.75\wcup} {0.5\hcup} 0px 0px,clip, width=\linewidth]{images/caltech/cup.png}}%
		&{\includegraphics[trim=0px {0.5\hcup} {0.75\wcup} 0px,clip, width=\linewidth]{images/caltech/cup.png}}%
		\\
				
		& (a) NIWE
		& (b) E2VID~\cite{Rebecq19pami}
		& (c) ECNN~\cite{Stoffregen20eccv}
		& (d) BTEB~\cite{Paredes21cvpr}
		& (e) Ours (TV)
		& (f) Ours (CNN)
		& (g) Ground truth
	\end{tabular}
	}
    \fi
	\caption{Qualitative comparison with the state of the art on sequences from~\cite{Orchard15fns}. 
	Histogram equalization is not used. %
	}
	\label{fig:compare:caltech}
\end{figure*}

Quantitative results using the above metrics for six different methods are provided in \cref{tab:imgrec:ijrr:equalized}.
The choice of evaluation metric and protocol highly influence the numbers. 
For example, our image prior method produces the best results in terms of SSIM, whereas E2VID and ECNN are better in other metrics.
Despite not using ground truth images and not having recurrent connections to past events beyond the IWE, 
our three methods (Tikh., TV and CNN) produce results in line with the state of the art in reconstruction quality.
\Cref{fig:violin:eq} plots the distribution of MSE, SSIM and LPIPS values for the six methods in \cref{tab:imgrec:ijrr:equalized}.
These violin plots are more informative than the median numbers in \cref{tab:imgrec:ijrr:equalized}.
Surprisingly, solutions as simple as classical Tikhonov or TV regularizers acting on a linear system of equations produce already similar quality results as complex learning-based approaches.
In scenes like dynamic and shapes, the proposed CNN regularizer provides better SSIM and LPIPS distributions than ECNN.
The distributions of the proposed regularizers (Tikh.-TV-CNN) follow monotonic trends in MSE and SSIM, but the trend is not clear in terms of LPIPS. %

\subsubsection*{Analysis without Histogram Equalization}
Histogram equalization makes the results of the comparison metrics more invariant to changes in average image intensity.
For example, the SSIM metric comprises two parts: one that depends on the means and one that depends on the variances.
Without histogram equalization, both terms contribute to the SSIM.
However, with histogram equalization, the effect of the term that depends on the means is greatly reduced, and therefore the term that depends on the variances dominates the metric. 

\Cref{tab:imgrec:ijrr:noneq} and \cref{fig:violin:noeq}
are analogous to \cref{tab:imgrec:ijrr:equalized} and \cref{fig:violin:eq}, but without histogram equalization.
The changes in average intensity are noticeable in the MSE metric. 
ECNN tends to produce dark images, like those of the DAVIS camera in \cite{Mueggler17ijrr}, which often only span a 7-bit range of intensities (from 0 to 127). 
Hence, without histogram equalization ECNN tends to outperform other methods on these sequences just by producing darker images. 
With histogram equalization, E2VID and ECNN are best in terms of MSE (\cref{tab:imgrec:ijrr:equalized}), but without it we see that ECNN is the top one and our image prior approach (CNN) is consistently the second best (\cref{tab:imgrec:ijrr:noneq}).

\def\figWidth{0.20\linewidth}
\begin{figure*}[t]
    \ifhideimages
    \else
	\centering
    {\small
    \setlength{\tabcolsep}{2pt}
	\begin{tabular}{
	>{\centering\arraybackslash}m{0.3cm}
	>{\centering\arraybackslash}m{\figWidth} 
	>{\centering\arraybackslash}m{\figWidth}
	>{\centering\arraybackslash}m{\figWidth}
	>{\centering\arraybackslash}m{\figWidth}}

		\rotatebox{90}{\makecell{boxes}}
		&\frame{\includegraphics[width=\linewidth]{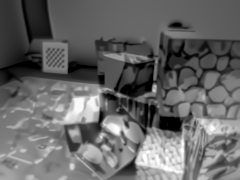}}
		&\frame{\includegraphics[width=\linewidth]{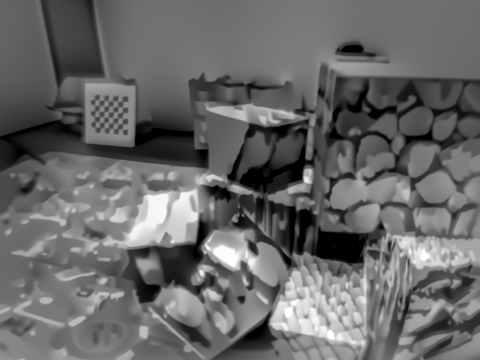}}
		&\frame{\includegraphics[width=\linewidth]{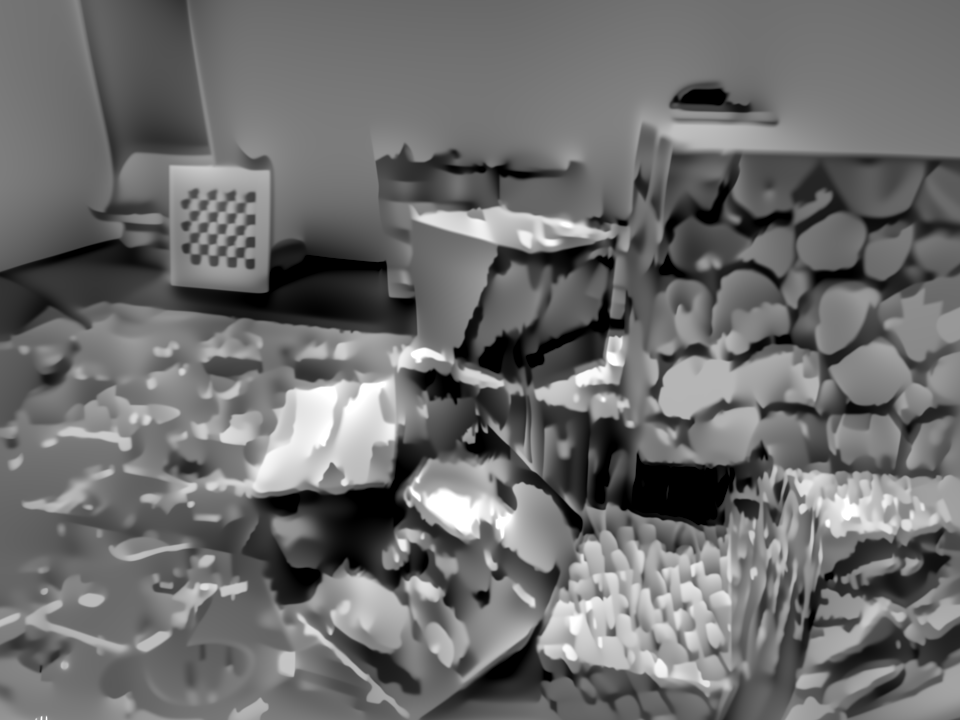}}
		&\frame{\includegraphics[width=\linewidth]{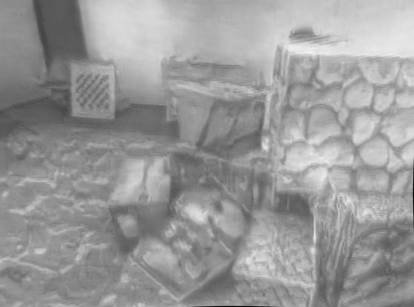}}
		\\
		
		\rotatebox{90}{\makecell{boxes-zoom}}
		&\frame{\includegraphics[width=\linewidth]{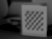}}
		&\frame{\includegraphics[width=\linewidth]{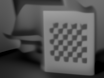}}
		&\frame{\includegraphics[width=\linewidth]{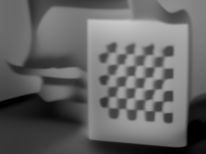}}
		&\frame{\includegraphics[width=\linewidth]{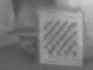}}
		\\
		
		\rotatebox{90}{\makecell{plane}}
		&\frame{\includegraphics[width=\linewidth]{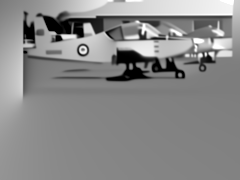}}
		&\frame{\includegraphics[width=\linewidth]{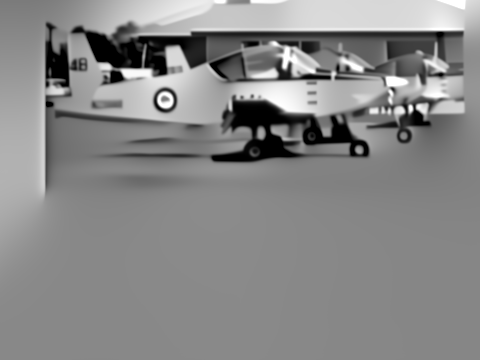}}
		&\frame{\includegraphics[width=\linewidth]{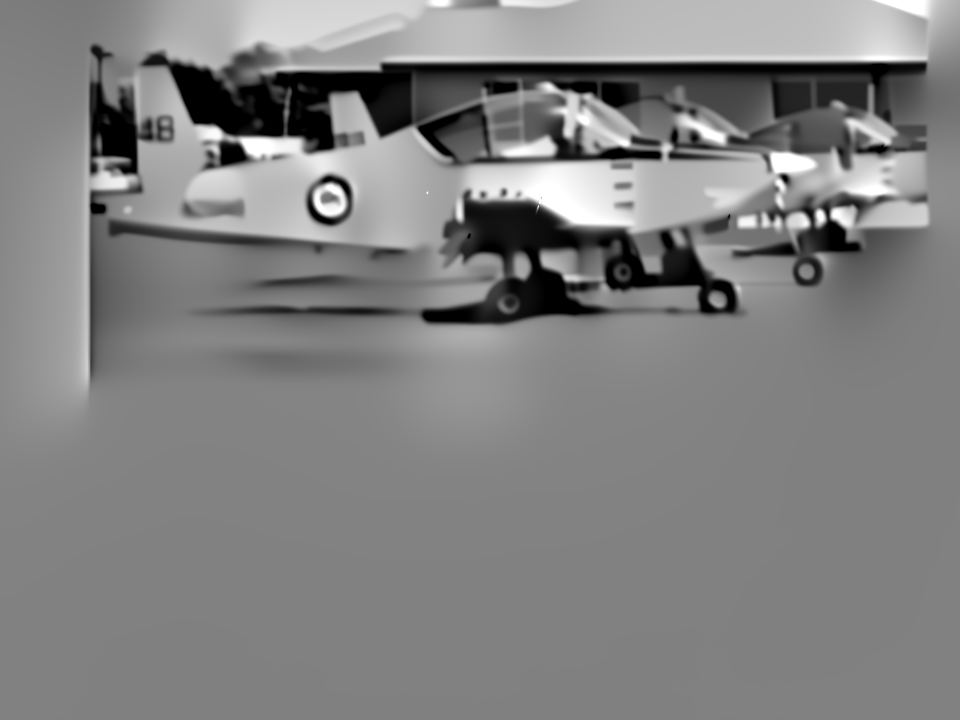}}
		&\frame{\includegraphics[width=\linewidth]{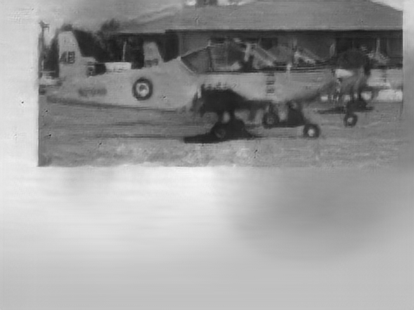}}
		\\
		
		\rotatebox{90}{\makecell{plane-zoom}}
		&\frame{\includegraphics[width=\linewidth]{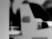}}
		&\frame{\includegraphics[width=\linewidth]{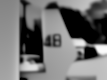}}
		&\frame{\includegraphics[width=\linewidth]{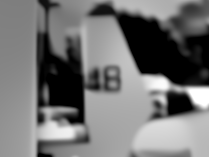}}
		&\frame{\includegraphics[width=\linewidth]{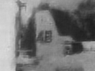}}
		\\
		
		& (a) Ours (CNN) $1\times$
		& (b) Ours (CNN) $2\times$
		& (c) Ours (CNN) $4\times$
		& (d) ESRI~\cite{Mostafavi21pami} $2\times$
	\end{tabular}
	}
	\fi
  \caption{\emph{Super-resolution}.
  Reconstructing and super-resolving brightness by factors of $2\times$ and $4\times$ in each spatial dimension.
  Details are best viewed by zooming in. %
  }
  \label{fig:sr}
\end{figure*}

Histogram equalization is a non-linear transformation unrelated to the event generation model \eqref{eq:generativeEventCondition}. 
Hence, it is reasonable that our methods (Tikhonov, TV and CNN image prior), which aim at solving a system of equations based on \eqref{eq:generativeEventCondition}, produce better values on metrics without image equalization than on metrics with it. 
The established evaluation protocol~\cite{Rebecq19pami,Stoffregen20eccv,Paredes21cvpr} is somehow contradictory: 
estimators are designed to solve some equations (i.e., based on \eqref{eq:generativeEventCondition}) and then their goodness of fit is measured using a different set of equations, which are non-linearly related.

In summary, the fact that the quantitative results highly depend on the metric, evaluation protocol (e.g., equalized vs.~not equalized) and scene texture calls for prudence when interpreting the numbers as the main means to finely rank the performance of image reconstructions methods.

\subsubsection{Results on the N-Caltech 101 Dataset}

\Cref{fig:compare:caltech} reports results on the N-Caltech 101 dataset \cite{Orchard15fns}.
It is complemented by Figs.~\ref{fig:compare:caltech:2}--\ref{fig:compare:caltech:3} and Tabs.~\ref{tab:imgrec:caltech:eq}--\ref{tab:imgrec:caltech:noneq} in the supplementary material.
We aligned the images with respect to the ground truth using Enhanced Correlation Coefficient maximization \cite{Evangelidis08pami} to be able to report quantitative results on the usual metrics (MSE, SSIM, LPIPS).
This dataset comprises sequences of 300 ms, with three saccades of 100 ms each. 
In the tests, we leave the first saccade as initialization for RNN baseline methods \cite{Rebecq19pami,Stoffregen20eccv,Paredes21cvpr}, 
and compare the results in the middle of the second saccade.
In our method, we only use 60 ms of data from the middle saccade. %

Quantitatively the total variation (TV) regularizer produces very good results in all three metrics using unequalized images (\cref{tab:imgrec:caltech:noneq}), with the CNN image prior being a competitive option in terms of MSE and SSIM, and E2VID in terms of LPIPS.
The performance of E2VID and ECNN improves if equalized images are used in the evaluation protocol. 
Qualitatively, we observe in Figs.~\ref{fig:compare:caltech}, \ref{fig:compare:caltech:2} and \ref{fig:compare:caltech:3}
compelling high-quality reconstructed images by our TV and image-prior methods compared to the state of the art.

\subsection{Super-resolution}
\label{sec:experim:sr}
Super-resolution is achieved by converting part of the high temporal resolution of the event camera into spatial resolution.
\Cref{fig:sr} shows results on simultaneous brightness reconstruction and super-resolution (\cref{sec:method:superresolution}) on sample sequences from \cite{Mueggler17ijrr,Orchard15fns}.
Input resolutions ($1\times$) are $240\times 180$ pixels for \cite{Mueggler17ijrr} (DAVIS camera) 
and $304\times 240$ pixels for \cite{Orchard15fns} (ATIS camera).
The super-resolved images have $2\times$ and $4\times$ more pixels in each dimension 
(i.e., $\approx$ 1Mpixel).
To build the IWE at resolutions $2\times, 4\times$ we slightly increased the number of events (less than doubled). 
There is a trade-off: using more events improves the SNR and fills in the super-resolved pixels of the IWE, 
but it increases the time spanned, and motion compensation may not be effective assuming constant optical flow \cite{Gu21iccv}.
In the images, we observe that increasing from $1\times$ to $2\times$ makes a big difference: 
at $2\times$ one can read the number ``48'' on the fin of the plane.
In the boxes scene, the checkerboard and the texture of the boxes become sharper at $2\times$.
From $2\times$ to $4\times$ the improvement is not as striking: 
the checkerboard and the contours of the folded foam mattress behind it become sharper, but we cannot read the text below the plane's fin. 
This may be difficult given the input spatial resolution. 

\begin{table}[t]
\centering
\caption{%
Image reconstruction from the Laplacian, using \cite{Duwek21cvprw} vs.~our method~\eqref{eq:recLaplacian}.
}
\label{tab:imgrec:lap}
\begin{adjustbox}{max width=\linewidth}
\setlength{\tabcolsep}{4pt}
\begin{tabular}{@{}l *{6}{S[table-format=1.4]}@{}}
             & \multicolumn{2}{c}{MSE $\downarrow$} & \multicolumn{2}{c}{SSIM $\uparrow$} & \multicolumn{2}{c}{LPIPS $\downarrow$}\\
             \cmidrule(l{2mm}r{2mm}){2-3} \cmidrule(l{2mm}r{2mm}){4-5} \cmidrule(l{2mm}r{2mm}){6-7} %
Sequence name   & \cite{Duwek21cvprw} & Ours
                & \cite{Duwek21cvprw} & Ours
                & \cite{Duwek21cvprw} & Ours\\
\midrule
brain & \bnum{0.0084} & 0.0102 & 0.7130 & \bnum{0.7394} & 0.2268 & \bnum{0.2105} \\
brain + noise & 0.0135 & \bnum{0.0130} & 0.6636 & \bnum{0.6921} & 0.3572 & \bnum{0.2758} \\
car & 0.0282 & \bnum{0.0235} & 0.6728 & \bnum{0.6938} & \bnum{0.3601} & 0.3836 \\
car + noise & 0.0368 & \bnum{0.0335} & 0.6193 & \bnum{0.6331} & \bnum{0.3797} & 0.4304 \\
chair & \bnum{0.0571} & 0.0638 & \bnum{0.8307} & 0.8246 & \bnum{0.2716} & 0.2956 \\
chair + noise & \bnum{0.0359} & 0.0433 & 0.7290 & \bnum{0.7221} & 0.3925 & \bnum{0.3589} \\
\bottomrule
\end{tabular}
\end{adjustbox}
\end{table}
\def\figWidth{0.3\linewidth}
\begin{figure}[t]
	\centering
    {\small
    \setlength{\tabcolsep}{2pt}
	\begin{tabular}{
	>{\centering\arraybackslash}m{0.3cm}
	>{\centering\arraybackslash}m{\figWidth}
	>{\centering\arraybackslash}m{\figWidth}
	>{\centering\arraybackslash}m{\figWidth}}

		\rotatebox{90}{\makecell{brain + noise}}
		&{\includegraphics[width=\linewidth]{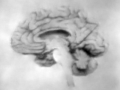}}
		&{\includegraphics[width=\linewidth]{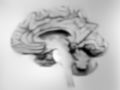}}
		&{\includegraphics[width=\linewidth]{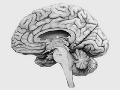}}
		\\

		\rotatebox{90}{\makecell{car + noise}}
		&\frame{\includegraphics[width=\linewidth]{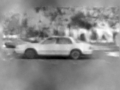}}
		&\frame{\includegraphics[width=\linewidth]{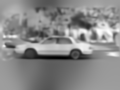}}
		&\frame{\includegraphics[width=\linewidth]{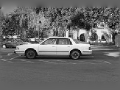}}
		\\

		\rotatebox{90}{\makecell{chair + noise}}
		&\frame{\includegraphics[width=\linewidth]{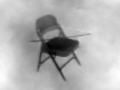}}
		&\frame{\includegraphics[width=\linewidth]{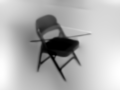}}
		&\frame{\includegraphics[width=\linewidth]{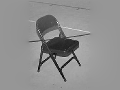}}
		\\

		& (a) \cite{Duwek21cvprw}
		& (b) Ours~\eqref{eq:recLaplacian}
		& (c) Ground truth
	\end{tabular}
	}
	\caption{\emph{Reconstruction using the Laplacian}.
    Results using the method of \cref{sec:method:LaplacianRec} with the CNN regularizer.
	}
	\label{fig:lap}
\end{figure}

\def\figWidth{0.2\linewidth}
\begin{figure*}[t]
    \ifhideimages
    \else
	\centering
    {\small
    \setlength{\tabcolsep}{2pt}
	\begin{tabular}{
	>{\centering\arraybackslash}m{0.3cm}
	>{\centering\arraybackslash}m{\figWidth} 
	>{\centering\arraybackslash}m{\figWidth}
	>{\centering\arraybackslash}m{\figWidth}
	>{\centering\arraybackslash}m{\figWidth}
	>{\centering\arraybackslash}m{\figWidth}}

		\rotatebox{90}{\makecell{$\bell_k$}}
		&
		\ifshowanimationcommentarxiv
		\animategraphics[width=\linewidth]{8}{images/evolution/boxes_xs/00}{00}{15}
		\else
		{\includegraphics[width=\linewidth]{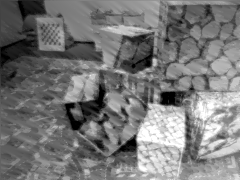}}
		\fi
		&{\includegraphics[width=\linewidth]{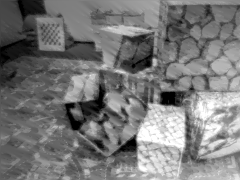}}
		&{\includegraphics[width=\linewidth]{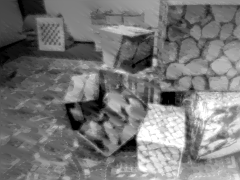}}
		&{\includegraphics[width=\linewidth]{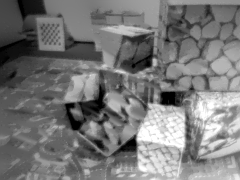}}
		\\

		\rotatebox{90}{\makecell{$\bz_k$}}
		&
		\ifshowanimationcommentarxiv
		\animategraphics[width=\linewidth]{8}{images/evolution/boxes_zs/00}{00}{15}
		\else
		{\includegraphics[width=\linewidth]{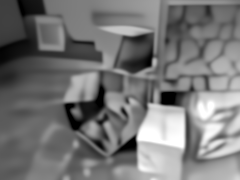}}
		\fi
		&{\includegraphics[width=\linewidth]{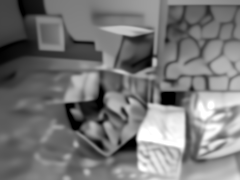}}
		&{\includegraphics[width=\linewidth]{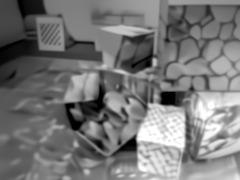}}
		&{\includegraphics[width=\linewidth]{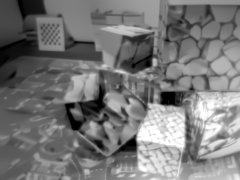}}
		\\\addlinespace[1ex]
		
		\rotatebox{90}{\makecell{$\bell_k$}}
		&
		\ifshowanimationcommentarxiv
		\animategraphics[width=\linewidth]{8}{images/evolution/dynamic_xs/00}{00}{15}
		\else
		{\includegraphics[width=\linewidth]{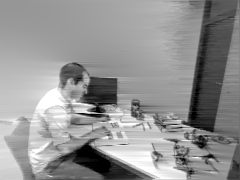}}
		\fi
		&{\includegraphics[width=\linewidth]{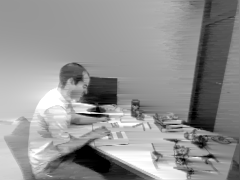}}
		&{\includegraphics[width=\linewidth]{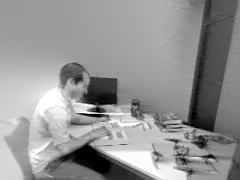}}
		&{\includegraphics[width=\linewidth]{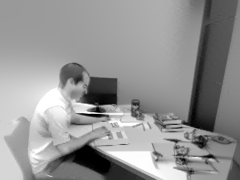}}
		\\

		\rotatebox{90}{\makecell{$\bz_k$}}
		&
		\ifshowanimationcommentarxiv
		\animategraphics[width=\linewidth]{8}{images/evolution/dynamic_zs/00}{00}{15}
		\else
		{\includegraphics[width=\linewidth]{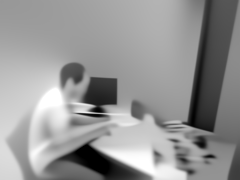}}
		\fi
		&{\includegraphics[width=\linewidth]{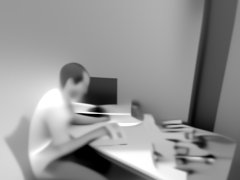}}
		&{\includegraphics[width=\linewidth]{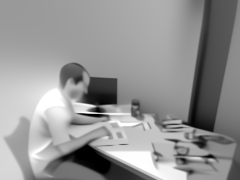}}
		&{\includegraphics[width=\linewidth]{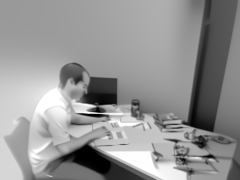}}
		\\
		
		& (a) iter $k=0$
		& (b) $k=3$
		& (c) $k=7$
		& (d) $k=15$
	\end{tabular}
	}
	\fi
	\caption{Evolution of the solution ($\bell_k$,$\bz_k$) in the alternating, iterative solver \eqref{eq:iterative} with CNN regularizer.
	We start $\bell_0$ from the solution provided by the total variation (TV).
	As iterations proceed, $\bz_k$ is refined, capturing more fine details, becoming less smooth.
	\ifshowanimationcommentarxiv
	\emph{The first column contains animations if opened (and clicked on) with Acrobat Reader, with the iteration counter in red}.
	\fi
	Event data from \cite{Mueggler17ijrr}.
	}
	\label{fig:evol:boxes}
\end{figure*}

Additionally, we compare with state-of-the-art method ESRI \cite{Mostafavi21pami} on the same data. 
The results for the best ESRI $2\times$ profile (7 bins) are shown in \cref{fig:sr}.
Our method produces overall better-looking images and can handle various super-resolution scales with minor modifications, while ESRI requires multiple ANNs to handle different scales.

\subsection{Image Reconstruction via the Laplacian}
\label{sec:experim:laplacian}
Here we demonstrate the method of \cref{sec:method:LaplacianRec}.
First we use the ANN in \cite{Duwek21cvprw} to predict the Laplacian image $\veclap$ from the events.
Subsequently this image is passed to our solver~\eqref{eq:recLaplacian}. %
\Cref{tab:imgrec:lap} and \cref{fig:lap} report results, where we also add Gaussian noise to the Laplacian. 
The Laplacian is given in the range $[-0.6,0.6]$, and the noise has standard deviation $\sigma = 0.02$. 
Given the limited spatial input resolution of $\veclap$ provided by the ANN ($120\times 90$ pixels), 
the differences between \cite{Duwek21cvprw} and our method are most noticeable in the way they deal with noise.
Our method yields higher contrast, sharper edges and smoother homogeneous regions due to the CNN denoiser.
Regarding \cref{tab:imgrec:lap}, the differences are not as large and depend on the metric.
Nevertheless, our method is clearly better in terms of SSIM.

\subsection{Analysis}
\label{sec:experim:analysis}
\subsubsection{Evolution of the Solution with the Iterations}
\label{sec:experim:evol}
\def\plotWidth{0.2213\linewidth} \def\figWidth{0.2\linewidth}
\begin{figure*}[t]
    \ifhideimages
    \else
	\centering
    {\small
    \setlength{\tabcolsep}{2pt}
	\begin{tabular}{
	>{\centering\arraybackslash}m{\plotWidth}
	>{\centering\arraybackslash}m{\figWidth} 
	>{\centering\arraybackslash}m{\figWidth}
	>{\centering\arraybackslash}m{\figWidth}}

        \includegraphics[trim=15.0cm 0.0cm 2.5cm 1.0cm,clip,width=\linewidth]{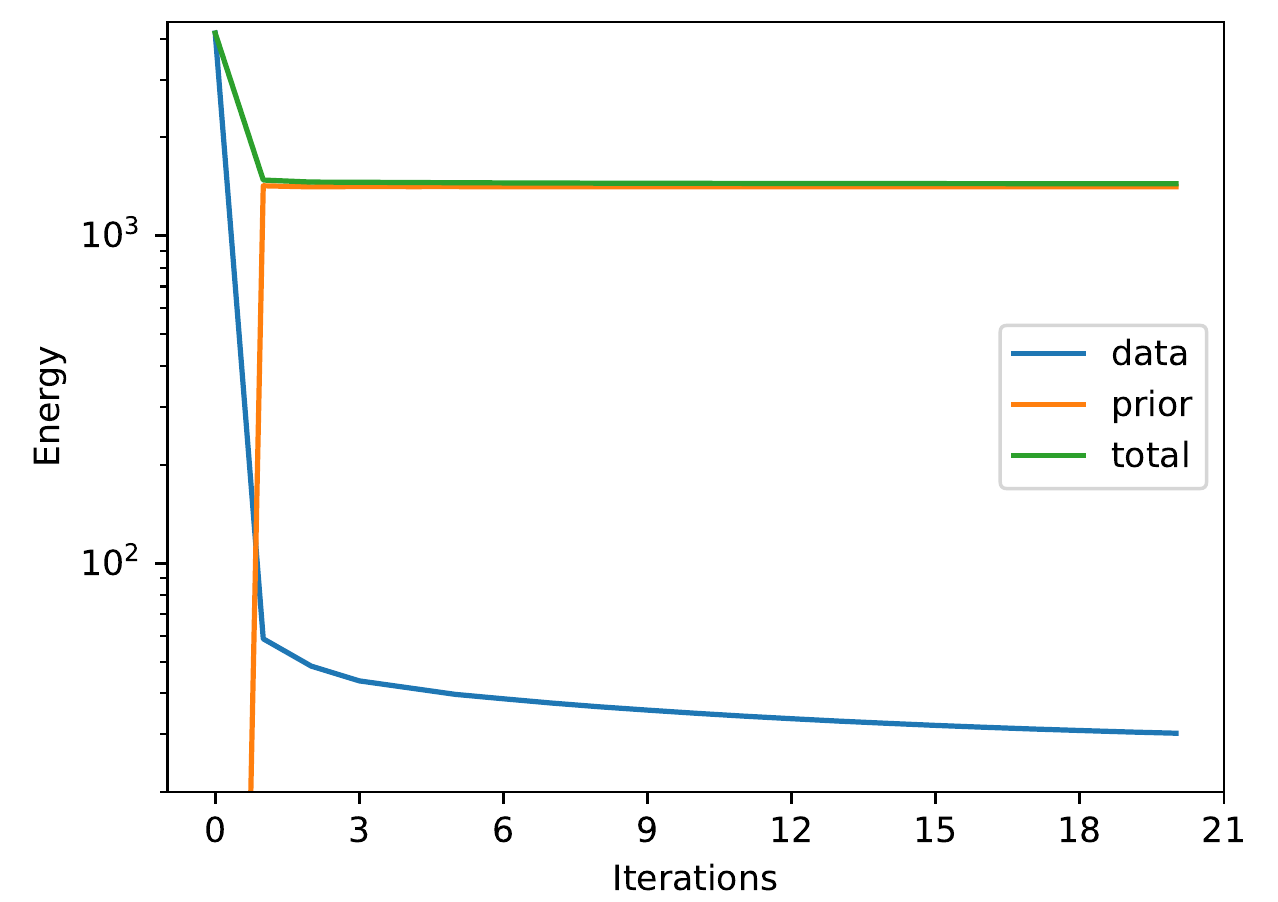}
		&
        \ifshowanimationcommentarxiv
        \animategraphics[width=\linewidth]{8}{images/evolution/l1/0000000}{00}{19}
        \else
        {\includegraphics[width=\linewidth]{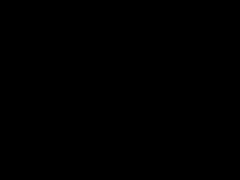}}
        \fi
		&{\includegraphics[width=\linewidth]{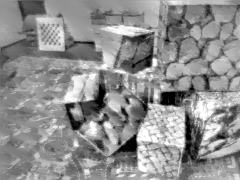}}
        &{\includegraphics[width=\linewidth]{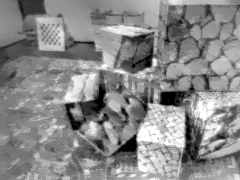}}
		\\
		
		(a) Cost evolution
		& (b) iter $k=0$
		& (c) $k=1$
		& (d) $k=2$
		\\\addlinespace[1ex]

		&{\includegraphics[width=\linewidth]{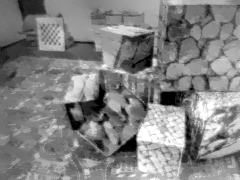}}
		&{\includegraphics[width=\linewidth]{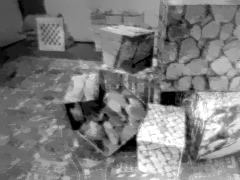}}
        &{\includegraphics[width=\linewidth]{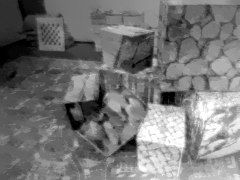}}
		\\

		& (e) $k=4$
		& (f) $k=8$
		& (g) $k=19$
	\end{tabular}
	}
	\fi
 	\caption{Evolution of the objective value and brightness image during reconstruction with total variation (TV) regularization \eqref{eq:map:minproblem}. 
 	(a) %
 	The value of the prior term (i.e., regularizer) is $\lambda \|\nabla \bell\|$. 
 	The value of the data term drops significantly in the first iteration and the structure of the scene shows up (c). 
 	Then, image details are refined in subsequent iterations (d)-(g). 
 	The prior term encourages image smoothness, hence preventing the solution from completely fitting the data term.
 	\ifshowanimationcommentarxiv
 	\emph{The first image (b) contains an animation of the evolution. Best seen in Acrobat Reader (possibly clicking on image (b)).}
 	\fi
 	}
	\label{fig:energy_evo:l1}
\end{figure*}

\def\plotWidth{0.2213\linewidth} \def\figWidth{0.2\linewidth}
\begin{figure*}[t]
    \ifhideimages
    \else
	\centering
    {\small
    \setlength{\tabcolsep}{2pt}
	\begin{tabular}{
	>{\centering\arraybackslash}m{\plotWidth}
	>{\centering\arraybackslash}m{\figWidth} 
	>{\centering\arraybackslash}m{\figWidth}
	>{\centering\arraybackslash}m{\figWidth}}

		\includegraphics[trim=15.0cm 0.0cm 2.5cm 1.0cm,clip,width=\linewidth]{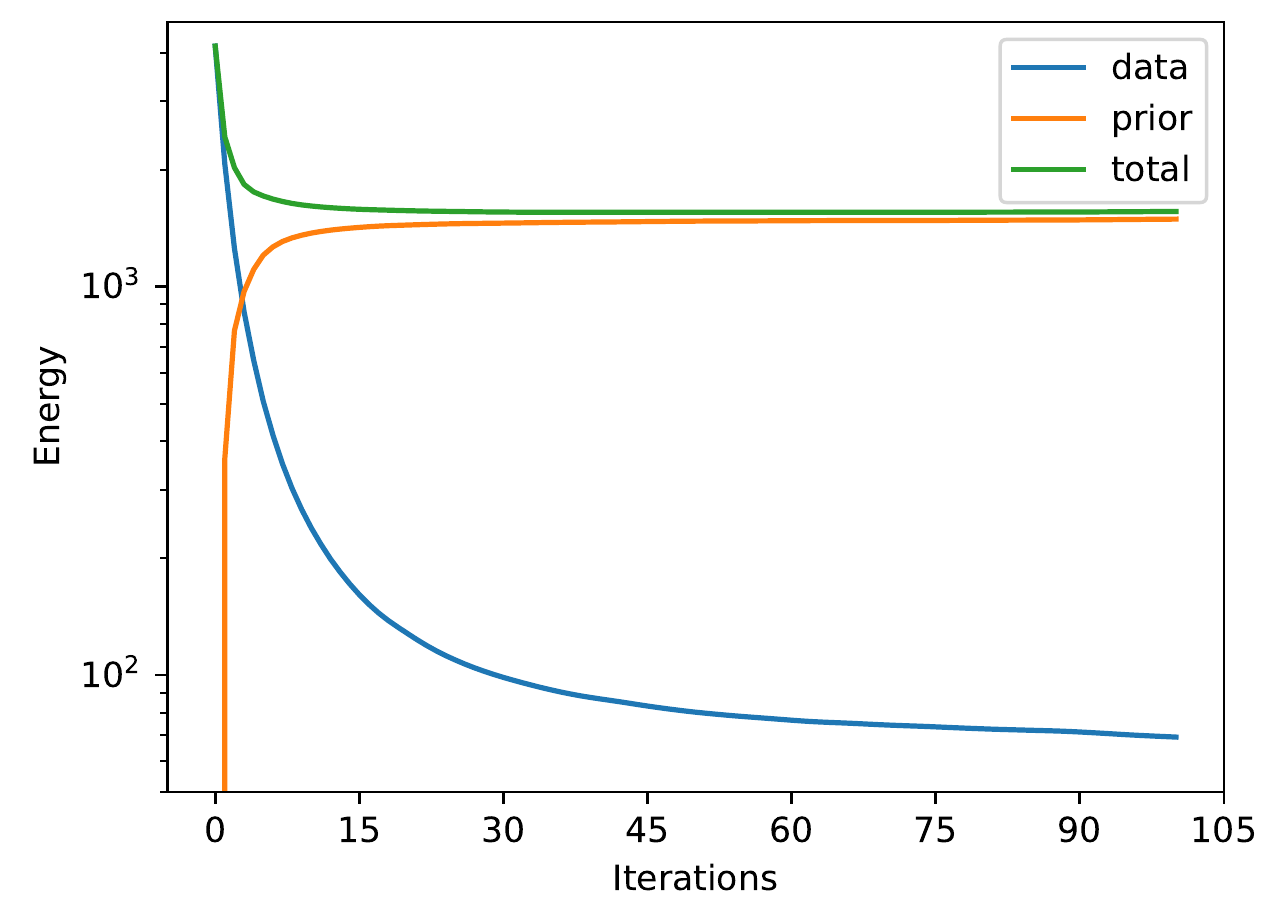}
		&
        \ifshowanimationcommentarxiv
        \animategraphics[width=\linewidth]{8}{images/evolution/l2/000000}{000}{100}
        \else 
        {\includegraphics[width=\linewidth]{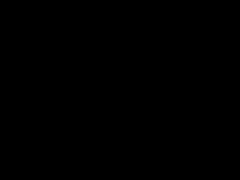}}
        \fi 
		&{\includegraphics[width=\linewidth]{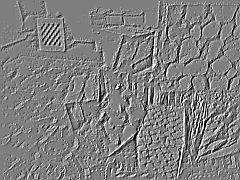}}
        &{\includegraphics[width=\linewidth]{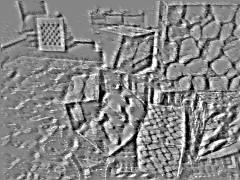}}
		\\
		
		(a) Cost evolution
		& (b) iter $k=0$
		& (c) $k=1$
		& (d) $k=4$
		\\\addlinespace[1ex]

		&{\includegraphics[width=\linewidth]{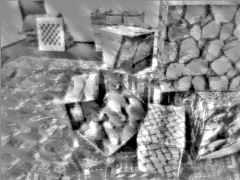}}
		&{\includegraphics[width=\linewidth]{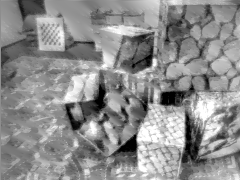}}
        &{\includegraphics[width=\linewidth]{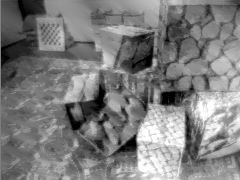}}
		\\

		& (e) $k=16$
		& (f) $k=48$
		& (g) $k=100$
	\end{tabular}
	}
	\fi
 	\caption{
 	Evolution of the objective value and brightness image during reconstruction with Tikhonov regularization (i.e., with prior term $\lambda \|\nabla \bell\|^2$). 
 	Similar comments as in \cref{fig:energy_evo:l1} apply.
 	Starting from zero brightness (b), the value of the data term drops significantly during the first iterations and the structure of the scene (mostly around edges) appears (c)-(d). 
 	Homogeneous regions are filled in subsequent iterations (e)-(g). 
 	\ifshowanimationcommentarxiv
 	\emph{The animation (b) is best seen in Acrobat Reader.}
 	\fi
 	}
	\label{fig:energy_evo:l2}
\end{figure*}

\Cref{fig:evol:boxes} shows the inner workings of the iterative solver \eqref{eq:iterative} on sample sequences.
We run the image-prior solver \eqref{eq:iterative} (CNN) for 16 iterations.
The figure depicts the evolution of the two variables $\bell_k, \bz_k$ as the iterations proceed. 
They start far apart, but converge to each other with the iterations.
The auxiliary variable $\bz_k$ starts as a very smooth and denoised version of the solution, and it achieves higher level of detail (less smoothing) with the iterations. 
The brightness image $\bell_k$ starts with very fine details and also artifacts in the direction of the optical flow (e.g., \cref{fig:compare:regularizers}a), and some of the details get lost along with the artifacts as it is denoised.
In the alternating process, the iterative solver \eqref{eq:iterative:a} enforces the data fidelity equations 
while the denoiser \eqref{eq:iterative:b} pulls the solution towards satisfying structural constraints (i.e., in the space of natural-looking images).

\Cref{fig:energy_evo:l1,fig:energy_evo:l2} show the evolution of cost function (energy) and brightness when TV and Tikhonov regularization are used to guide the solution of the linear system. 
The objective value is measured in squared units of the IWE \eqref{eq:IWEfromevents}.
We use event count as IWE units, thus discarding the contrast sensitivity $C$ (which can be regarded as an unknown scaling factor). 
The split Bregman method \cite{goldstein2009split} is used to solve the cost function in the case of the TV regularization, 
while a sparse least square solver \cite{paige1982lsqr,saunders1995solution} is used to solve the cost function with Tikhonov regularization. 
The optimization starts from zero brightness (b).
The algorithms converge in few iterations and the majority of scene structure comes into view. 
The regularizer (prior term) hinders the total energy from dropping sharply along with the data fidelity term and thus guides the solution. 
The refinement of brightness is less pronounced quantitatively as iterations proceed. 
The evolution in \cref{fig:energy_evo:l2} resembles a heat diffusion process; the integration effect, which decreases the cost of the data term, happens from the edges to the homogeneous regions.
The regularization factor $\lambda$ \eqref{eq:map:minproblem} is set to 0.1 for TV and 0.3 for Tikhonov to create the plots.

\subsubsection{Short-time Image Reconstruction for Monitoring}
\label{sec:experim:shorttime}
\def\figWidth{0.49\linewidth}
\begin{figure}[t]
  \centering
\begin{subfigure}{\figWidth}
    \frame{\includegraphics[width=\linewidth]{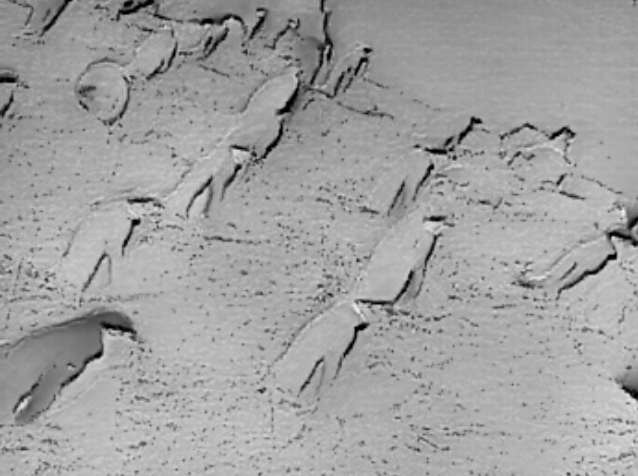}}
    \caption{E2VID \cite{Rebecq19pami}\label{fig:penguins:e2vid}}
\end{subfigure} 
\begin{subfigure}{\figWidth}
    \frame{\includegraphics[width=\linewidth]{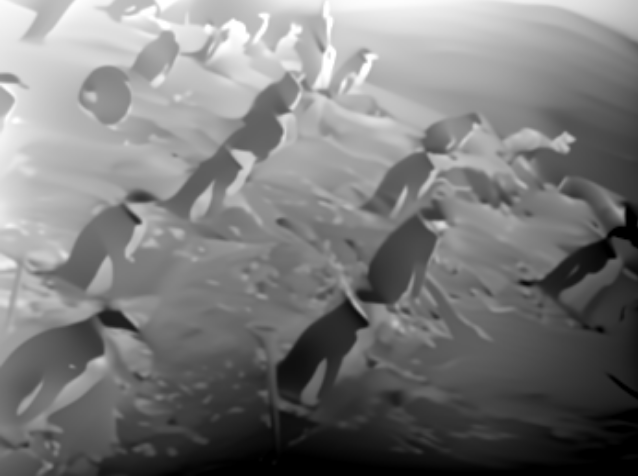}}
    \caption{Ours (CNN)\label{fig:penguins:ours}}
\end{subfigure} 
  \caption{Brightness reconstruction from a stationary event camera ($346 \times 260$ pixels) 
  observing penguins in Antarctica while it is moved by a wind gust.
  }
  \label{fig:penguins}
\end{figure}

There are scenarios where short-time reconstruction (i.e., image reconstruction, as opposed to video reconstruction) can be particularly useful, such as surveillance or monitoring. 
\Cref{fig:penguins} shows an example of animal behavior monitoring.
In this scenario, the event camera stays almost stationary and provides data bandwidth savings for long term and power-constrained monitoring. 
Small vibrations of the camera (unintended or triggered) are useful to reconstruct the brightness of the scene sparingly in time. 
This is a difficult scenario for RNN methods (\cref{fig:penguins:e2vid}), which require an initialization phase with continuous motion to build up the internal state and fill in homogeneous brightness regions, where no events are triggered. 
Our method (\cref{fig:penguins:ours}) is able to fill in homogeneous regions by decreasing the data-fidelity cost, and therefore produces good results.

\subsubsection{Effect of Varying the Regularizer Weight}
\label{sec:experim:analysis:regweight}
The main hyperparameters of the method are $N_e$ and the regularizer weight $\lambda$. 
$N_e$ is set depending on the scene texture, as mentioned in \cref{sec:experim:imgquality:datasets}.
The effect of varying the regularizer weight $\lambda$ is shown in \cref{fig:regweight}, 
which reports a trade-off between artifact removal, detail preservation and oversmoothing. 
A small $\lambda$ (\cref{fig:regweight:small}) preserves details but does not remove the artifacts in the flow streamlines that arise from the rank deficiency of the operator $D$.
A large $\lambda$ (\cref{fig:regweight:large}) produces an over-smooth reconstructed image, where only strong edges are reconstructed, albeit in a non-linear way in case of the CNN regularizer due to its anisotropic nature. 
An intermediate $\lambda$ (\cref{fig:regweight:middle}) provides a good compromise among the two above extreme cases.
\def\figWidth{0.325\linewidth}
\begin{figure}[h]
  \centering
\begin{subfigure}{\figWidth}
    \includegraphics[width=\linewidth]{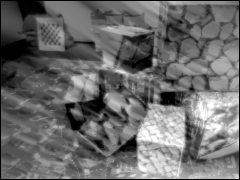} %
    \caption{$\lambda=.023$ (small)\label{fig:regweight:small}}
\end{subfigure} 
\begin{subfigure}{\figWidth}
    \includegraphics[width=\linewidth]{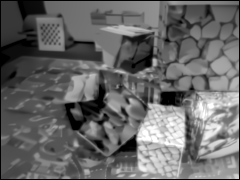} %
    \caption{$\lambda=.23$ (good)\label{fig:regweight:middle}}
\end{subfigure} 
\begin{subfigure}{\figWidth}
    \includegraphics[width=\linewidth]{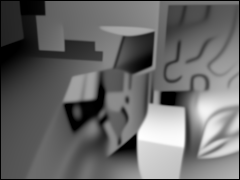} %
    \caption{$\lambda=2.3$ (large)\label{fig:regweight:large}}
\end{subfigure} 
  \caption{Effect of varying the CNN regularizer weight. 
  }
  \label{fig:regweight}
\end{figure}

\subsection{Segmentation and Color}
\label{sec:experim:extensions}

\subsubsection{Motion Segmentation and Image Reconstruction}
\label{sec:experim:extensions:motseg}
\Cref{fig:motsegm} provides qualitative results on the combination of our method and motion segmentation.
The scene consists of a moving car on a street while the camera is panning \cite{Stoffregen19iccv}.
\Cref{fig:motsegm}a shows the input events in space-time, 
which are clustered in two by running \cite{Zhou21tnnls} (\cref{fig:motsegm}b): 
events $\{\cE_1, \cE_2\}$ and their motion parameters $\{\btheta_1, \btheta_2\}$.
The majority of events ($\cE_1$) are due to the apparent motion of the static parts of the scene (e.g., buildings) as the camera moves.
The car, segmented in a separate cluster as an independent moving object with respect to the background, produces fewer events ($\cE_2$).
Our image reconstruction method can be applied to each cluster independently, i.e., to the IWE produced by each cluster (\cref{fig:motsegm}c).
As \cref{fig:motsegm}d shows, the method is able to recover the brightness of the foreground and the background separately, 
as if it had split a grayscale image into its independently moving components.
\def\figWidth{0.46\linewidth}
\begin{figure}[t]
    \ifhideimages
    \else
	\centering
    {\small
    \setlength{\tabcolsep}{2pt}
	\begin{tabular}{
	>{\centering\arraybackslash}m{0.3cm}
	>{\centering\arraybackslash}m{\figWidth} 
	>{\centering\arraybackslash}m{\figWidth}}

        \rotatebox{90}{\makecell{Segmentation}}
		&\includegraphics[width=\linewidth]{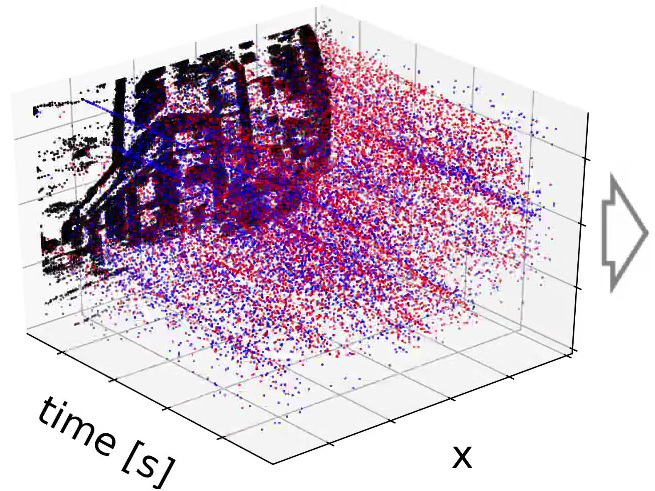}
        &\frame{\includegraphics[width=\linewidth]{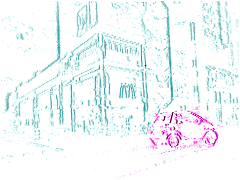}}
		\\
		
		& (a) Input events (space-time)
		& (b) Event clusters by \cite{Zhou21tnnls}
		\\\addlinespace[2ex]
		
        \rotatebox{90}{\makecell{Cluster 1 \;(BG)}}
		&{\includegraphics[width=\linewidth]{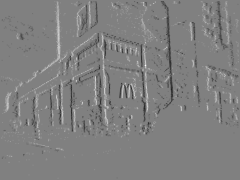}}
        &{\includegraphics[width=\linewidth]{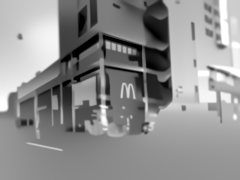}}
		\\

		\rotatebox{90}{\makecell{Cluster 2 \;(FG)}}
		&{\includegraphics[width=\linewidth]{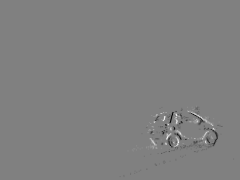}}
        &{\includegraphics[width=\linewidth]{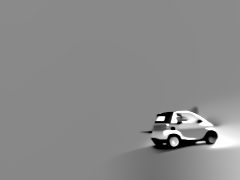}}
		\\
		
		& (c) IWEs (events)
		& (d) Ours (CNN)
	\end{tabular}
	}
	\fi
 	\caption{
  	Image reconstruction after event-based motion segmentation into foreground (FG) and background (BG).
 	}
	\label{fig:motsegm}
\end{figure}

\subsubsection{Color Image Reconstruction}
\label{sec:experim:extensions:color}
\Cref{fig:color} shows that our method can be used to reconstruct color images using data from a color event camera \cite{Li15iscas,Scheerlinck19cvprw}.
The reconstruction quality may not be as good as that of the recurrent neural network \cite{Rebecq19pami}, which is a high-capacity supervised learning method that exploits the history of all past events and handles color in a different color space (CIELAB color, by combining full and low-resolution reconstructions for the luminance and chroma channels).
Nevertheless, the reconstructed images exhibit HDR characteristics (e.g., one can see through the overexposed windows) and suffer from less spike color spots than the RNN method.
While color is not the target of this work, we think it is interesting to highlight the connection: color, super-resolution and reconstruction can be easily combined in our method. %
\def\figWidth{0.325\linewidth}
\begin{figure}[h]
  \centering
\begin{subfigure}{\figWidth}
    \frame{\includegraphics[width=\linewidth]{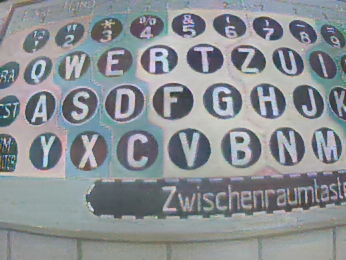}}
\end{subfigure} 
\begin{subfigure}{\figWidth}
    {\includegraphics[width=\linewidth]{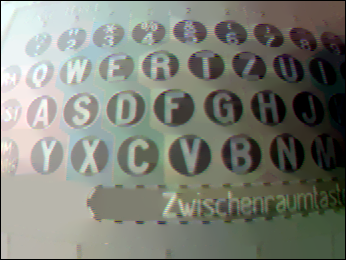}}
\end{subfigure} 
\begin{subfigure}{\figWidth}
    \frame{\includegraphics[width=\linewidth]{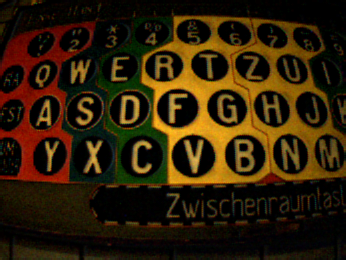}}
\end{subfigure}\\[0.8ex]
\begin{subfigure}{\figWidth}
    \frame{\includegraphics[width=\linewidth]{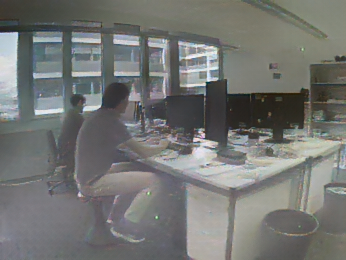}}
    \caption{E2VID \cite{Rebecq19pami}\label{fig:color:a}}
\end{subfigure} 
\begin{subfigure}{\figWidth}
    {\includegraphics[width=\linewidth]{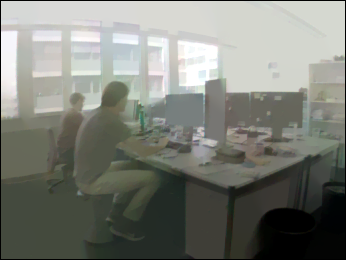}}
    \caption{Ours (CNN)\label{fig:color:b}}
\end{subfigure} 
\begin{subfigure}{\figWidth}
    \frame{\includegraphics[width=\linewidth]{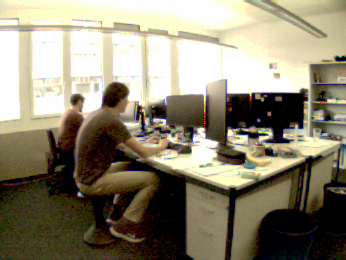}}
    \caption{DAVIS frame\label{fig:color:c}}
\end{subfigure}
  \caption{Color image reconstruction from events.}
  \label{fig:color}
\end{figure}

\subsection{Combination with State-of-the-art Dense Optical Flow Methods}
\label{sec:experim:denseflow}

\def\figWidth{0.46\linewidth}
\begin{figure}[t]
	
	\centering
    {\small
    \setlength{\tabcolsep}{2pt}
	\begin{tabular}{
	>{\centering\arraybackslash}m{0.3cm}
	>{\centering\arraybackslash}m{\figWidth} 
	>{\centering\arraybackslash}m{\figWidth}}
	
        \rotatebox{90}{\makecell{E-RAFT \;\cite{Gehrig21threedv}}}
        &{\includegraphics[width=\linewidth]{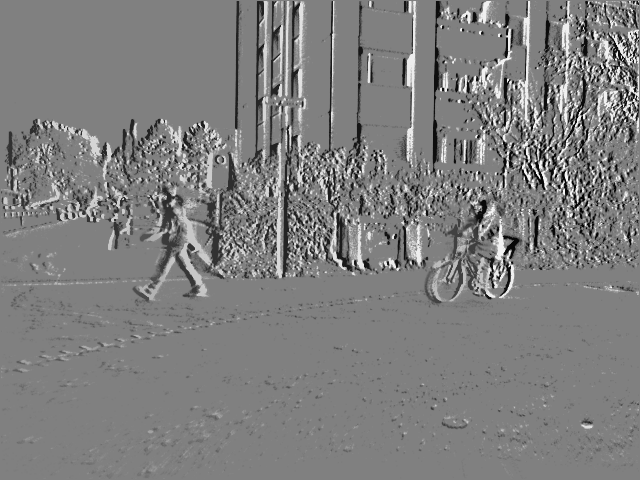}}
		&{\includegraphics[width=\linewidth]{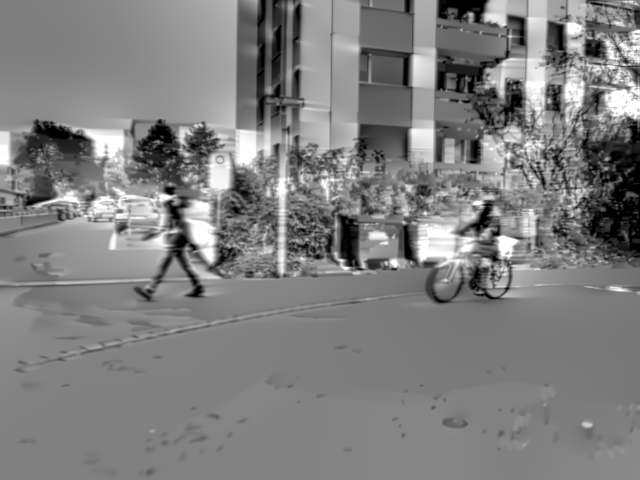}}
		\\

		\rotatebox{90}{\makecell{MCM \;\cite{Shiba22eccv}}}
		&{\includegraphics[width=\linewidth]{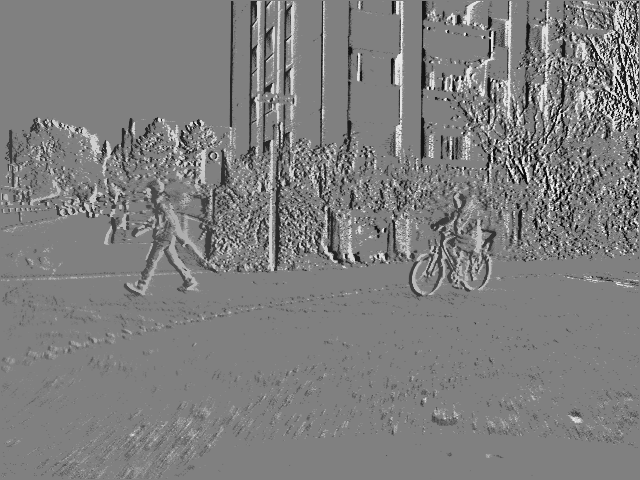}}
		&{\includegraphics[width=\linewidth]{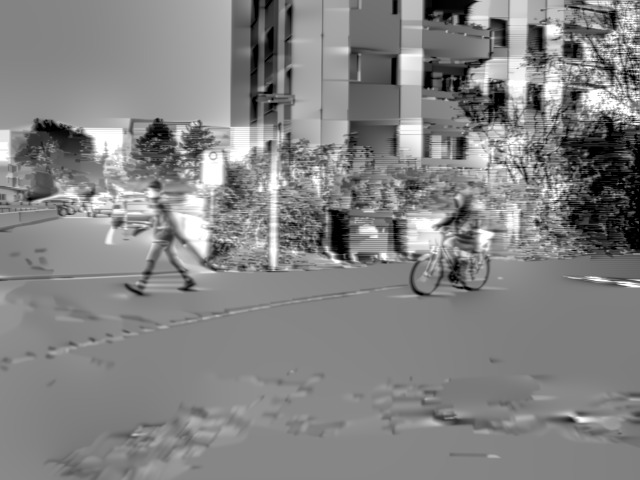}}
		\\
		
		& (a) IWE (events)
		& (b) Ours (CNN)
	\end{tabular}
	}
 	\caption{
  	Image reconstruction in combination with off-the-shelf dense optical flow methods \cite{Gehrig21threedv} and \cite{Shiba22eccv}.
  	The driving scene is from the DSEC dataset \cite{Gehrig21ral} ($640\times 480$ pixels).
 	}
	\label{fig:denseoflow}
\end{figure}

Our IWE method is based on the assumption of known optical flow, like~\cite{Paredes21cvpr}. 
This allowed us to focus on reformulating the image reconstruction part as a linear inverse problem and propose efficient, explainable solvers.
The experiments so far used the flow estimated by methods designed for low degrees-of-freedom (DOF) motions \cite{Gallego18cvpr,Nunes21pami,Peng21pami,Shiba22sensors}, which are often a good approximation 
and produce very accurate results if the true motion satisfies their assumptions~\cite{Gu21iccv,Ghosh22aisy}.
More complex scenes may require dense optical flow estimation, as given by \cite{Gehrig21threedv,Paredes21cvpr,Shiba22eccv} (i.e., higher DOFs).

\Cref{fig:denseoflow} shows the result of our IWE method when combined with two state-of-the-art event-based dense optical flow methods: E-RAFT \cite{Gehrig21threedv} (top row) and Multi-reference Contrast Maximization \cite{Shiba22eccv} (MCM, bottom row).
E-RAFT is a supervised deep learning method based on RAFT \cite{Teed20eccv}, and it uses the events in a time window of 100 ms.
MCM is a model-based method that uses a fixed number of events (500k in the example). 
The data comes from the DSEC dataset \cite{Gehrig21ral}, recorded with a VGA resolution Prophesee Gen3 CD event camera (which does not have grayscale output for comparison).
The results are qualitatively similar (both IWEs in \cref{fig:denseoflow} are reasonably sharp and the reconstructed images have many of the natural details of the city scene), showing that our method can be easily combined with modern dense optical flow estimation methods.
Looking into the differences, E-RAFT receives more events than MCM and is trained on DSEC data, so its optical flow produces an IWE with higher SNR in stationary parts of the scene, yielding slightly better reconstructions than the flow by MCM. 
However E-RAFT does not produce accurate flow for independently moving objects (IMOs) such as the pedestrian and the bicycle (since it is trained using the motion field induced by the vehicle), hence the reconstructed IMOs are more blurred than those by MCM.

In short, image reconstruction results are very good if the motion is accurately provided, as shown in low-DOF scenarios \cite{Gallego18cvpr,Nunes21pami,Peng21pami}, which allowed us to explore the possibilities of the framework (Secs.~\ref{sec:experim:imgquality}--\ref{sec:experim:extensions}).
The method delivers good results, i.e., degrades gracefully, if combined with modern dense optical flow methods (high-DOF scenarios, 
\cref{sec:experim:denseflow}), which have higher errors in optical flow than low-DOF scenarios. 
The influence of erroneous optical flow on image reconstruction quality is further explored next.

\subsubsection{Effect of Erroneous Optical Flow}
\label{sec:erroneousflow}

\begin{figure}[t]
	\centering
  \centering
\begin{subfigure}{0.48\linewidth}
    \includegraphics[width=\linewidth]{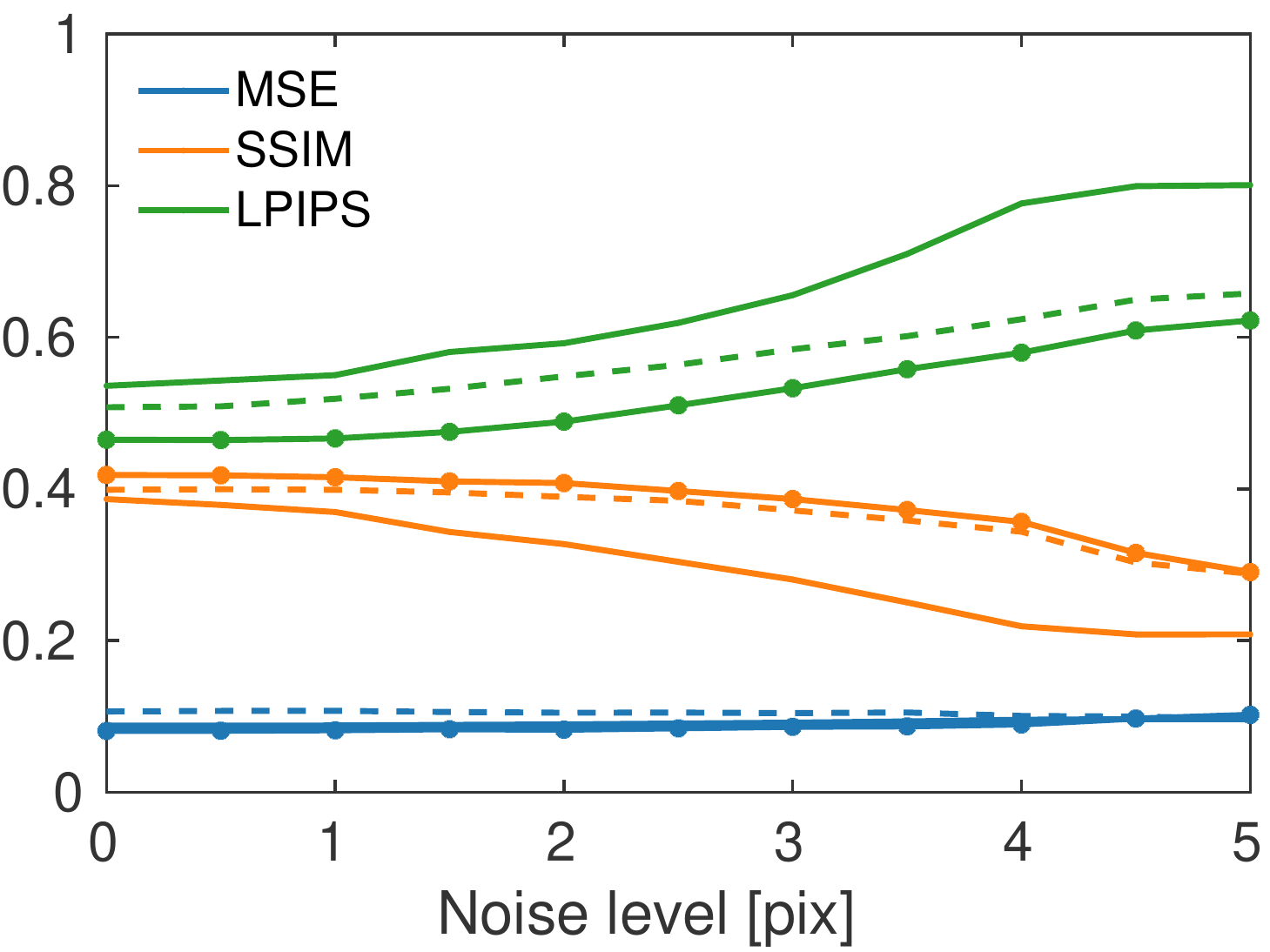}
    \caption{boxes scene\label{fig:noisyflow:curves:boxes}}
\end{subfigure} 
\begin{subfigure}{0.48\linewidth}
    \includegraphics[width=\linewidth]{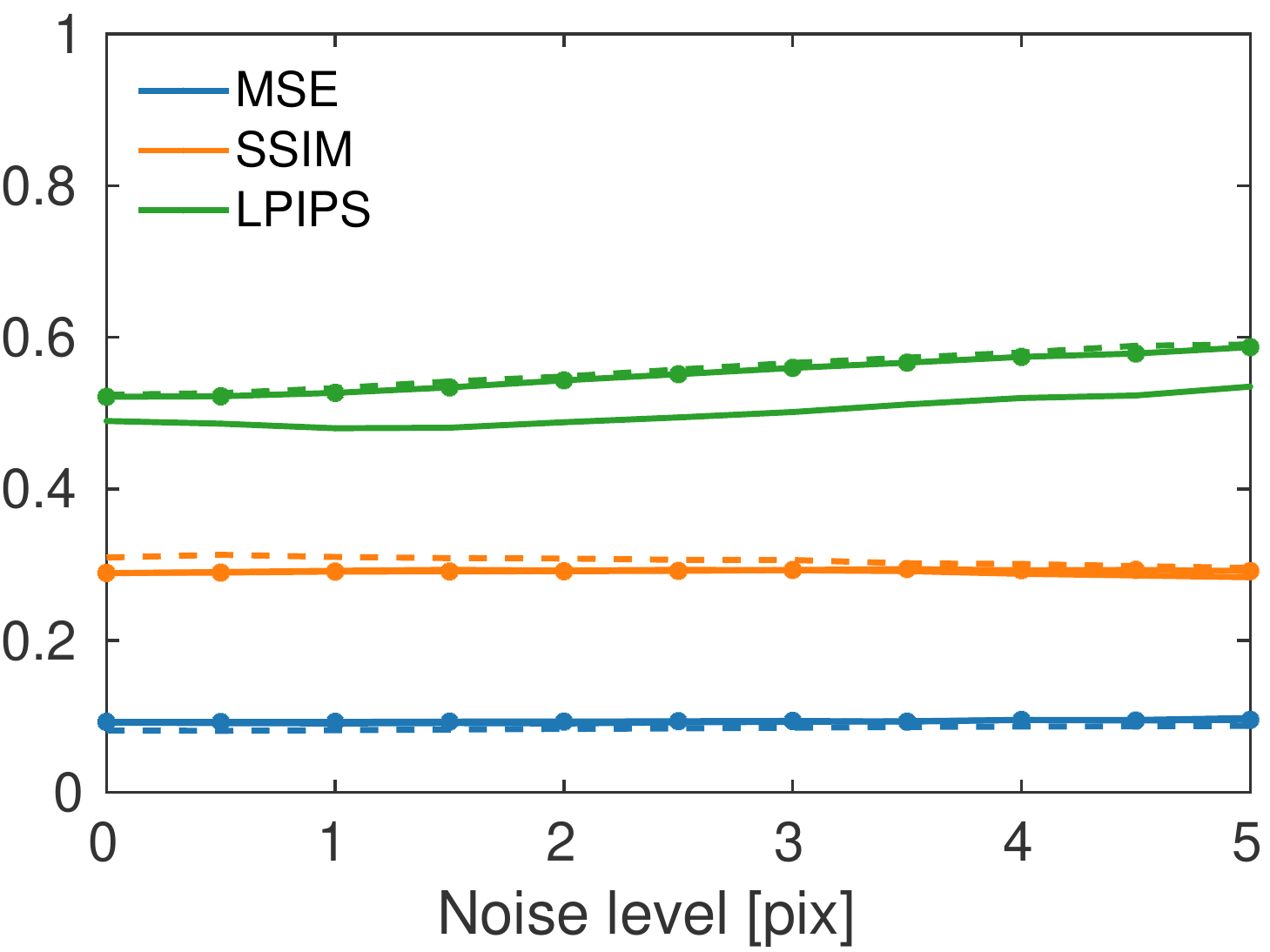}
    \caption{dynamic scene\label{fig:noisyflow:curves:dynamic}}
\end{subfigure}    
	\caption{Sensitivity study:
    Image quality metrics vs.~noise in the flow field for the three regularizers:
    CNN (solid line), TV (dashed line) and Tikhonov (solid-circled line).
    }
	\label{fig:noisyflow:curves}
\end{figure}
We further analyze the effect that estimated optical flow quality has on image reconstruction.
To this end, we use data from \cite{Mueggler17ijrr}, with high quality flow (``ground truth'') given by CMax \cite{Gallego17ral}, and corrupt the optical flow with increasing values of uniform white noise in $[-b,b]$, with $b=1,\ldots,5$ pixels (in each coordinate independently). 
Then, we assess how the quality of the reconstructed image degrades as a function of the noise power.
\Cref{fig:noisyflow:curves} quantitatively summarizes the results (the experiment was repeated 10 times for each noise level and mean values of image quality metrics are reported), while \cref{fig:noisyflow:combined} provides sample reconstructed images (see also suppl.~material).

The results look clean up to $\pm 1$ pixel additional noise (for a 240 $\times$ 180 pixel image).
Afterwards, visual image quality may degrade.
On the dynamic scene, the degradation of the visual quality is not well captured by the standard metrics used (MSE, SSIM, LPIPS): 
they all seem to stay approximately constant.
The effect is most noticeable quantitatively on the boxes scene (for $b > 2$ pixels), specially using the SSIM and LPIPS metrics (SSIM decreases and LPIPS increases with noise).
The images reconstructed with the CNN regularizer degrade more abruptly than with other regularizers.
While this is a contrived experiment (no algorithm would produce flow corrupted by large uniform white noise), 
it reassures the motivation of this work: a large portion of the problem is already solved by providing accurate optical flow.
Hence, this stimulates future research on better event-based dense optical flow estimation methods. 

\def\figWidth{0.155\linewidth} %
\begin{figure*}[th]
	\centering
    {\small
    \setlength{\tabcolsep}{2pt}
	\begin{tabular}{
	>{\centering\arraybackslash}m{0.3cm}
	>{\centering\arraybackslash}m{\figWidth} 
	>{\centering\arraybackslash}m{\figWidth}
	>{\centering\arraybackslash}m{\figWidth}
	>{\centering\arraybackslash}m{\figWidth}
	>{\centering\arraybackslash}m{\figWidth}
	>{\centering\arraybackslash}m{\figWidth}}

		\rotatebox{90}{\makecell{Optical flow}}
		&{\includegraphics[trim={0 180px 240px 0},clip,width=\linewidth]{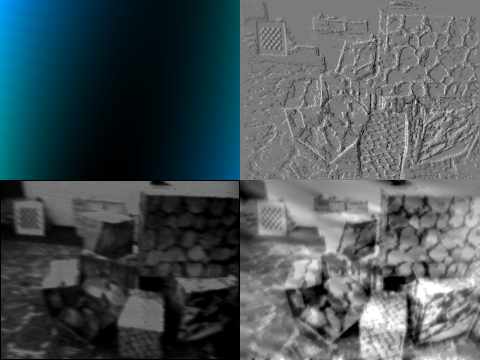}}
		&{\includegraphics[trim={0 180px 240px 0},clip,width=\linewidth]{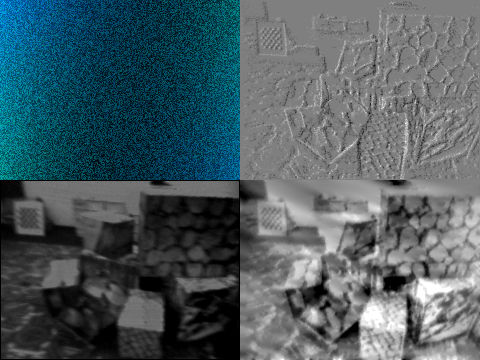}}
		&{\includegraphics[trim={0 180px 240px 0},clip,width=\linewidth]{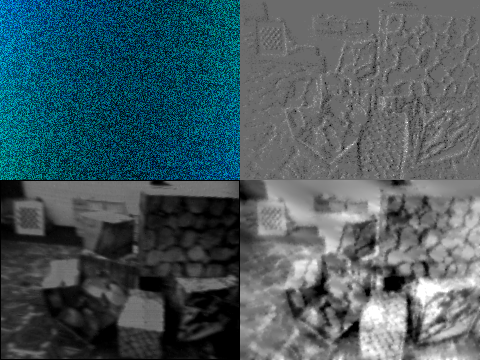}}
		&{\includegraphics[trim={0 180px 240px 0},clip,width=\linewidth]{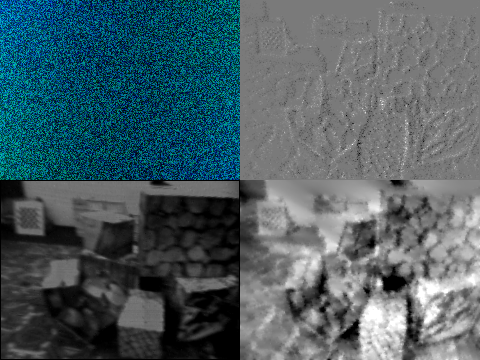}}
		&{\includegraphics[trim={0 180px 240px 0},clip,width=\linewidth]{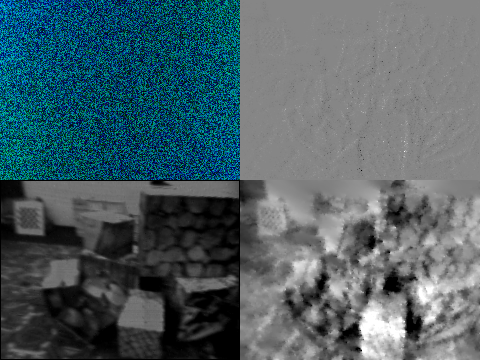}}
		&{\includegraphics[trim={0 180px 240px 0},clip,width=\linewidth]{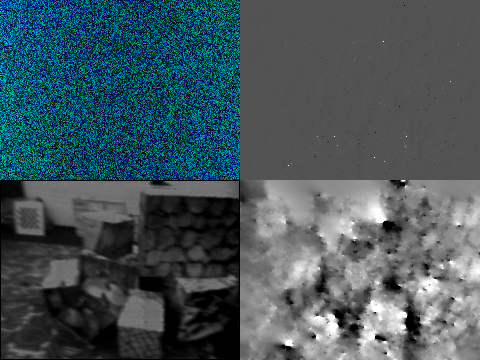}}
		\\

        \rotatebox{90}{\makecell{CNN}}
		&{\includegraphics[trim={240px 0 0 180px},clip,width=\linewidth]{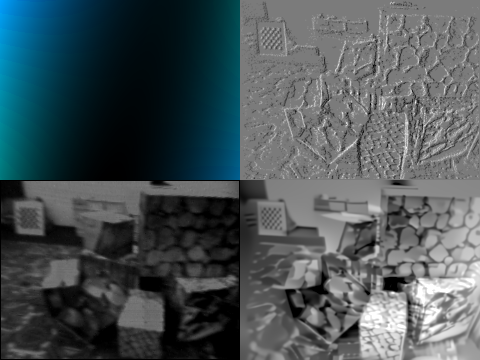}}
		&{\includegraphics[trim={240px 0 0 180px},clip,width=\linewidth]{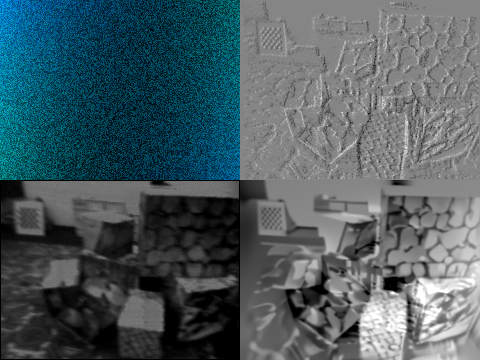}}
		&{\includegraphics[trim={240px 0 0 180px},clip,width=\linewidth]{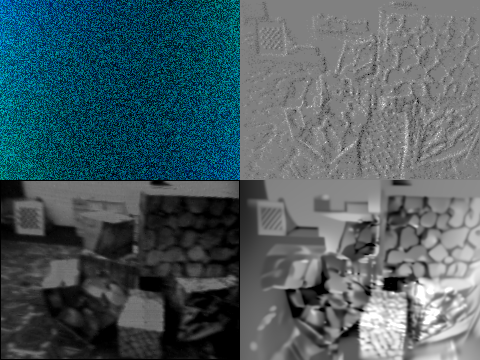}}
		&{\includegraphics[trim={240px 0 0 180px},clip,width=\linewidth]{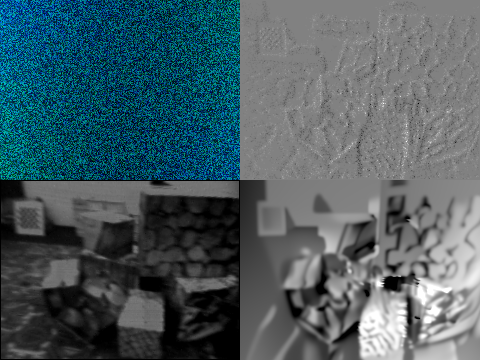}}
		&{\includegraphics[trim={240px 0 0 180px},clip,width=\linewidth]{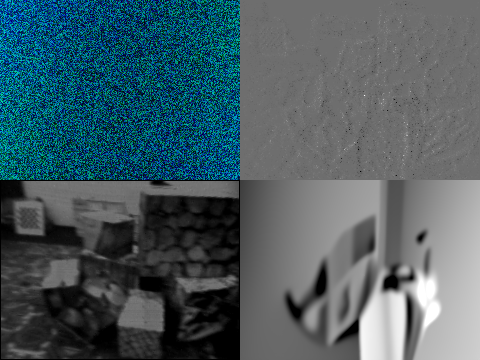}}
		&{\includegraphics[trim={240px 0 0 180px},clip,width=\linewidth]{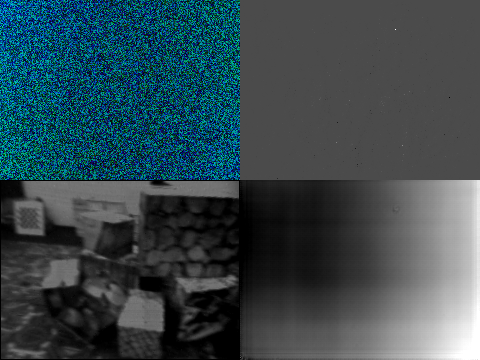}}
		\\[0.5ex]%

		\rotatebox{90}{\makecell{Optical flow}}
		&{\includegraphics[trim={0 180px 240px 0},clip,width=\linewidth]{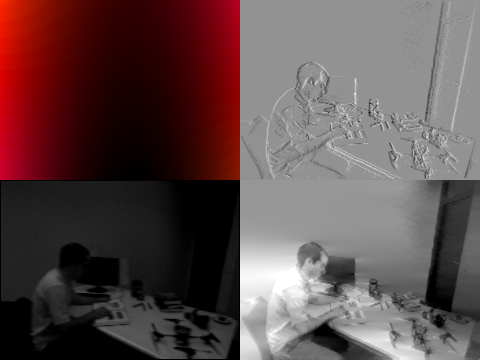}}
		&{\includegraphics[trim={0 180px 240px 0},clip,width=\linewidth]{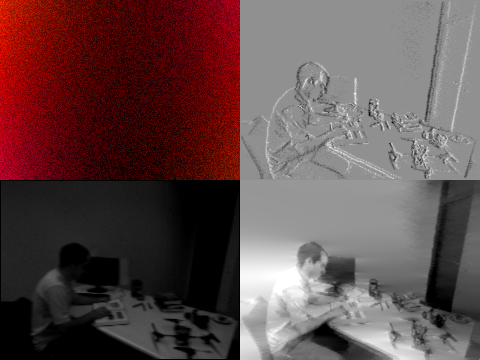}}
		&{\includegraphics[trim={0 180px 240px 0},clip,width=\linewidth]{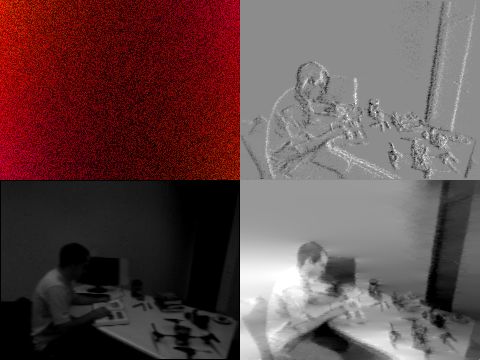}}
		&{\includegraphics[trim={0 180px 240px 0},clip,width=\linewidth]{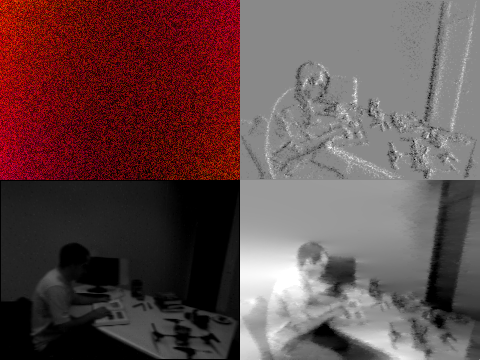}}
		&{\includegraphics[trim={0 180px 240px 0},clip,width=\linewidth]{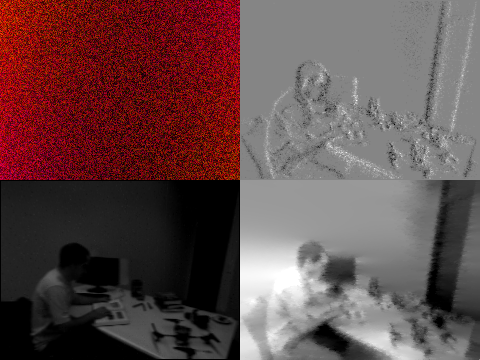}}
		&{\includegraphics[trim={0 180px 240px 0},clip,width=\linewidth]{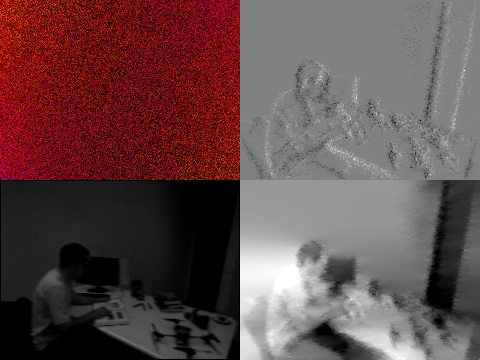}}
		\\

        \rotatebox{90}{\makecell{CNN}}
		&{\includegraphics[trim={240px 0 0 180px},clip,width=\linewidth]{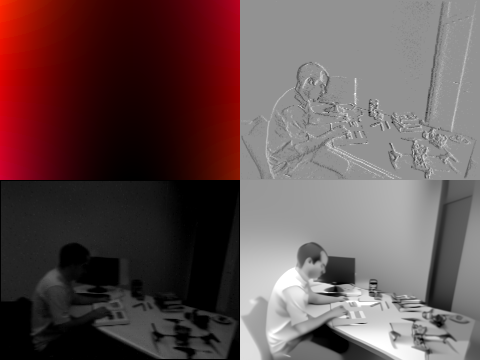}}
		&{\includegraphics[trim={240px 0 0 180px},clip,width=\linewidth]{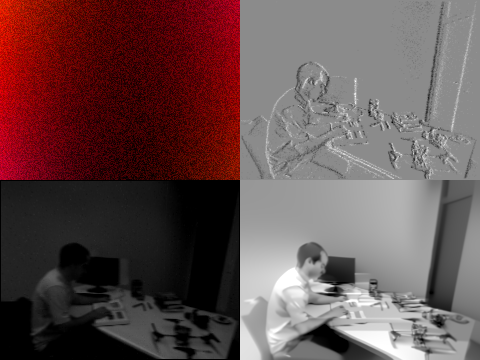}}
		&{\includegraphics[trim={240px 0 0 180px},clip,width=\linewidth]{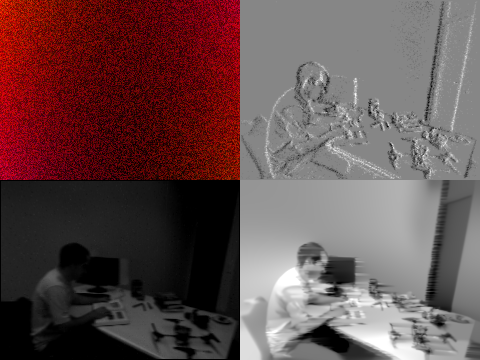}}
		&{\includegraphics[trim={240px 0 0 180px},clip,width=\linewidth]{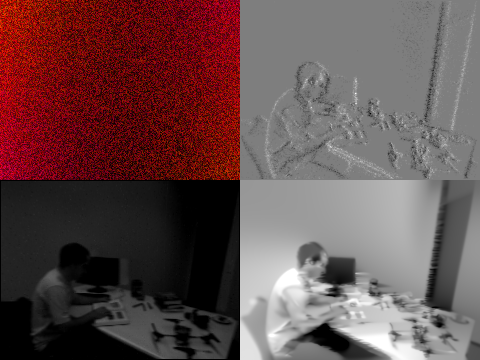}}
		&{\includegraphics[trim={240px 0 0 180px},clip,width=\linewidth]{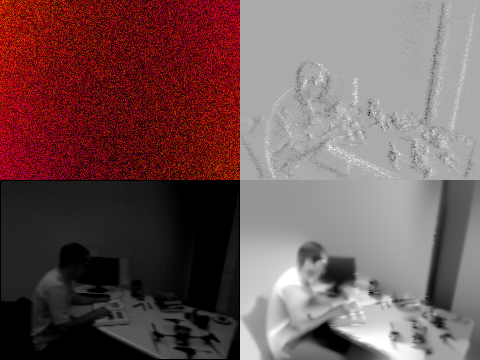}}
		&{\includegraphics[trim={240px 0 0 180px},clip,width=\linewidth]{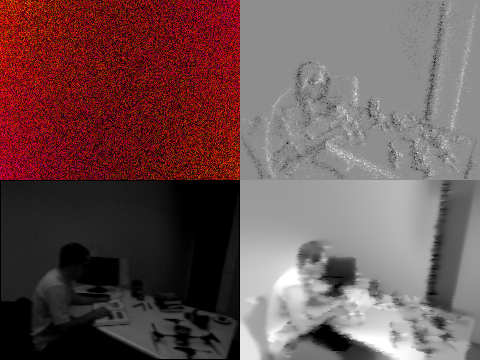}}
		\\
		
		& (a) $b=0$ pix
		& (b) $b=1$ pix
		& (c) $b=2$ pix
		& (d) $b=3$ pix
		& (e) $b=4$ pix
		& (f) $b=5$ pix
	\end{tabular}
	}
	\caption{Image reconstruction sensitivity with respect to uniform white noise in optical flow, in the range $[-b,b]$ pixels.}
	\label{fig:noisyflow:combined}
\end{figure*}

\subsection{Computational Performance}
\label{sec:experim:runtime}

The proposed method consists of multiple iterations of linear solvers with regularizers (e.g., Tikhonov, TV or CNN).
Runtime is roughly proportional to the number of iterations.
We implement the method using the standard library Pylops \cite{Ravasi20pylops}, running on a laptop's CPU 
(Intel i7-8650U CPU at 1.90 GHZ, single-threaded).
The Plug-and-Play module provided by~\cite{zhang2021plug} is used as CNN regularizer.
The following numbers are calculated by reconstructing 100 images using 30k events/image from \cite{Mueggler17ijrr} (240 $\times$ 180 pixel resolution),
collecting the runtime and taking the average.

Assuming the optical flow and NIWE are known, the image reconstruction time of our off-the-shelf implementation takes 0.4401~s (Tikhonov by regularized inversion algorithm with 100 iterations), 4.0443~s (TV by the split Bregman algorithm with 20 outer and 10 inner iterations) and 28.3904~s (CNN, 16 iterations).
For comparison, RNNs using optimized implementations take: 0.2448~s (E2VID), 0.2839~s (ECNN) and 0.4059~s (BTEB) on the same CPU.
The time to covert the events into a voxel grid to feed to the RNNs is 0.01 s.
Likewise, the unoptimized MCM method \cite{Shiba22eccv} takes 54.544 s to compute optical flow, 
while the Ev-FlowNet CNN \cite{Zhu18rss} (trained in \cite{Stoffregen20eccv,Shiba22eccv} using different loss functions) takes 0.1521 s (i.e., 350$\times$ speed-up).
Additional speed gains can be obtained when running the ANNs on a GPU, as reported by the corresponding publications.
The proposed method contains elementary operations that are local and parallelizable, hence we envision that with the appropriate hardware and software implementation considerable speed gains are to be achieved.

\def\figWidth{0.155\linewidth} %
\begin{figure*}[th]
	\centering
    {\small
    \setlength{\tabcolsep}{2pt}
	\begin{tabular}{
	>{\centering\arraybackslash}m{0.3cm}
	>{\centering\arraybackslash}m{\figWidth} 
	>{\centering\arraybackslash}m{\figWidth}
	>{\centering\arraybackslash}m{\figWidth}
	>{\centering\arraybackslash}m{\figWidth}
	>{\centering\arraybackslash}m{\figWidth}
	>{\centering\arraybackslash}m{\figWidth}}

		\rotatebox{90}{\makecell{boxes\_rot}}
		&{\includegraphics[width=\linewidth]{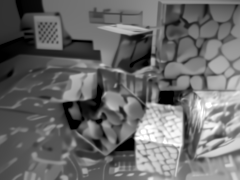}}
		&{\includegraphics[width=\linewidth]{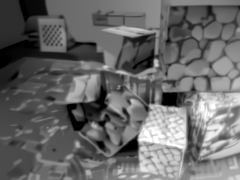}}
		&{\includegraphics[width=\linewidth]{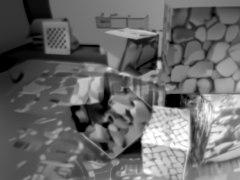}}
		&{\includegraphics[width=\linewidth]{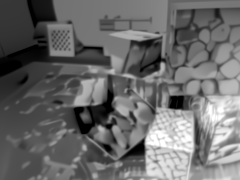}}
		&{\includegraphics[width=\linewidth]{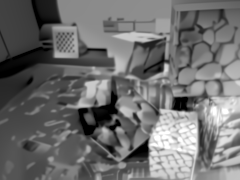}}
		&{\includegraphics[width=\linewidth]{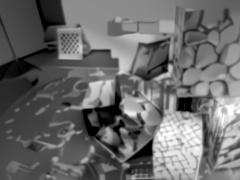}}
		\\

		\rotatebox{90}{\makecell{dynamic\_rot}}
		&{\includegraphics[width=\linewidth]{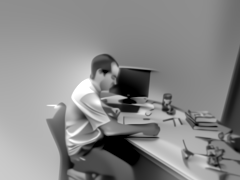}}
		&{\includegraphics[width=\linewidth]{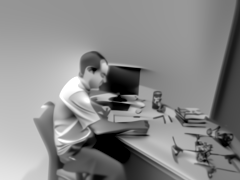}}
		&{\includegraphics[width=\linewidth]{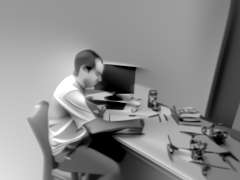}}
		&{\includegraphics[width=\linewidth]{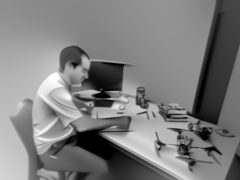}}
		&{\includegraphics[width=\linewidth]{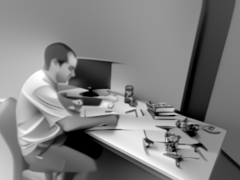}}
		&{\includegraphics[width=\linewidth]{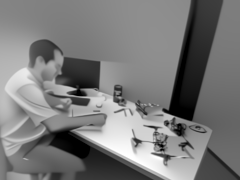}}
		\\
		
		& (a) Time index $k$ 
		& (b) $k+2$
		& (c) $k+4$
		& (d) $k+6$
		& (e) $k+8$
		& (f) $k+10$
	\end{tabular}
	}
	\caption{Temporal consistency of our method (with CNN regularizer) despite not enforcing it. 
	Event data from \cite{Mueggler17ijrr}.
	}
	\label{fig:temporalconsitency}
\end{figure*}

\def\figWidth{0.32\linewidth}
\begin{figure}[t]
	\centering
    {\small
    \setlength{\tabcolsep}{1pt}
	\begin{tabular}{
	>{\centering\arraybackslash}m{\figWidth}
	>{\centering\arraybackslash}m{\figWidth}
	>{\centering\arraybackslash}m{\figWidth}}

		{\includegraphics[width=\linewidth]{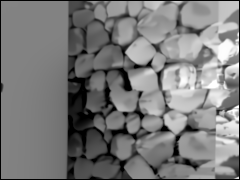}}
		&{\includegraphics[width=\linewidth]{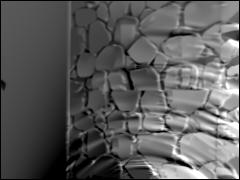}}
		&{\includegraphics[width=\linewidth]{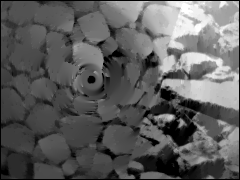}}
		\\		
		\multicolumn{2}{c}{(a) Change of direction}
		& (b) Rotation artifact
	\end{tabular}
	}
	\caption{Artifacts showing limitations of the method (CNN).}
	\label{fig:limitations}
\end{figure}

\subsection{Limitations}
\label{sec:limitations}

\emph{Artifacts}:
Despite our method not explicitly modeling temporal consistency, some degree of consistency between consecutive image reconstructions is attained, as shown in \cref{fig:temporalconsitency}.
We use CNN image priors for regularization and to remove artifacts.
While the majority of artifacts are removed, some remain (\cref{fig:limitations}).
For example, when the flow changes direction abruptly within an event packet, motion compensation with a single motion model may fail to give a good NIWE. 
Hence, two consecutively reconstructed images may change appearance considerably (\cref{fig:limitations}(a)). 
Another artifact that may arise is a black/white spot (\cref{fig:limitations}(b)), produced if the center of rotation lands in the middle of the image (no events in the NIWE).
These artifacts are due to the lack of a temporal consistency prior or recurrent connections.
Future research could look into mitigating these artifacts via temporal filtering, 
i.e., targeting video reconstruction instead of image reconstruction.

\emph{Brightness Constancy}:
Like other reconstruction methods, ours is based on the brightness constancy assumption (e.g., in \eqref{eq:niwedotproduct}).
This has pros and cons. 
Among the advantages, the method does not require supervisory signal (ground truth and/or grayscale frames), 
and it does not even need to train a network because it can leverage existing ANNs for image denoising.
The downside is that it only recovers brightness of moving edges, 
hence ($i$) it may struggle to handle events caused by flickering lights, 
and ($ii$) it might lead to inconsistencies~\cite{Paredes21cvpr} if the camera is stationary.
Supervised or semi-supervised learning methods (like \cite{Rebecq19pami,Stoffregen20eccv,Paredes21cvpr}) may forego this assumption given sufficient \emph{high quality} data, which may be difficult to obtain.
The stationary camera problem may be solved by combining events and grayscale frames~\cite{Scheerlinck18accv}.

\emph{Slow and fast objects}:
Dealing with slow and fast motions in the same scene could be problematic in image reconstruction because, in the same time window, few events would be present from the slow motion while many events would be present from the faster moving parts of the scene.
While the NIWE partially accounts for this, the brightness of the slow object would not be as reliably reconstructed as that of the fast object if the speeds were too different.
Incorporating recurrent connections that propagate information through time could mitigate this issue, 
but it would introduce others, such as increasing the initialization phase (i.e., not being able to produce short-time reconstructions \cref{fig:penguins}). There is a trade-off.
We think that the above issue could be addressed in a more sensible manner using motion segmentation \cite{Stoffregen19iccv,Zhou21tnnls}, before building the NIWE (\cref{sec:method:motseg}).

Despite the limitations, the proposed method has many appealing properties and we hope it will foster novel research on estimating both flow and intensity, and in noting that flow, while difficult, facilitates the recovery of intensity. %

\section{Conclusion}
\label{sec:conclusion}

Instead of mainstream end-to-end brightness reconstruction from events based on RNNs, 
we have emphasized the framework of simultaneously estimating both physically entangled quantities in the events: brightness and motion (optical flow).
By exploiting the asymmetries of the estimation problem, we have shown how, using motion to generate images of warped events, brightness reconstruction becomes a linear problem, which we have tackled with classical regularizers and image priors.
This novel approach is explainable, combines the physics of event generation with years of developments in linear inverse problem solvers, 
and leverages existing image denoising networks (i.e., does not require ground truth data).
In its competitive form (CNN image denoising prior), the resulting method is not strictly linear.
Beyond the particular results showing reconstruction quality in line with the state of the art, 
we believe our approach (linear equations plus image prior) is appealing because it unifies multiple problems
(including super-resolution and reconstruction based on either first --NIWE-- or second derivative --Laplacian-- of brightness),
and can be combined with motion segmentation, color demosaicing and state-of-the-art dense optical flow estimation methods.

\section*{Acknowledgments}
We thank our collaborators at University of Oxford: I. Juarez Martinez and Profs. A. Kacelnik and T. Hart for data collection in the experiment of \cref{fig:penguins}.
We also thank Shintaro Shiba for assistance with MCM \cite{Shiba22eccv},
and the anonymous reviewers for valuable suggestions.
Funded by the Deutsche Forschungsgemeinschaft (DFG, German Research Foundation) under Germany's Excellence Strategy – EXC 2002/1 ``Science of Intelligence'' – project number 390523135.

\ifCLASSOPTIONcaptionsoff
  \newpage
\fi

\ifshowanimationcommentarxiv

\onecolumn

\section*{Supplementary Material}
\begin{itemize}
    \item \Cref{fig:noisyflow:boxes:images,fig:noisyflow:dynamic:images} show the effect of erroneous optical flow. 
They are the expanded version of \cref{fig:noisyflow:combined}.
\item Tables \ref{tab:imgrec:caltech:eq}--\ref{tab:imgrec:caltech:noneq} and Figs. \ref{fig:compare:caltech:2}--\ref{fig:compare:caltech:3}
provide additional sample image reconstructions on the N-Caltech 101 dataset. 
\end{itemize}

\def\figWidth{0.155\linewidth} %
\begin{figure*}[th]
	\centering
    {\small
    \setlength{\tabcolsep}{2pt}
	\begin{tabular}{
	>{\centering\arraybackslash}m{0.3cm}
	>{\centering\arraybackslash}m{\figWidth} 
	>{\centering\arraybackslash}m{\figWidth}
	>{\centering\arraybackslash}m{\figWidth}
	>{\centering\arraybackslash}m{\figWidth}
	>{\centering\arraybackslash}m{\figWidth}
	>{\centering\arraybackslash}m{\figWidth}}

		\rotatebox{90}{\makecell{Optical flow}}
		&{\includegraphics[trim={0 180px 240px 0},clip,width=\linewidth]{images/noisy_flow_exp/boxes/noise_level_0.0_l2/000000001.png}}
		&{\includegraphics[trim={0 180px 240px 0},clip,width=\linewidth]{images/noisy_flow_exp/boxes/noise_level_1.0_l2/000000001.png}}
		&{\includegraphics[trim={0 180px 240px 0},clip,width=\linewidth]{images/noisy_flow_exp/boxes/noise_level_2.0_l2/000000001.png}}
		&{\includegraphics[trim={0 180px 240px 0},clip,width=\linewidth]{images/noisy_flow_exp/boxes/noise_level_3.0_l2/000000001.png}}
		&{\includegraphics[trim={0 180px 240px 0},clip,width=\linewidth]{images/noisy_flow_exp/boxes/noise_level_4.0_l2/000000001.png}}
		&{\includegraphics[trim={0 180px 240px 0},clip,width=\linewidth]{images/noisy_flow_exp/boxes/noise_level_5.0_l2/000000001.png}}
		\\

		\rotatebox{90}{\makecell{Tikhonov}}
        &{\includegraphics[trim={240px 0 0 180px},clip,width=\linewidth]{images/noisy_flow_exp/boxes/noise_level_0.0_l2/000000001.png}}
		&{\includegraphics[trim={240px 0 0 180px},clip,width=\linewidth]{images/noisy_flow_exp/boxes/noise_level_1.0_l2/000000001.png}}
		&{\includegraphics[trim={240px 0 0 180px},clip,width=\linewidth]{images/noisy_flow_exp/boxes/noise_level_2.0_l2/000000001.png}}
		&{\includegraphics[trim={240px 0 0 180px},clip,width=\linewidth]{images/noisy_flow_exp/boxes/noise_level_3.0_l2/000000001.png}}
		&{\includegraphics[trim={240px 0 0 180px},clip,width=\linewidth]{images/noisy_flow_exp/boxes/noise_level_4.0_l2/000000001.png}}
		&{\includegraphics[trim={240px 0 0 180px},clip,width=\linewidth]{images/noisy_flow_exp/boxes/noise_level_5.0_l2/000000001.png}}
		\\

		\rotatebox{90}{\makecell{TV}}
		&{\includegraphics[trim={240px 0 0 180px},clip,width=\linewidth]{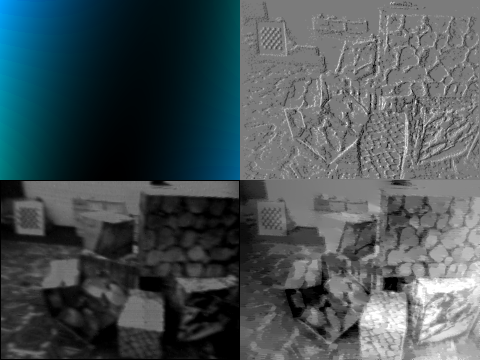}}
		&{\includegraphics[trim={240px 0 0 180px},clip,width=\linewidth]{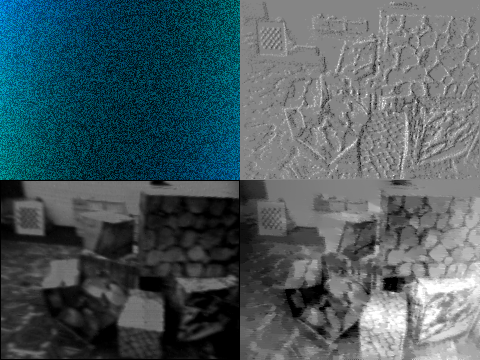}}
		&{\includegraphics[trim={240px 0 0 180px},clip,width=\linewidth]{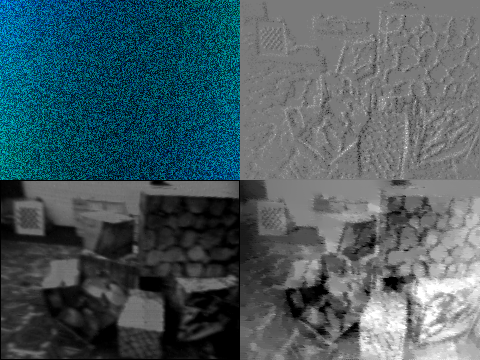}}
		&{\includegraphics[trim={240px 0 0 180px},clip,width=\linewidth]{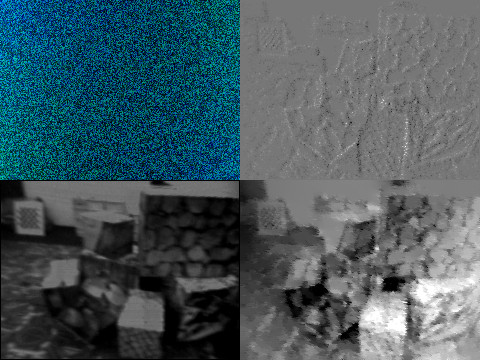}}
		&{\includegraphics[trim={240px 0 0 180px},clip,width=\linewidth]{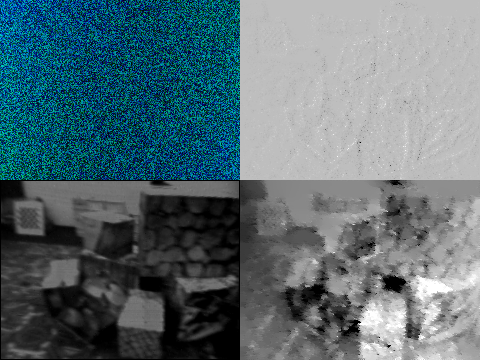}}
		&{\includegraphics[trim={240px 0 0 180px},clip,width=\linewidth]{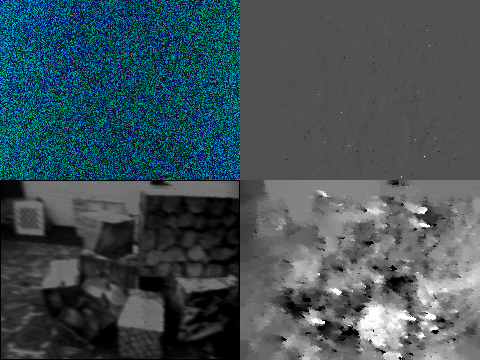}}
		\\

        \rotatebox{90}{\makecell{CNN}}
		&{\includegraphics[trim={240px 0 0 180px},clip,width=\linewidth]{images/noisy_flow_exp/boxes/noise_level_0.0_cnn/000000001.png}}
		&{\includegraphics[trim={240px 0 0 180px},clip,width=\linewidth]{images/noisy_flow_exp/boxes/noise_level_1.0_cnn/000000001.png}}
		&{\includegraphics[trim={240px 0 0 180px},clip,width=\linewidth]{images/noisy_flow_exp/boxes/noise_level_2.0_cnn/000000001.png}}
		&{\includegraphics[trim={240px 0 0 180px},clip,width=\linewidth]{images/noisy_flow_exp/boxes/noise_level_3.0_cnn/000000001.png}}
		&{\includegraphics[trim={240px 0 0 180px},clip,width=\linewidth]{images/noisy_flow_exp/boxes/noise_level_4.0_cnn/000000001.png}}
		&{\includegraphics[trim={240px 0 0 180px},clip,width=\linewidth]{images/noisy_flow_exp/boxes/noise_level_5.0_cnn/000000001.png}}
		\\
		
		& (a) $b=0$ pix
		& (b) $b=1$ pix
		& (c) $b=2$ pix
		& (d) $b=3$ pix
		& (e) $b=4$ pix
		& (f) $b=5$ pix
	\end{tabular}
	}
	\caption{Image reconstruction sensitivity with respect to uniform white noise in optical flow, in the range $[-b,b]$ pixels.}
	\label{fig:noisyflow:boxes:images}
\end{figure*}

\def\figWidth{0.155\linewidth} %
\begin{figure*}[th]
	\centering
    {\small
    \setlength{\tabcolsep}{2pt}
	\begin{tabular}{
	>{\centering\arraybackslash}m{0.3cm}
	>{\centering\arraybackslash}m{\figWidth} 
	>{\centering\arraybackslash}m{\figWidth}
	>{\centering\arraybackslash}m{\figWidth}
	>{\centering\arraybackslash}m{\figWidth}
	>{\centering\arraybackslash}m{\figWidth}
	>{\centering\arraybackslash}m{\figWidth}}

		\rotatebox{90}{\makecell{Optical flow}}
		&{\includegraphics[trim={0 180px 240px 0},clip,width=\linewidth]{images/noisy_flow_exp/dynamic/noise_level_0.0_l2/000000004.png}}
		&{\includegraphics[trim={0 180px 240px 0},clip,width=\linewidth]{images/noisy_flow_exp/dynamic/noise_level_1.0_l2/000000004.png}}
		&{\includegraphics[trim={0 180px 240px 0},clip,width=\linewidth]{images/noisy_flow_exp/dynamic/noise_level_2.0_l2/000000004.png}}
		&{\includegraphics[trim={0 180px 240px 0},clip,width=\linewidth]{images/noisy_flow_exp/dynamic/noise_level_3.0_l2/000000004.png}}
		&{\includegraphics[trim={0 180px 240px 0},clip,width=\linewidth]{images/noisy_flow_exp/dynamic/noise_level_4.0_l2/000000004.png}}
		&{\includegraphics[trim={0 180px 240px 0},clip,width=\linewidth]{images/noisy_flow_exp/dynamic/noise_level_5.0_l2/000000004.png}}
		\\

		\rotatebox{90}{\makecell{Tikhonov}}
        &{\includegraphics[trim={240px 0 0 180px},clip,width=\linewidth]{images/noisy_flow_exp/dynamic/noise_level_0.0_l2/000000004.png}}
		&{\includegraphics[trim={240px 0 0 180px},clip,width=\linewidth]{images/noisy_flow_exp/dynamic/noise_level_1.0_l2/000000004.png}}
		&{\includegraphics[trim={240px 0 0 180px},clip,width=\linewidth]{images/noisy_flow_exp/dynamic/noise_level_2.0_l2/000000004.png}}
		&{\includegraphics[trim={240px 0 0 180px},clip,width=\linewidth]{images/noisy_flow_exp/dynamic/noise_level_3.0_l2/000000004.png}}
		&{\includegraphics[trim={240px 0 0 180px},clip,width=\linewidth]{images/noisy_flow_exp/dynamic/noise_level_4.0_l2/000000004.png}}
		&{\includegraphics[trim={240px 0 0 180px},clip,width=\linewidth]{images/noisy_flow_exp/dynamic/noise_level_5.0_l2/000000004.png}}
		\\

		\rotatebox{90}{\makecell{TV}}
		&{\includegraphics[trim={240px 0 0 180px},clip,width=\linewidth]{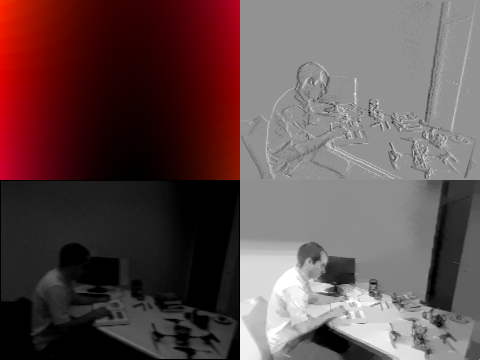}}
		&{\includegraphics[trim={240px 0 0 180px},clip,width=\linewidth]{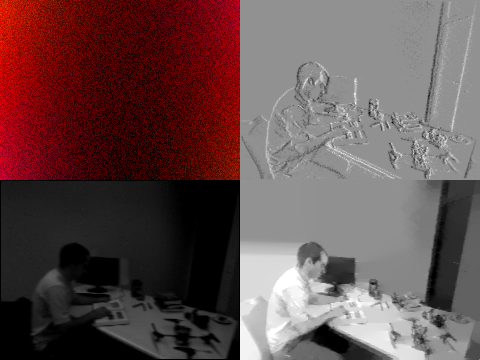}}
		&{\includegraphics[trim={240px 0 0 180px},clip,width=\linewidth]{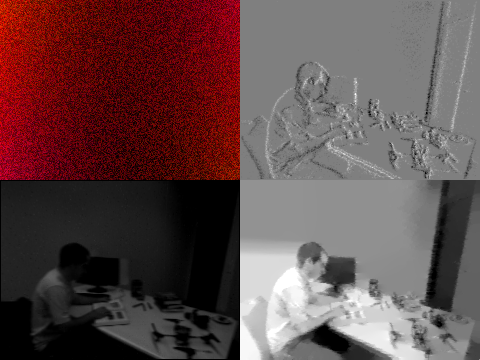}}
		&{\includegraphics[trim={240px 0 0 180px},clip,width=\linewidth]{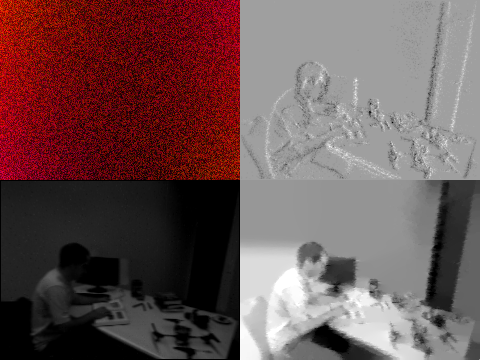}}
		&{\includegraphics[trim={240px 0 0 180px},clip,width=\linewidth]{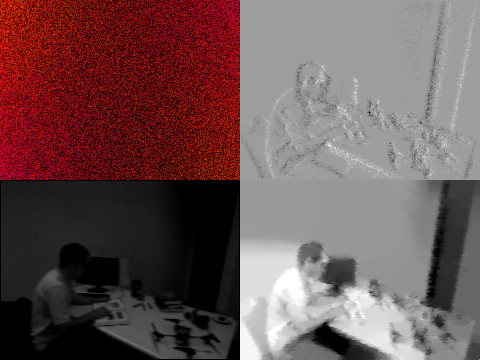}}
		&{\includegraphics[trim={240px 0 0 180px},clip,width=\linewidth]{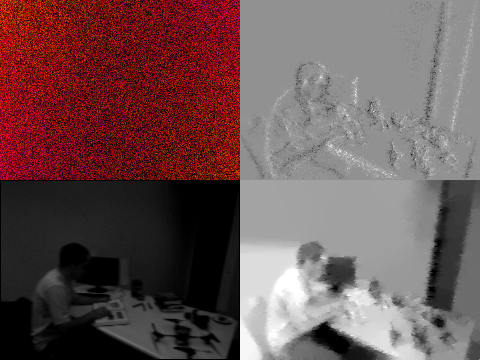}}
		\\

        \rotatebox{90}{\makecell{CNN}}
		&{\includegraphics[trim={240px 0 0 180px},clip,width=\linewidth]{images/noisy_flow_exp/dynamic/noise_level_0.0_cnn/000000004.png}}
		&{\includegraphics[trim={240px 0 0 180px},clip,width=\linewidth]{images/noisy_flow_exp/dynamic/noise_level_1.0_cnn/000000004.png}}
		&{\includegraphics[trim={240px 0 0 180px},clip,width=\linewidth]{images/noisy_flow_exp/dynamic/noise_level_2.0_cnn/000000004.png}}
		&{\includegraphics[trim={240px 0 0 180px},clip,width=\linewidth]{images/noisy_flow_exp/dynamic/noise_level_3.0_cnn/000000004.png}}
		&{\includegraphics[trim={240px 0 0 180px},clip,width=\linewidth]{images/noisy_flow_exp/dynamic/noise_level_4.0_cnn/000000004.png}}
		&{\includegraphics[trim={240px 0 0 180px},clip,width=\linewidth]{images/noisy_flow_exp/dynamic/noise_level_5.0_cnn/000000004.png}}
		\\
		
		& (a) $b=0$ pix
		& (b) $b=1$ pix
		& (c) $b=2$ pix
		& (d) $b=3$ pix
		& (e) $b=4$ pix
		& (f) $b=5$ pix
	\end{tabular}
	}
	\caption{Image reconstruction sensitivity with respect to uniform white noise in optical flow, in the range $[-b,b]$ pixels.}
	\label{fig:noisyflow:dynamic:images}
\end{figure*}

\def\figWidth{0.135\linewidth}
\newlength{\wbonsai} \setlength{\wbonsai}{552px} %
\newlength{\hbonsai} \setlength{\hbonsai}{344px} %
\newlength{\wbrain} \setlength{\wbrain}{736px} %
\newlength{\hbrain} \setlength{\hbrain}{344px} %
\newlength{\wbutterfly} \setlength{\wbutterfly}{752px} %
\newlength{\hbutterfly} \setlength{\hbutterfly}{344px} %
\newlength{\wcannon} \setlength{\wcannon}{776px} %
\newlength{\hcannon} \setlength{\hcannon}{344px} %
\newlength{\wcarside} \setlength{\wcarside}{928px} %
\newlength{\hcarside} \setlength{\hcarside}{300px} %
\newlength{\wchair} \setlength{\wchair}{448px} %
\newlength{\hchair} \setlength{\hchair}{344px} %
\begin{figure*}[t]
    \ifhideimages
    \else
	\centering
    {\small
    \setlength{\tabcolsep}{1pt}
	\begin{tabular}{
	>{\centering\arraybackslash}m{0.3cm}
	>{\centering\arraybackslash}m{\figWidth} 
	>{\centering\arraybackslash}m{\figWidth}
	>{\centering\arraybackslash}m{\figWidth}
	>{\centering\arraybackslash}m{\figWidth}
	>{\centering\arraybackslash}m{\figWidth}
	>{\centering\arraybackslash}m{\figWidth}
	>{\centering\arraybackslash}m{\figWidth}}

        \rotatebox{90}{\makecell{bonsai}}
		&{\includegraphics[trim=0px 0px {0.75\wbonsai} {0.5\hbonsai},clip, width=\linewidth]{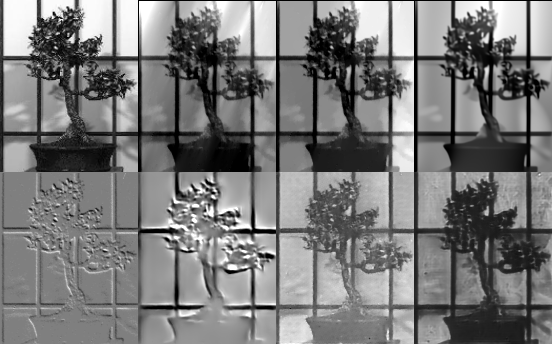}}%
		&{\includegraphics[trim={0.5\wbonsai} 0px {0.25\wbonsai} {0.5\hbonsai},clip, width=\linewidth]{images/caltech/bonsai.png}}%
		&{\includegraphics[trim={0.75\wbonsai} 0px 0px {0.5\hbonsai},clip, width=\linewidth]{images/caltech/bonsai.png}}%
		&{\includegraphics[trim={0.25\wbonsai} 0px {0.5\wbonsai} {0.5\hbonsai},clip, width=\linewidth]{images/caltech/bonsai.png}}%
		&{\includegraphics[trim={0.5\wbonsai} {0.5\hbonsai} {0.25\wbonsai} 0px,clip, width=\linewidth]{images/caltech/bonsai.png}}%
		&{\includegraphics[trim={0.75\wbonsai} {0.5\hbonsai} 0px 0px,clip, width=\linewidth]{images/caltech/bonsai.png}}%
		&{\includegraphics[trim=0px {0.5\hbonsai} {0.75\wbonsai} 0px,clip, width=\linewidth]{images/caltech/bonsai.png}}%
		\\
		
		\rotatebox{90}{\makecell{brain}}
		&{\includegraphics[trim=0px 0px {0.75\wbrain} {0.5\hbrain},clip, width=\linewidth]{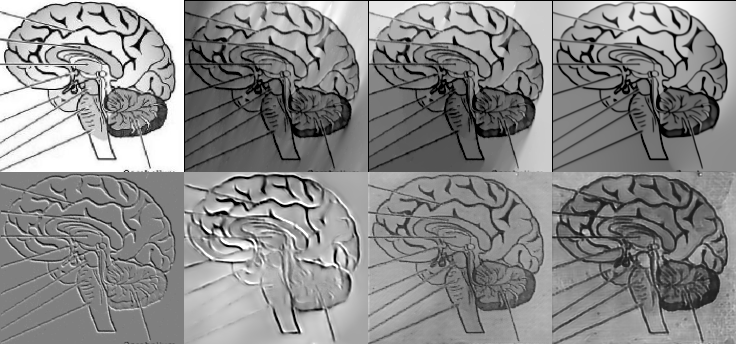}}%
		&{\includegraphics[trim={0.5\wbrain} 0px {0.25\wbrain} {0.5\hbrain},clip, width=\linewidth]{images/caltech/brain.png}}%
		&{\includegraphics[trim={0.75\wbrain} 0px 0px {0.5\hbrain},clip, width=\linewidth]{images/caltech/brain.png}}%
		&{\includegraphics[trim={0.25\wbrain} 0px {0.5\wbrain} {0.5\hbrain},clip, width=\linewidth]{images/caltech/brain.png}}%
		&{\includegraphics[trim={0.5\wbrain} {0.5\hbrain} {0.25\wbrain} 0px,clip, width=\linewidth]{images/caltech/brain.png}}%
		&{\includegraphics[trim={0.75\wbrain} {0.5\hbrain} 0px 0px,clip, width=\linewidth]{images/caltech/brain.png}}%
		&{\includegraphics[trim=0px {0.5\hbrain} {0.75\wbrain} 0px,clip, width=\linewidth]{images/caltech/brain.png}}%
		\\
		
		\rotatebox{90}{\makecell{butterfly}}
		&{\includegraphics[trim=0px 0px {0.75\wbutterfly} {0.5\hbutterfly},clip, width=\linewidth]{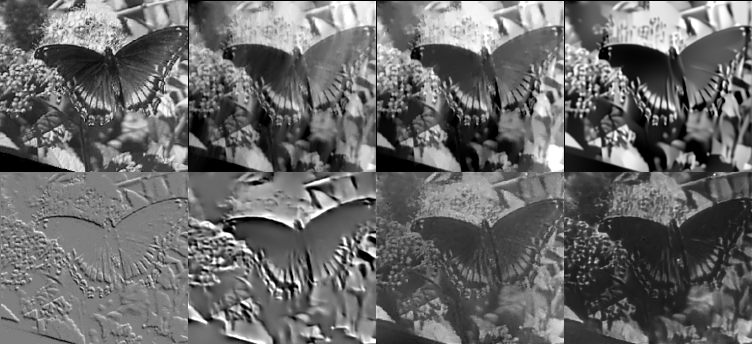}}%
		&{\includegraphics[trim={0.5\wbutterfly} 0px {0.25\wbutterfly} {0.5\hbutterfly},clip, width=\linewidth]{images/caltech/butterfly.png}}%
		&{\includegraphics[trim={0.75\wbutterfly} 0px 0px {0.5\hbutterfly},clip, width=\linewidth]{images/caltech/butterfly.png}}%
		&{\includegraphics[trim={0.25\wbutterfly} 0px {0.5\wbutterfly} {0.5\hbutterfly},clip, width=\linewidth]{images/caltech/butterfly.png}}%
		&{\includegraphics[trim={0.5\wbutterfly} {0.5\hbutterfly} {0.25\wbutterfly} 0px,clip, width=\linewidth]{images/caltech/butterfly.png}}%
		&{\includegraphics[trim={0.75\wbutterfly} {0.5\hbutterfly} 0px 0px,clip, width=\linewidth]{images/caltech/butterfly.png}}%
		&{\includegraphics[trim=0px {0.5\hbutterfly} {0.75\wbutterfly} 0px,clip, width=\linewidth]{images/caltech/butterfly.png}}%
		\\
		
		\rotatebox{90}{\makecell{camera}}
		&{\includegraphics[trim=0px 0px {0.75\wcamera} {0.5\hcamera},clip, width=\linewidth]{images/caltech/camera.png}}%
		&{\includegraphics[trim={0.5\wcamera} 0px {0.25\wcamera} {0.5\hcamera},clip, width=\linewidth]{images/caltech/camera.png}}%
		&{\includegraphics[trim={0.75\wcamera} 0px 0px {0.5\hcamera},clip, width=\linewidth]{images/caltech/camera.png}}%
		&{\includegraphics[trim={0.25\wcamera} 0px {0.5\wcamera} {0.5\hcamera},clip, width=\linewidth]{images/caltech/camera.png}}%
		&{\includegraphics[trim={0.5\wcamera} {0.5\hcamera} {0.25\wcamera} 0px,clip, width=\linewidth]{images/caltech/camera.png}}%
		&{\includegraphics[trim={0.75\wcamera} {0.5\hcamera} 0px 0px,clip, width=\linewidth]{images/caltech/camera.png}}%
		&{\includegraphics[trim=0px {0.5\hcamera} {0.75\wcamera} 0px,clip, width=\linewidth]{images/caltech/camera.png}}%
		\\
		
		\rotatebox{90}{\makecell{cannon}}
		&{\includegraphics[trim=0px 0px {0.75\wcannon} {0.5\hcannon},clip, width=\linewidth]{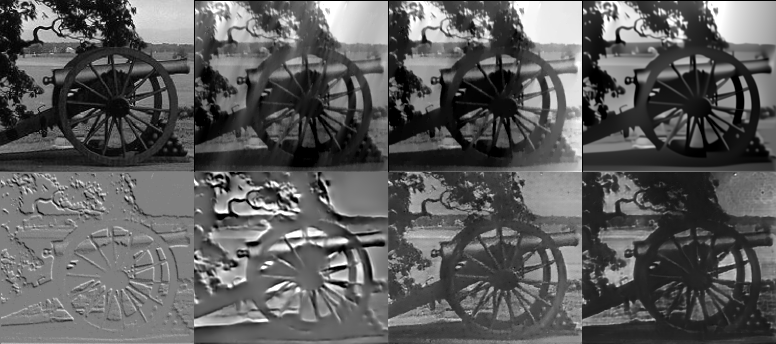}}%
		&{\includegraphics[trim={0.5\wcannon} 0px {0.25\wcannon} {0.5\hcannon},clip, width=\linewidth]{images/caltech/cannon.png}}%
		&{\includegraphics[trim={0.75\wcannon} 0px 0px {0.5\hcannon},clip, width=\linewidth]{images/caltech/cannon.png}}%
		&{\includegraphics[trim={0.25\wcannon} 0px {0.5\wcannon} {0.5\hcannon},clip, width=\linewidth]{images/caltech/cannon.png}}%
		&{\includegraphics[trim={0.5\wcannon} {0.5\hcannon} {0.25\wcannon} 0px,clip, width=\linewidth]{images/caltech/cannon.png}}%
		&{\includegraphics[trim={0.75\wcannon} {0.5\hcannon} 0px 0px,clip, width=\linewidth]{images/caltech/cannon.png}}%
		&{\includegraphics[trim=0px {0.5\hcannon} {0.75\wcannon} 0px,clip, width=\linewidth]{images/caltech/cannon.png}}%
		\\
		
		\rotatebox{90}{\makecell{car\_side}}
		&{\includegraphics[trim=0px 0px {0.75\wcarside} {0.5\hcarside},clip, width=\linewidth]{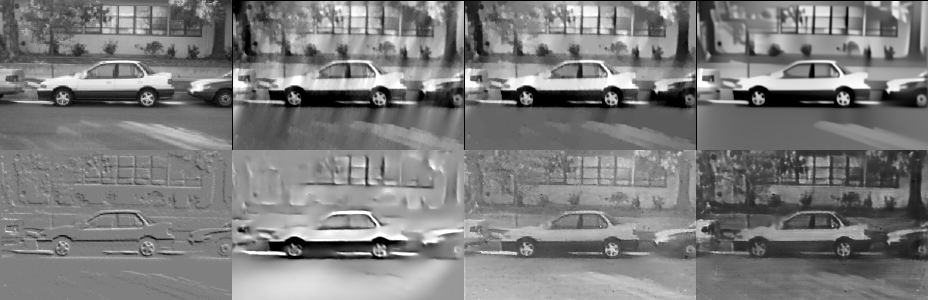}}%
		&{\includegraphics[trim={0.5\wcarside} 0px {0.25\wcarside} {0.5\hcarside},clip, width=\linewidth]{images/caltech/carside.png}}%
		&{\includegraphics[trim={0.75\wcarside} 0px 0px {0.5\hcarside},clip, width=\linewidth]{images/caltech/carside.png}}%
		&{\includegraphics[trim={0.25\wcarside} 0px {0.5\wcarside} {0.5\hcarside},clip, width=\linewidth]{images/caltech/carside.png}}%
		&{\includegraphics[trim={0.5\wcarside} {0.5\hcarside} {0.25\wcarside} 0px,clip, width=\linewidth]{images/caltech/carside.png}}%
		&{\includegraphics[trim={0.75\wcarside} {0.5\hcarside} 0px 0px,clip, width=\linewidth]{images/caltech/carside.png}}%
		&{\includegraphics[trim=0px {0.5\hcarside} {0.75\wcarside} 0px,clip, width=\linewidth]{images/caltech/carside.png}}%
		\\
		
		\rotatebox{90}{\makecell{cellphone}}
		&{\includegraphics[trim=0px 0px {0.75\wcellphone} {0.5\hcellphone},clip, width=\linewidth]{images/caltech/cellphone.png}}%
		&{\includegraphics[trim={0.5\wcellphone} 0px {0.25\wcellphone} {0.5\hcellphone},clip, width=\linewidth]{images/caltech/cellphone.png}}%
		&{\includegraphics[trim={0.75\wcellphone} 0px 0px {0.5\hcellphone},clip, width=\linewidth]{images/caltech/cellphone.png}}%
		&{\includegraphics[trim={0.25\wcellphone} 0px {0.5\wcellphone} {0.5\hcellphone},clip, width=\linewidth]{images/caltech/cellphone.png}}%
		&{\includegraphics[trim={0.5\wcellphone} {0.5\hcellphone} {0.25\wcellphone} 0px,clip, width=\linewidth]{images/caltech/cellphone.png}}%
		&{\includegraphics[trim={0.75\wcellphone} {0.5\hcellphone} 0px 0px,clip, width=\linewidth]{images/caltech/cellphone.png}}%
		&{\includegraphics[trim=0px {0.5\hcellphone} {0.75\wcellphone} 0px,clip, width=\linewidth]{images/caltech/cellphone.png}}%
		\\
		
		\rotatebox{90}{\makecell{chair}}
		&{\includegraphics[trim=0px 0px {0.75\wchair} {0.5\hchair},clip, width=\linewidth]{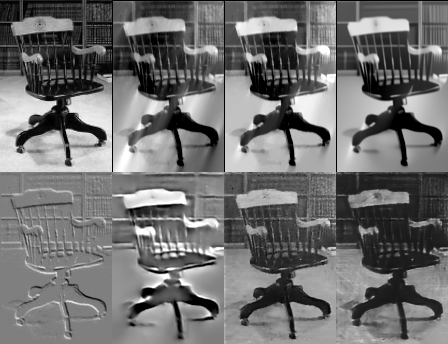}}%
		&{\includegraphics[trim={0.5\wchair} 0px {0.25\wchair} {0.5\hchair},clip, width=\linewidth]{images/caltech/chair.png}}%
		&{\includegraphics[trim={0.75\wchair} 0px 0px {0.5\hchair},clip, width=\linewidth]{images/caltech/chair.png}}%
		&{\includegraphics[trim={0.25\wchair} 0px {0.5\wchair} {0.5\hchair},clip, width=\linewidth]{images/caltech/chair.png}}%
		&{\includegraphics[trim={0.5\wchair} {0.5\hchair} {0.25\wchair} 0px,clip, width=\linewidth]{images/caltech/chair.png}}%
		&{\includegraphics[trim={0.75\wchair} {0.5\hchair} 0px 0px,clip, width=\linewidth]{images/caltech/chair.png}}%
		&{\includegraphics[trim=0px {0.5\hchair} {0.75\wchair} 0px,clip, width=\linewidth]{images/caltech/chair.png}}%
		\\

		& (a) NIWE %
		& (b) E2VID~\cite{Rebecq19pami}
		& (c) ECNN~\cite{Stoffregen20eccv}
		& (d) BTEB~\cite{Paredes21cvpr}
		& (e) Ours (TV)
		& (f) Ours (CNN)
		& (g) Ground truth
	\end{tabular}
	}
	\fi
 	\caption{Sample reconstruction results on N-Caltech 101 dataset \cite{Orchard15fns}.
 	Continuation of Fig.~8.
 	}
	\label{fig:compare:caltech:2}
\end{figure*}
\def\figWidth{0.135\linewidth}
\newlength{\wceilingfan} \setlength{\wceilingfan}{928px} %
\newlength{\hceilingfan} \setlength{\hceilingfan}{260px} %
\newlength{\wchandelier} \setlength{\wchandelier}{928px} %
\newlength{\hchandelier} \setlength{\hchandelier}{344px} %
\newlength{\wcougar} \setlength{\wcougar}{928px} %
\newlength{\hcougar} \setlength{\hcougar}{344px} %
\newlength{\wcrab} \setlength{\wcrab}{928px} %
\newlength{\hcrab} \setlength{\hcrab}{308px} %
\newlength{\wcrayfish} \setlength{\wcrayfish}{928px} %
\newlength{\hcrayfish} \setlength{\hcrayfish}{340px} %
\newlength{\wcrocodile} \setlength{\wcrocodile}{928px} %
\newlength{\hcrocodile} \setlength{\hcrocodile}{296px} %
\newlength{\wdalmatian} \setlength{\wdalmatian}{800px} %
\newlength{\hdalmatian} \setlength{\hdalmatian}{344px} %
\newlength{\wdollar} \setlength{\wdollar}{928px} %
\newlength{\hdollar} \setlength{\hdollar}{188px} %
\newlength{\wdolphin} \setlength{\wdolphin}{928px} %
\newlength{\hdolphin} \setlength{\hdolphin}{272px} %
\newlength{\wdragonfly} \setlength{\wdragonfly}{544px} %
\newlength{\hdragonfly} \setlength{\hdragonfly}{344px} %
\begin{figure*}[t]
    \ifhideimages
    \else
	\centering
    {\small
    \setlength{\tabcolsep}{1pt}
	\begin{tabular}{
	>{\centering\arraybackslash}m{0.3cm}
	>{\centering\arraybackslash}m{\figWidth} 
	>{\centering\arraybackslash}m{\figWidth}
	>{\centering\arraybackslash}m{\figWidth}
	>{\centering\arraybackslash}m{\figWidth}
	>{\centering\arraybackslash}m{\figWidth}
	>{\centering\arraybackslash}m{\figWidth}
	>{\centering\arraybackslash}m{\figWidth}}

        \rotatebox{90}{\makecell{ceilingfan}}
		&{\includegraphics[trim=0px 0px {0.75\wceilingfan} {0.5\hceilingfan},clip, width=\linewidth]{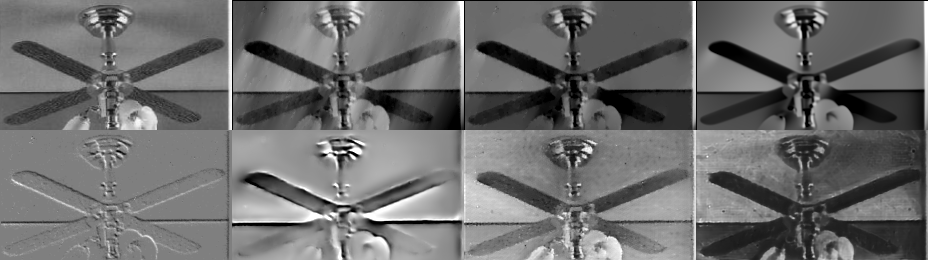}}%
		&{\includegraphics[trim={0.5\wceilingfan} 0px {0.25\wceilingfan} {0.5\hceilingfan},clip, width=\linewidth]{images/caltech/ceilingfan.png}}%
		&{\includegraphics[trim={0.75\wceilingfan} 0px 0px {0.5\hceilingfan},clip, width=\linewidth]{images/caltech/ceilingfan.png}}%
		&{\includegraphics[trim={0.25\wceilingfan} 0px {0.5\wceilingfan} {0.5\hceilingfan},clip, width=\linewidth]{images/caltech/ceilingfan.png}}%
		&{\includegraphics[trim={0.5\wceilingfan} {0.5\hceilingfan} {0.25\wceilingfan} 0px,clip, width=\linewidth]{images/caltech/ceilingfan.png}}%
		&{\includegraphics[trim={0.75\wceilingfan} {0.5\hceilingfan} 0px 0px,clip, width=\linewidth]{images/caltech/ceilingfan.png}}%
		&{\includegraphics[trim=0px {0.5\hceilingfan} {0.75\wceilingfan} 0px,clip, width=\linewidth]{images/caltech/ceilingfan.png}}%
		\\
		
		\rotatebox{90}{\makecell{chandelier}}
		&{\includegraphics[trim=0px 0px {0.75\wchandelier} {0.5\hchandelier},clip, width=\linewidth]{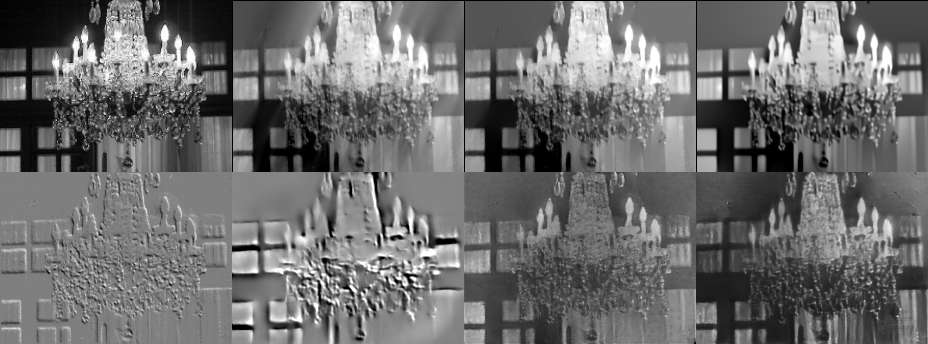}}%
		&{\includegraphics[trim={0.5\wchandelier} 0px {0.25\wchandelier} {0.5\hchandelier},clip, width=\linewidth]{images/caltech/chandelier.png}}%
		&{\includegraphics[trim={0.75\wchandelier} 0px 0px {0.5\hchandelier},clip, width=\linewidth]{images/caltech/chandelier.png}}%
		&{\includegraphics[trim={0.25\wchandelier} 0px {0.5\wchandelier} {0.5\hchandelier},clip, width=\linewidth]{images/caltech/chandelier.png}}%
		&{\includegraphics[trim={0.5\wchandelier} {0.5\hchandelier} {0.25\wchandelier} 0px,clip, width=\linewidth]{images/caltech/chandelier.png}}%
		&{\includegraphics[trim={0.75\wchandelier} {0.5\hchandelier} 0px 0px,clip, width=\linewidth]{images/caltech/chandelier.png}}%
		&{\includegraphics[trim=0px {0.5\hchandelier} {0.75\wchandelier} 0px,clip, width=\linewidth]{images/caltech/chandelier.png}}%
		\\
		
		\rotatebox{90}{\makecell{cougar}}
		&{\includegraphics[trim=0px 0px {0.75\wcougar} {0.5\hcougar},clip, width=\linewidth]{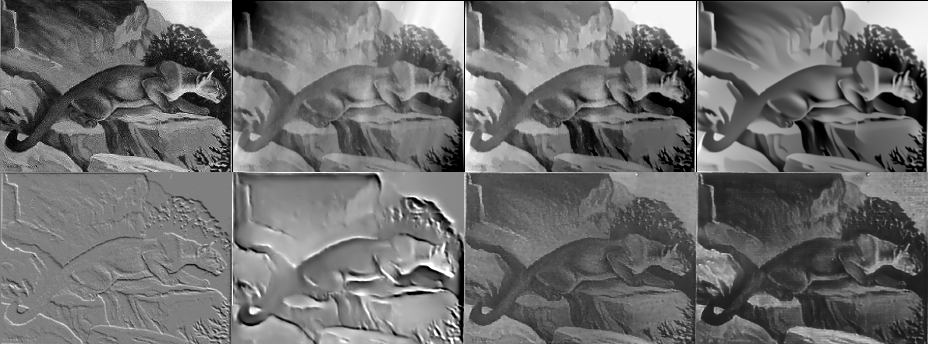}}%
		&{\includegraphics[trim={0.5\wcougar} 0px {0.25\wcougar} {0.5\hcougar},clip, width=\linewidth]{images/caltech/cougar.png}}%
		&{\includegraphics[trim={0.75\wcougar} 0px 0px {0.5\hcougar},clip, width=\linewidth]{images/caltech/cougar.png}}%
		&{\includegraphics[trim={0.25\wcougar} 0px {0.5\wcougar} {0.5\hcougar},clip, width=\linewidth]{images/caltech/cougar.png}}%
		&{\includegraphics[trim={0.5\wcougar} {0.5\hcougar} {0.25\wcougar} 0px,clip, width=\linewidth]{images/caltech/cougar.png}}%
		&{\includegraphics[trim={0.75\wcougar} {0.5\hcougar} 0px 0px,clip, width=\linewidth]{images/caltech/cougar.png}}%
		&{\includegraphics[trim=0px {0.5\hcougar} {0.75\wcougar} 0px,clip, width=\linewidth]{images/caltech/cougar.png}}%
		\\
		
		\rotatebox{90}{\makecell{crab}}
		&{\includegraphics[trim=0px 0px {0.75\wcrab} {0.5\hcrab},clip, width=\linewidth]{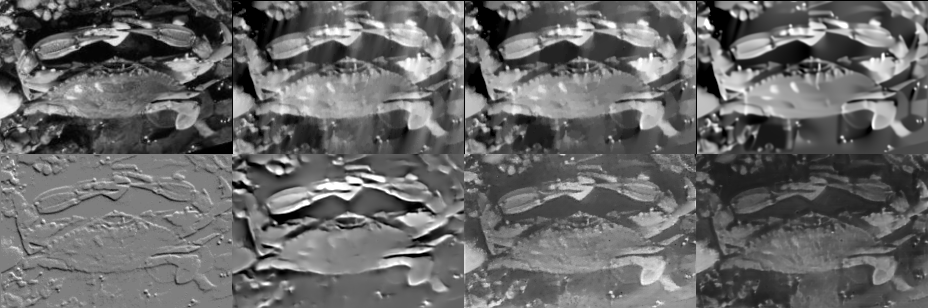}}%
		&{\includegraphics[trim={0.5\wcrab} 0px {0.25\wcrab} {0.5\hcrab},clip, width=\linewidth]{images/caltech/crab.png}}%
		&{\includegraphics[trim={0.75\wcrab} 0px 0px {0.5\hcrab},clip, width=\linewidth]{images/caltech/crab.png}}%
		&{\includegraphics[trim={0.25\wcrab} 0px {0.5\wcrab} {0.5\hcrab},clip, width=\linewidth]{images/caltech/crab.png}}%
		&{\includegraphics[trim={0.5\wcrab} {0.5\hcrab} {0.25\wcrab} 0px,clip, width=\linewidth]{images/caltech/crab.png}}%
		&{\includegraphics[trim={0.75\wcrab} {0.5\hcrab} 0px 0px,clip, width=\linewidth]{images/caltech/crab.png}}%
		&{\includegraphics[trim=0px {0.5\hcrab} {0.75\wcrab} 0px,clip, width=\linewidth]{images/caltech/crab.png}}%
		\\
		
		\rotatebox{90}{\makecell{crayfish}}
		&{\includegraphics[trim=0px 0px {0.75\wcrayfish} {0.5\hcrayfish},clip, width=\linewidth]{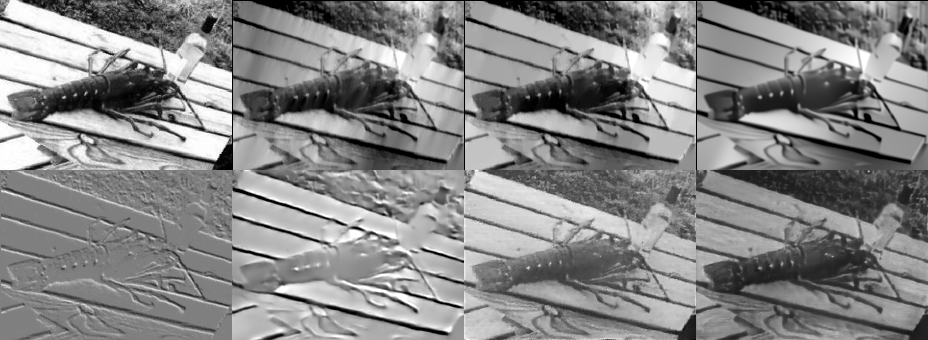}}%
		&{\includegraphics[trim={0.5\wcrayfish} 0px {0.25\wcrayfish} {0.5\hcrayfish},clip, width=\linewidth]{images/caltech/crayfish.png}}%
		&{\includegraphics[trim={0.75\wcrayfish} 0px 0px {0.5\hcrayfish},clip, width=\linewidth]{images/caltech/crayfish.png}}%
		&{\includegraphics[trim={0.25\wcrayfish} 0px {0.5\wcrayfish} {0.5\hcrayfish},clip, width=\linewidth]{images/caltech/crayfish.png}}%
		&{\includegraphics[trim={0.5\wcrayfish} {0.5\hcrayfish} {0.25\wcrayfish} 0px,clip, width=\linewidth]{images/caltech/crayfish.png}}%
		&{\includegraphics[trim={0.75\wcrayfish} {0.5\hcrayfish} 0px 0px,clip, width=\linewidth]{images/caltech/crayfish.png}}%
		&{\includegraphics[trim=0px {0.5\hcrayfish} {0.75\wcrayfish} 0px,clip, width=\linewidth]{images/caltech/crayfish.png}}%
		\\
		
		\rotatebox{90}{\makecell{crocodile}}
		&{\includegraphics[trim=0px 0px {0.75\wcrocodile} {0.5\hcrocodile},clip, width=\linewidth]{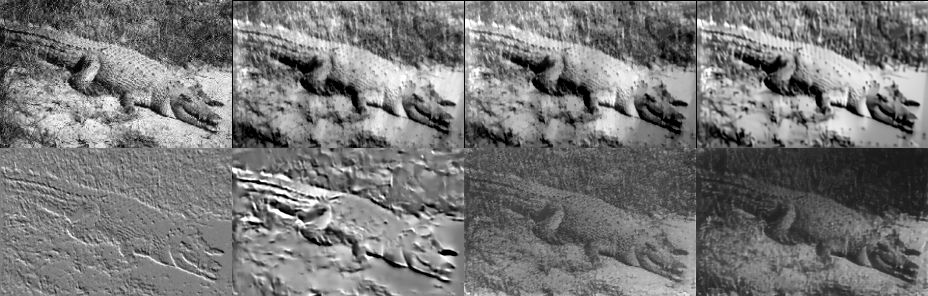}}%
		&{\includegraphics[trim={0.5\wcrocodile} 0px {0.25\wcrocodile} {0.5\hcrocodile},clip, width=\linewidth]{images/caltech/crocodile.png}}%
		&{\includegraphics[trim={0.75\wcrocodile} 0px 0px {0.5\hcrocodile},clip, width=\linewidth]{images/caltech/crocodile.png}}%
		&{\includegraphics[trim={0.25\wcrocodile} 0px {0.5\wcrocodile} {0.5\hcrocodile},clip, width=\linewidth]{images/caltech/crocodile.png}}%
		&{\includegraphics[trim={0.5\wcrocodile} {0.5\hcrocodile} {0.25\wcrocodile} 0px,clip, width=\linewidth]{images/caltech/crocodile.png}}%
		&{\includegraphics[trim={0.75\wcrocodile} {0.5\hcrocodile} 0px 0px,clip, width=\linewidth]{images/caltech/crocodile.png}}%
		&{\includegraphics[trim=0px {0.5\hcrocodile} {0.75\wcrocodile} 0px,clip, width=\linewidth]{images/caltech/crocodile.png}}%
		\\
		
		\rotatebox{90}{\makecell{cup}}
		&{\includegraphics[trim=0px 0px {0.75\wcup} {0.5\hcup},clip, width=\linewidth]{images/caltech/cup.png}}%
		&{\includegraphics[trim={0.5\wcup} 0px {0.25\wcup} {0.5\hcup},clip, width=\linewidth]{images/caltech/cup.png}}%
		&{\includegraphics[trim={0.75\wcup} 0px 0px {0.5\hcup},clip, width=\linewidth]{images/caltech/cup.png}}%
		&{\includegraphics[trim={0.25\wcup} 0px {0.5\wcup} {0.5\hcup},clip, width=\linewidth]{images/caltech/cup.png}}%
		&{\includegraphics[trim={0.5\wcup} {0.5\hcup} {0.25\wcup} 0px,clip, width=\linewidth]{images/caltech/cup.png}}%
		&{\includegraphics[trim={0.75\wcup} {0.5\hcup} 0px 0px,clip, width=\linewidth]{images/caltech/cup.png}}%
		&{\includegraphics[trim=0px {0.5\hcup} {0.75\wcup} 0px,clip, width=\linewidth]{images/caltech/cup.png}}%
		\\
		
		\rotatebox{90}{\makecell{dalmatian}}
		&{\includegraphics[trim=0px 0px {0.75\wdalmatian} {0.5\hdalmatian},clip, width=\linewidth]{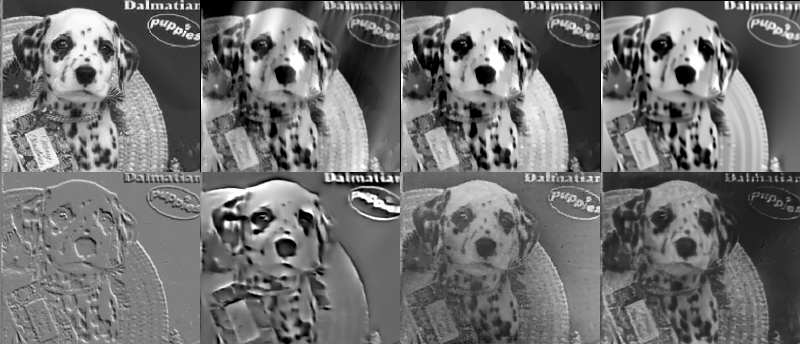}}%
		&{\includegraphics[trim={0.5\wdalmatian} 0px {0.25\wdalmatian} {0.5\hdalmatian},clip, width=\linewidth]{images/caltech/dalmatian.png}}%
		&{\includegraphics[trim={0.75\wdalmatian} 0px 0px {0.5\hdalmatian},clip, width=\linewidth]{images/caltech/dalmatian.png}}%
		&{\includegraphics[trim={0.25\wdalmatian} 0px {0.5\wdalmatian} {0.5\hdalmatian},clip, width=\linewidth]{images/caltech/dalmatian.png}}%
		&{\includegraphics[trim={0.5\wdalmatian} {0.5\hdalmatian} {0.25\wdalmatian} 0px,clip, width=\linewidth]{images/caltech/dalmatian.png}}%
		&{\includegraphics[trim={0.75\wdalmatian} {0.5\hdalmatian} 0px 0px,clip, width=\linewidth]{images/caltech/dalmatian.png}}%
		&{\includegraphics[trim=0px {0.5\hdalmatian} {0.75\wdalmatian} 0px,clip, width=\linewidth]{images/caltech/dalmatian.png}}%
		\\
		
		\rotatebox{90}{\makecell{dollar}}
		&{\includegraphics[trim=0px 0px {0.75\wdollar} {0.5\hdollar},clip, width=\linewidth]{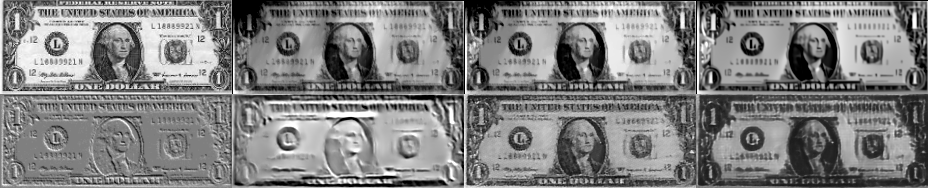}}%
		&{\includegraphics[trim={0.5\wdollar} 0px {0.25\wdollar} {0.5\hdollar},clip, width=\linewidth]{images/caltech/dollar.png}}%
		&{\includegraphics[trim={0.75\wdollar} 0px 0px {0.5\hdollar},clip, width=\linewidth]{images/caltech/dollar.png}}%
		&{\includegraphics[trim={0.25\wdollar} 0px {0.5\wdollar} {0.5\hdollar},clip, width=\linewidth]{images/caltech/dollar.png}}%
		&{\includegraphics[trim={0.5\wdollar} {0.5\hdollar} {0.25\wdollar} 0px,clip, width=\linewidth]{images/caltech/dollar.png}}%
		&{\includegraphics[trim={0.75\wdollar} {0.5\hdollar} 0px 0px,clip, width=\linewidth]{images/caltech/dollar.png}}%
		&{\includegraphics[trim=0px {0.5\hdollar} {0.75\wdollar} 0px,clip, width=\linewidth]{images/caltech/dollar.png}}%
		\\
		
		\rotatebox{90}{\makecell{dolphin}}
		&{\includegraphics[trim=0px 0px {0.75\wdolphin} {0.5\hdolphin},clip, width=\linewidth]{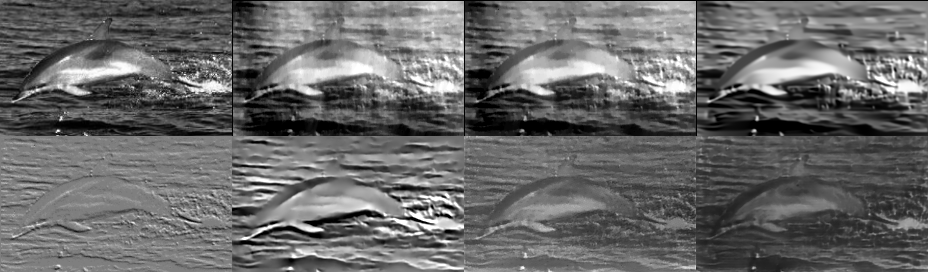}}%
		&{\includegraphics[trim={0.5\wdolphin} 0px {0.25\wdolphin} {0.5\hdolphin},clip, width=\linewidth]{images/caltech/dolphin.png}}%
		&{\includegraphics[trim={0.75\wdolphin} 0px 0px {0.5\hdolphin},clip, width=\linewidth]{images/caltech/dolphin.png}}%
		&{\includegraphics[trim={0.25\wdolphin} 0px {0.5\wdolphin} {0.5\hdolphin},clip, width=\linewidth]{images/caltech/dolphin.png}}%
		&{\includegraphics[trim={0.5\wdolphin} {0.5\hdolphin} {0.25\wdolphin} 0px,clip, width=\linewidth]{images/caltech/dolphin.png}}%
		&{\includegraphics[trim={0.75\wdolphin} {0.5\hdolphin} 0px 0px,clip, width=\linewidth]{images/caltech/dolphin.png}}%
		&{\includegraphics[trim=0px {0.5\hdolphin} {0.75\wdolphin} 0px,clip, width=\linewidth]{images/caltech/dolphin.png}}%
		\\
		
		\rotatebox{90}{\makecell{dragonfly}}
		&{\includegraphics[trim=0px 0px {0.75\wdragonfly} {0.5\hdragonfly},clip, width=\linewidth]{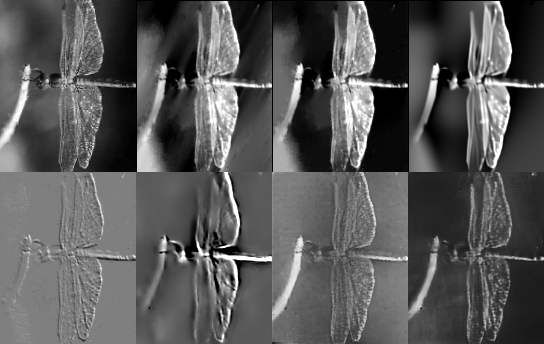}}%
		&{\includegraphics[trim={0.5\wdragonfly} 0px {0.25\wdragonfly} {0.5\hdragonfly},clip, width=\linewidth]{images/caltech/dragonfly.png}}%
		&{\includegraphics[trim={0.75\wdragonfly} 0px 0px {0.5\hdragonfly},clip, width=\linewidth]{images/caltech/dragonfly.png}}%
		&{\includegraphics[trim={0.25\wdragonfly} 0px {0.5\wdragonfly} {0.5\hdragonfly},clip, width=\linewidth]{images/caltech/dragonfly.png}}%
		&{\includegraphics[trim={0.5\wdragonfly} {0.5\hdragonfly} {0.25\wdragonfly} 0px,clip, width=\linewidth]{images/caltech/dragonfly.png}}%
		&{\includegraphics[trim={0.75\wdragonfly} {0.5\hdragonfly} 0px 0px,clip, width=\linewidth]{images/caltech/dragonfly.png}}%
		&{\includegraphics[trim=0px {0.5\hdragonfly} {0.75\wdragonfly} 0px,clip, width=\linewidth]{images/caltech/dragonfly.png}}%
		\\

		& (a) NIWE %
		& (b) E2VID~\cite{Rebecq19pami}
		& (c) ECNN~\cite{Stoffregen20eccv}
		& (d) BTEB~\cite{Paredes21cvpr}
		& (e) Ours (TV)
		& (f) Ours (CNN)
		& (g) Ground truth
	\end{tabular}
	}
	\fi
 	\caption{
 	Continuation of Fig.~21.
 	}
	\label{fig:compare:caltech:3}
\end{figure*}

\begin{table*}[t]
\centering
\caption{Quantitative evaluation of our method and state-of-the-art methods on sequences from N-Caltech \cite{Orchard15fns}.
We report median values of MSE, SSIM and LPIPS, since they are more robust to outliers than the mean.
Images are histogram-equalized before computing the metrics.}
\label{tab:imgrec:caltech:eq}
\ifhideimages
\else
\begin{adjustbox}{max width=.9\linewidth}
\setlength{\tabcolsep}{4pt}
\begin{tabular}{@{}l *{18}{S[table-format=1.4]}@{}}
             & \multicolumn{6}{c}{MSE $\downarrow$} & \multicolumn{6}{c}{SSIM $\uparrow$} & \multicolumn{6}{c}{LPIPS $\downarrow$}\\
             \cmidrule(l{2mm}r{2mm}){2-7} \cmidrule(l{2mm}r{2mm}){8-13} \cmidrule(l{2mm}r{2mm}){14-19} %
               & & & & \multicolumn{3}{c}{Ours}
               & & & & \multicolumn{3}{c}{Ours}
               & & & & \multicolumn{3}{c}{Ours}\\
             \cmidrule(l{2mm}r{2mm}){5-7} \cmidrule(l{2mm}r{2mm}){11-13} \cmidrule(l{2mm}r{2mm}){17-19} %
Sequence name  & $\text{E2VID}$ & $\text{ECNN}$ & $\text{BTEB}$ & $\text{Tikh.}$ & $\text{TV}$ & $\text{CNN}$
               & $\text{E2VID}$ & $\text{ECNN}$ & $\text{BTEB}$ & $\text{Tikh.}$ & $\text{TV}$ & $\text{CNN}$
               & $\text{E2VID}$ & $\text{ECNN}$ & $\text{BTEB}$ & $\text{Tikh.}$ & $\text{TV}$ & $\text{CNN}$\\
	
\midrule
accordion & 0.05920569 & \bnum{0.04133756} & 0.09192133 & 0.05387457 & 0.04631127 & \unum{0.04423198} & 0.46505177 & 0.4392318 & 0.27290267 & 0.5294351 & \bnum{0.5382289} & \unum{0.5300511} & \bnum{0.36127275} & \unum{0.3781837} & 0.49198204 & 0.37534684 & 0.39216262 & 0.41052425\\
airplanes & 0.04434557 & \bnum{0.03315404} & 0.09494756 & 0.08677819 & \unum{0.04434109} & 0.04640213 & 0.48904812 & 0.43381727 & 0.3593611 & 0.55019516 & \bnum{0.63780785} & \unum{0.6178733} & \unum{0.40091142} & 0.41816688 & 0.49376547 & 0.42437786 & \bnum{0.38407898} & 0.40787825\\
anchor & \bnum{0.07895362377166748} & \unum{0.08014652132987976} & 0.14517439901828766 & 0.1119549497961998 & 0.08265967667102814 & 0.10606463998556137 & 0.35056912899017334 & 0.2931300401687622 & 0.33214420080184937 & 0.4596564769744873 & \bnum{0.496509850025177} & \unum{0.47631335258483887} & \unum{0.4630645513534546} & \bnum{0.4622228741645813} & 0.5624786615371704 & 0.4963926076889038 & 0.4657703638076782 & 0.5144096612930298  \\
ant & 0.06838316470384598 & \unum{0.05460770055651665} & 0.10222232341766357 & 0.07041549682617188 & \bnum{0.054568637162446976} & 0.06461641937494278 & 0.3723912239074707 & 0.38958972692489624 & 0.2932029962539673 & 0.44560372829437256 & \bnum{0.48722946643829346} & \unum{0.4474900960922241} & \unum{0.43296945095062256} & \bnum{0.42934781312942505} & 0.5620637536048889 & 0.45405980944633484 & 0.4591345489025116 & 0.523811399936676  \\
barrel & 0.04756002128124237 & 0.04426400735974312 & 0.08994776755571365 & 0.060764130204916 & \unum{0.04172297194600105} & \bnum{0.040528975427150726} & 0.3999851942062378 & 0.39429283142089844 & 0.30231720209121704 & 0.5369876623153687 & \bnum{0.5551163554191589} & \unum{0.5505121946334839} & \bnum{0.3611087203025818} & \unum{0.3699135184288025} & 0.5019426345825195 & 0.39370742440223694 & 0.37823474407196045 & 0.4327196180820465  \\
bass & 0.06340480595827103 & \bnum{0.036082107573747635} & 0.08687697350978851 & 0.05962083116173744 & 0.04928707331418991 & \unum{0.048419781029224396} & 0.3803844451904297 & 0.3897179961204529 & 0.3273437023162842 & 0.5028129816055298 & \unum{0.5286797285079956} & \bnum{0.542693555355072} & \unum{0.44413304328918457} & \bnum{0.43620288372039795} & 0.5535010695457458 & 0.45362424850463867 & 0.462230384349823 & 0.4792623519897461  \\
beaver & 0.06035503000020981 & \unum{0.04186065122485161} & 0.07708714157342911 & 0.05560486763715744 & \bnum{0.03508351370692253} & 0.04298391938209534 & 0.4494360089302063 & 0.4654979109764099 & 0.3304046392440796 & \unum{0.5983344912528992} & \bnum{0.6200309991836548} & 0.5961582064628601 & \unum{0.39806419610977173} & 0.4060823321342468 & 0.5353305339813232 & \bnum{0.3960363268852234} & 0.39997732639312744 & 0.4868474304676056  \\
binocular & \unum{0.03167545795440674} & \bnum{0.02752923220396042} & 0.07824055105447769 & 0.06106695905327797 & 0.03899497911334038 & 0.03838518634438515 & 0.44337910413742065 & 0.4374879002571106 & 0.3318655490875244 & 0.533951997756958 & \bnum{0.573147714138031} & \unum{0.5386011004447937} & 0.4264776408672333 & \bnum{0.4100019037723541} & 0.5089706778526306 & 0.4267992377281189 & \unum{0.41466647386550903} & 0.4510359466075897  \\
bonsai & 0.05586470291018486 & \unum{0.041108570992946625} & 0.09891459345817566 & 0.06017015501856804 & 0.04253366217017174 & \bnum{0.039393018931150436} & 0.4179765582084656 & 0.3799135684967041 & 0.3089304566383362 & 0.5538474917411804 & \bnum{0.579352080821991} & \unum{0.5727180242538452} & \bnum{0.4146419167518616} & 0.45867788791656494 & 0.5533571243286133 & 0.43438127636909485 & \unum{0.4153587222099304} & 0.48880714178085327  \\
brain & 0.1054486557841301 & 0.07398700714111328 & 0.13757596909999847 & 0.07727378606796265 & \bnum{0.061683837324380875} & \unum{0.06986674666404724} & 0.39223921298980713 & 0.4027059078216553 & 0.24518990516662598 & 0.5370796918869019 & \bnum{0.5714069604873657} & \unum{0.5491352081298828} & 0.43724125623703003 & 0.4137842655181885 & 0.5280857086181641 & \bnum{0.37202611565589905} & \unum{0.3737241327762604} & 0.3817756474018097  \\
brontosaurus & 0.055112529546022415 & \bnum{0.0418318472802639} & 0.10467161983251572 & 0.0655779093503952 & \unum{0.051161784678697586} & 0.054910808801651 & 0.40591108798980713 & 0.3872390389442444 & 0.32378894090652466 & \unum{0.5199416875839233} & \bnum{0.5438218116760254} & 0.5185747146606445 & \bnum{0.4121262729167938} & 0.43439263105392456 & 0.5397356152534485 & 0.43581581115722656 & \unum{0.4320756793022156} & 0.4968794882297516  \\
buddha & \unum{0.05010215938091278} & \bnum{0.040443722158670425} & 0.10248655080795288 & 0.06029251590371132 & 0.05316578596830368 & 0.052091680467128754 & 0.39291948080062866 & 0.3512565493583679 & 0.28191787004470825 & \unum{0.5415487289428711} & \bnum{0.5520004034042358} & 0.5308400392532349 & \bnum{0.4178471565246582} & 0.4473504424095154 & 0.5330840349197388 & 0.4338293671607971 & \unum{0.42823970317840576} & 0.4369816184043884  \\
butterfly & 0.052644629031419754 & 0.05584395304322243 & 0.08825968205928802 & 0.06131717562675476 & \unum{0.050945546478033066} & \bnum{0.048045773059129715} & 0.446377158164978 & 0.40295910835266113 & 0.33132344484329224 & 0.5539247989654541 & \bnum{0.5936809778213501} & \unum{0.581508994102478} & 0.40813949704170227 & \unum{0.39452064037323} & 0.5155377388000488 & 0.41696634888648987 & \bnum{0.3804284930229187} & 0.431555837392807  \\
camera & \unum{0.031807489693164825} & \bnum{0.02767108380794525} & 0.08118672668933868 & 0.06380262970924377 & 0.03864750638604164 & 0.046748898923397064 & 0.4718201160430908 & 0.4537389278411865 & 0.3109561800956726 & 0.5463349223136902 & \bnum{0.5850237607955933} & \unum{0.5743359923362732} & \bnum{0.3615739941596985} & 0.3695060610771179 & 0.48924684524536133 & 0.39428403973579407 & 0.36479121446609497 & \unum{0.36420339345932007}  \\
cannon & 0.04493533819913864 & \bnum{0.036298323422670364} & 0.08818970620632172 & 0.05520068481564522 & 0.04203244298696518 & \unum{0.0407850444316864} & 0.5035405158996582 & 0.4605047106742859 & 0.33229392766952515 & 0.5705517530441284 & \bnum{0.5996747016906738} & \unum{0.5887491703033447} & \bnum{0.3568233251571655} & \unum{0.3921821415424347} & 0.5013376474380493 & 0.4025394916534424 & 0.39647096395492554 & 0.4166979193687439  \\
car\_side & 0.050085924565792084 & 0.05068323761224747 & 0.07043207436800003 & 0.043637096881866455 & \bnum{0.03285269811749458} & \unum{0.03873759135603905} & 0.4602116346359253 & 0.4059053659439087 & 0.3295777440071106 & 0.5691913962364197 & \bnum{0.5895100831985474} & \unum{0.5853464603424072} & \bnum{0.397295206785202} & \unum{0.39814114570617676} & 0.5128684639930725 & 0.42056211829185486 & 0.40623122453689575 & 0.43578192591667175  \\
ceiling\_fan & 0.08070658892393112 & \bnum{0.05825161561369896} & 0.10847032815217972 & 0.09325104206800461 & \unum{0.06692247092723846} & 0.07406622171401978 & 0.33761662244796753 & 0.25470399856567383 & 0.32629328966140747 & 0.4413711428642273 & \bnum{0.5270103812217712} & \unum{0.5180511474609375} & \unum{0.49957242608070374} & \bnum{0.4842064082622528} & 0.5443176031112671 & 0.5106675624847412 & 0.5083732604980469 & 0.5038788914680481  \\
cellphone & \bnum{0.05016696825623512} & \unum{0.05327599495649338} & 0.08221014589071274 & 0.06462471187114716 & 0.06705645471811295 & 0.06389880180358887 & 0.36738884449005127 & 0.2827969193458557 & 0.3122937083244324 & 0.49555057287216187 & \bnum{0.5230194926261902} & \unum{0.5139230489730835} & 0.43428799510002136 & 0.4614194929599762 & 0.5269297957420349 & \unum{0.4330025017261505} & \bnum{0.4190545082092285} & 0.4509209394454956  \\
chair & 0.059580255299806595 & 0.0632527768611908 & 0.1041107028722763 & 0.06591349840164185 & \bnum{0.049352001398801804} & \unum{0.057878874242305756} & 0.3760153651237488 & 0.33842217922210693 & 0.31748300790786743 & 0.5083968043327332 & \bnum{0.5707113742828369} & \unum{0.541695237159729} & \bnum{0.3793850541114807} & \unum{0.40725070238113403} & 0.510323166847229 & 0.4087839424610138 & 0.4085971415042877 & 0.43486490845680237  \\
chandelier & 0.08108989149332047 & 0.058993566781282425 & 0.10311378538608551 & \unum{0.05303599685430527} & 0.05600418895483017 & \bnum{0.04097024351358414} & 0.3721616864204407 & 0.3867626190185547 & 0.30065327882766724 & \unum{0.576994776725769} & 0.5693923234939575 & \bnum{0.5988384485244751} & 0.4413093328475952 & 0.45141518115997314 & 0.5189170837402344 & \bnum{0.4041717052459717} & \unum{0.4259799122810364} & 0.4545212984085083  \\
cougar & 0.06179078295826912 & \unum{0.04386880621314049} & 0.09191004186868668 & 0.055381473153829575 & 0.053011227399110794 & \bnum{0.043137311935424805} & 0.3842419385910034 & 0.38355231285095215 & 0.2955048084259033 & \unum{0.5537697672843933} & \bnum{0.5691473484039307} & 0.5166137218475342 & \bnum{0.4024360477924347} & \unum{0.411210834980011} & 0.5382527709007263 & 0.41276973485946655 & 0.45109328627586365 & 0.505876898765564  \\
crab & \unum{0.048894692212343216} & \bnum{0.04353783652186394} & 0.09890098124742508 & 0.06469515711069107 & 0.04945109784603119 & 0.050996504724025726 & 0.4633876085281372 & 0.424114465713501 & 0.30781275033950806 & \bnum{0.5555942058563232} & \unum{0.5544618964195251} & 0.5478019714355469 & \unum{0.3963928818702698} & \bnum{0.3957732915878296} & 0.5370287895202637 & 0.42675355076789856 & 0.423483669757843 & 0.47972655296325684  \\
crayfish & 0.05825410410761833 & \bnum{0.04379455745220184} & 0.09468171745538712 & 0.07187914103269577 & \unum{0.05560005456209183} & 0.06814546138048172 & 0.3985081911087036 & 0.34788650274276733 & 0.2907261848449707 & 0.4325811266899109 & \unum{0.47241920232772827} & \bnum{0.48267245292663574} & \bnum{0.4213452935218811} & \unum{0.44564494490623474} & 0.5529583096504211 & 0.4873311519622803 & 0.4927322566509247 & 0.5391013622283936  \\
crocodile & 0.07699239253997803 & 0.06375645101070404 & 0.0953725278377533 & 0.04916372895240784 & \unum{0.045627422630786896} & \bnum{0.044034551829099655} & 0.3990076184272766 & 0.40907710790634155 & 0.2707560062408447 & \unum{0.5835325717926025} & \bnum{0.5895148515701294} & 0.5698639154434204 & \bnum{0.4072498083114624} & 0.4383198022842407 & 0.5709301233291626 & \unum{0.4201224446296692} & 0.4334113299846649 & 0.503730297088623  \\
cup & 0.053679127246141434 & 0.049010325223207474 & 0.08974333107471466 & 0.057435452938079834 & \bnum{0.0338597409427166} & \unum{0.04542430117726326} & 0.38677895069122314 & 0.35887449979782104 & 0.2854449152946472 & 0.45915746688842773 & \unum{0.5129889249801636} & \bnum{0.5227615833282471} & 0.47929811477661133 & 0.46530160307884216 & 0.5471017360687256 & \unum{0.45681262016296387} & \bnum{0.45339471101760864} & 0.4597465991973877  \\
dalmatian & 0.04461020603775978 & 0.0373668372631073 & 0.07551831007003784 & 0.043098919093608856 & \bnum{0.030232327058911324} & \unum{0.032228756695985794} & 0.5079151391983032 & 0.4817243218421936 & 0.36600399017333984 & 0.5781332850456238 & \bnum{0.6342573165893555} & \unum{0.6077073812484741} & \unum{0.39270931482315063} & 0.4130457043647766 & 0.5031092166900635 & 0.4069434404373169 & \bnum{0.379324734210968} & 0.42127421498298645  \\
dollar\_bill & 0.040366675704717636 & \bnum{0.030370719730854034} & 0.07159542292356491 & 0.04729548841714859 & 0.04309890419244766 & \unum{0.037491217255592346} & 0.5259498357772827 & 0.5094071626663208 & 0.3714723587036133 & \unum{0.6255489587783813} & \bnum{0.6483675241470337} & 0.6243689060211182 & \unum{0.30921298265457153} & 0.33179235458374023 & 0.4515431523323059 & 0.317617803812027 & \bnum{0.3057001829147339} & 0.3848824203014374  \\
dolphin & 0.08505678176879883 & 0.07144179940223694 & 0.09129410982131958 & 0.0616462416946888 & \unum{0.061646223068237305} & \bnum{0.056533921509981155} & 0.34157097339630127 & 0.32977861166000366 & 0.27936798334121704 & \unum{0.46010822057724} & \bnum{0.4716986417770386} & 0.4336060881614685 & \unum{0.44675159454345703} & \bnum{0.4461768865585327} & 0.5355566740036011 & 0.46188920736312866 & 0.46609294414520264 & 0.5203084945678711  \\
dragonfly & 0.10002018511295319 & \bnum{0.07126935571432114} & 0.13358855247497559 & 0.09218084067106247 & \unum{0.08565560728311539} & 0.09385768324136734 & 0.31379711627960205 & 0.30256789922714233 & 0.2700846791267395 & 0.47110670804977417 & \unum{0.4782138466835022} & \bnum{0.4831279516220093} & \bnum{0.449353963136673} & 0.4655168950557709 & 0.5538030862808228 & 0.4747101068496704 & \unum{0.4523850679397583} & 0.5117101669311523  \\
\midrule
\end{tabular}
\end{adjustbox}
\fi
\end{table*}
\begin{table*}[t]
\centering
\caption{Quantitative evaluation like in \cref{tab:imgrec:caltech:eq} (N-Caltech data), but \emph{without} histogram equalization.
}
\label{tab:imgrec:caltech:noneq}
\ifhideimages
\else
\begin{adjustbox}{max width=.9\linewidth}
\setlength{\tabcolsep}{4pt}
\begin{tabular}{@{}l *{18}{S[table-format=1.4]}@{}}
             & \multicolumn{6}{c}{MSE $\downarrow$} & \multicolumn{6}{c}{SSIM $\uparrow$} & \multicolumn{6}{c}{LPIPS $\downarrow$}\\
             \cmidrule(l{2mm}r{2mm}){2-7} \cmidrule(l{2mm}r{2mm}){8-13} \cmidrule(l{2mm}r{2mm}){14-19} %
               & & & & \multicolumn{3}{c}{Ours}
               & & & & \multicolumn{3}{c}{Ours}
               & & & & \multicolumn{3}{c}{Ours}\\
             \cmidrule(l{2mm}r{2mm}){5-7} \cmidrule(l{2mm}r{2mm}){11-13} \cmidrule(l{2mm}r{2mm}){17-19} %
Sequence name  & $\text{E2VID}$ & $\text{ECNN}$ & $\text{BTEB}$ & $\text{Tikh.}$ & $\text{TV}$ & $\text{CNN}$
               & $\text{E2VID}$ & $\text{ECNN}$ & $\text{BTEB}$ & $\text{Tikh.}$ & $\text{TV}$ & $\text{CNN}$
               & $\text{E2VID}$ & $\text{ECNN}$ & $\text{BTEB}$ & $\text{Tikh.}$ & $\text{TV}$ & $\text{CNN}$\\
	
\midrule
accordion & 0.04976331 & 0.03664065 & 0.09021059 & 0.04515319 & \unum{0.0365205} & \bnum{0.03307699} & 0.4629091 & 0.46193713 & 0.3209064 & 0.5807396 & \unum{0.60255873} & \bnum{0.63323015} & \unum{0.33585393} & 0.35358733 & 0.46132484 & 0.34597668 & \bnum{0.32761377} & 0.3723488\\
airplanes & \bnum{0.03085367} & 0.05462648 & 0.05813661 & 0.07094332 & 0.03879557 & \unum{0.03640466} & 0.6767205 & 0.5580559 & 0.54091376 & 0.69461095 & \unum{0.76444304}  & \bnum{0.77523446} & \unum{0.3336416} & 0.38723946 & 0.45810476 & 0.39395446 & \bnum{0.32884565} & 0.3718313\\
anchor & 0.06180606037378311 & 0.0930490791797638 & 0.07975093275308609 & 0.08189629018306732 & \bnum{0.04923838749527931} & \unum{0.051993612200021744} & 0.6516189575195312 & 0.5245109796524048 & 0.5949443578720093 & 0.6760560870170593 & \unum{0.746990442276001} & \bnum{0.7775198817253113} & \unum{0.3763543963432312} & 0.4067254662513733 & 0.46679872274398804 & 0.41052258014678955 & \bnum{0.35768038034439087} & 0.39489203691482544\\
ant & 0.04071618244051933 & 0.04377692565321922 & 0.05873226374387741 & 0.040262673050165176 & \bnum{0.028321359306573868} & \unum{0.03321472182869911} & 0.618435800075531 & 0.5139859914779663 & 0.4768108129501343 & 0.666693389415741 & \bnum{0.7150713205337524} & \unum{0.7094737887382507} & \unum{0.3731774091720581} & 0.3898458480834961 & 0.49021339416503906 & 0.39000439643859863 & \bnum{0.34894442558288574} & 0.40468746423721313  \\
barrel & 0.035187214612960815 & 0.04448249191045761 & 0.0660376250743866 & 0.03568468987941742 & \unum{0.03181754797697067} & \bnum{0.03156011924147606} & 0.5680216550827026 & 0.518051266670227 & 0.39395666122436523 & 0.6605573892593384 & \bnum{0.697859525680542} & \unum{0.6624187231063843} & \unum{0.3250042200088501} & 0.3429376184940338 & 0.47203993797302246 & 0.34223756194114685 & \bnum{0.32344505190849304} & 0.4011757969856262  \\
bass & 0.034290287643671036 & \unum{0.029283829033374786} & 0.05156983435153961 & 0.03958059847354889 & \bnum{0.026496965438127518} & 0.029461847618222237 & 0.5716130137443542 & 0.5657792091369629 & 0.458915114402771 & \unum{0.6736642718315125} & \bnum{0.6834713816642761} & 0.6608256101608276 & 0.3944709897041321 & \unum{0.3919573426246643} & 0.4906713366508484 & 0.41113099455833435 & \bnum{0.37223610281944275} & 0.4253334105014801  \\
beaver & 0.03194933384656906 & 0.04360319301486015 & 0.04845285043120384 & 0.03903388977050781 & \bnum{0.024341469630599022} & \unum{0.026879798620939255} & 0.5669040083885193 & 0.5078815221786499 & 0.4088093042373657 & \unum{0.6811832189559937} & \bnum{0.7153145670890808} & 0.6778385043144226 & \unum{0.37067848443984985} & 0.38868388533592224 & 0.511306643486023 & 0.3761388063430786 & \bnum{0.35463273525238037} & 0.4213109016418457  \\
binocular & \unum{0.036041803658008575} & 0.06584908813238144 & 0.08646529167890549 & 0.05535653978586197 & 0.04362916573882103 & \bnum{0.034986015409231186} & 0.6386640071868896 & 0.5534910559654236 & 0.464155375957489 & 0.6584338545799255 & \unum{0.7081717252731323} & \bnum{0.7144693732261658} & 0.3689989447593689 & 0.37382781505584717 & 0.4825684130191803 & \unum{0.36093786358833313} & \bnum{0.30620765686035156} & 0.36134475469589233  \\
bonsai & 0.06336567550897598 & 0.061586905270814896 & 0.06983886659145355 & 0.06674055755138397 & \bnum{0.0461239218711853} & \unum{0.046139683574438095} & 0.5383121371269226 & 0.4541221857070923 & 0.4377036690711975 & 0.6238788366317749 & \bnum{0.7190674543380737} & \unum{0.6500680446624756} & \unum{0.39261022210121155} & 0.43533360958099365 & 0.5156527757644653 & 0.4054640233516693 & \bnum{0.35212916135787964} & 0.4320231080055237  \\
brain & 0.06402547657489777 & 0.0609394796192646 & 0.09184809029102325 & 0.09609640389680862 & \unum{0.054687775671482086} & \bnum{0.04285589978098869} & 0.43784379959106445 & 0.3647860884666443 & 0.33246082067489624 & 0.5309394598007202 & \bnum{0.5928435325622559} & \unum{0.5803854465484619} & 0.3924979567527771 & 0.3781609833240509 & 0.4582332372665405 & 0.3453632593154907 & \bnum{0.2956019639968872} & \unum{0.3254023790359497}  \\
brontosaurus & 0.03773048520088196 & 0.04663918912410736 & 0.06298764050006866 & 0.038722556084394455 & \bnum{0.031252775341272354} & \unum{0.0362987145781517} & 0.5975801348686218 & 0.497841477394104 & 0.4960881471633911 & 0.6806848049163818 & \bnum{0.7084444761276245} & \unum{0.683430552482605} & \unum{0.37672847509384155} & 0.4102165102958679 & 0.4995083808898926 & 0.3977210819721222 & \bnum{0.36370664834976196} & 0.4457191824913025  \\
buddha & 0.06971053034067154 & 0.08084006607532501 & 0.07101466506719589 & 0.06255567818880081 & \bnum{0.03837340325117111} & \unum{0.03907007724046707} & 0.5797809362411499 & 0.4828444719314575 & 0.4699946641921997 & 0.6278610229492188 & \bnum{0.7324010729789734} & \unum{0.6934065818786621} & \unum{0.3941202759742737} & 0.41811293363571167 & 0.502776026725769 & 0.4023207426071167 & \bnum{0.3587852716445923} & 0.41354992985725403  \\
butterfly & 0.048133160918951035 & 0.04532857611775398 & 0.06552338600158691 & 0.0490778312087059 & \bnum{0.03291098400950432} & \unum{0.03835543617606163} & 0.6535226106643677 & 0.5622613430023193 & 0.5025684833526611 & 0.6886083483695984 & \bnum{0.7242658138275146} & \unum{0.7178820967674255} & 0.3510746359825134 & 0.37409326434135437 & 0.4429849088191986 & \unum{0.3411543369293213} & \bnum{0.3219161033630371} & 0.37175875902175903  \\
camera & 0.0492192842066288 & 0.0896405354142189 & 0.07502599060535431 & 0.06769154220819473 & \bnum{0.03609273210167885} & \unum{0.04630478471517563} & 0.6370570659637451 & 0.5204405784606934 & 0.4645770788192749 & 0.6664062738418579 & \bnum{0.7247664332389832} & \unum{0.7180036902427673} & 0.334725946187973 & 0.33952999114990234 & 0.4256764054298401 & 0.3380775451660156 & \bnum{0.28031671047210693} & \unum{0.31685084104537964}  \\
cannon & \bnum{0.029198868200182915} & 0.047215670347213745 & 0.06548941880464554 & 0.0383414551615715 & 0.034444648772478104 & \unum{0.02920343726873398} & 0.6214393377304077 & 0.5296037197113037 & 0.46532386541366577 & 0.6711844205856323 & \bnum{0.710956335067749} & \unum{0.7019774317741394} & \unum{0.32987675070762634} & 0.3590615391731262 & 0.47204071283340454 & 0.371154248714447 & \bnum{0.31689244508743286} & 0.3613506257534027  \\
car\_side & 0.03530892729759216 & 0.024061409756541252 & 0.04221191629767418 & 0.022595852613449097 & \unum{0.020805275067687035} & \bnum{0.017860492691397667} & 0.659225344657898 & 0.6039735078811646 & 0.5191807746887207 & 0.722502589225769 & \bnum{0.7665976285934448} & \unum{0.7477446794509888} & 0.37164241075515747 & \unum{0.37093091011047363} & 0.5058566331863403 & 0.4109450578689575 & \bnum{0.3624769449234009} & 0.40981125831604004  \\
ceiling\_fan & \bnum{0.0351455882191658} & 0.03855845704674721 & 0.056701067835092545 & 0.03818156570196152 & 0.046673208475112915 & \unum{0.03551475331187248} & 0.6809443235397339 & 0.525669515132904 & 0.6100777387619019 & 0.6965609788894653 & \unum{0.7342981100082397} & \bnum{0.7377433776855469} & 0.4074757993221283 & 0.43928804993629456 & 0.45088112354278564 & 0.42269790172576904 & \unum{0.38666588068008423} & \bnum{0.3840453624725342}  \\
cellphone & 0.056975413113832474 & 0.06604500114917755 & 0.09180566668510437 & 0.06556160002946854 & \bnum{0.0521044060587883} & \unum{0.056491997092962265} & 0.5370196104049683 & 0.47150909900665283 & 0.4561718702316284 & 0.6166521310806274 & \bnum{0.6533278226852417} & \unum{0.6495931148529053} & 0.3948538303375244 & 0.39908748865127563 & 0.4765971302986145 & 0.3817363977432251 & \bnum{0.33638110756874084} & \unum{0.3789931535720825}  \\
chair & 0.03649096190929413 & 0.07403799146413803 & 0.06662656366825104 & 0.04770655930042267 & \bnum{0.02987484075129032} & \unum{0.03287600725889206} & 0.672193169593811 & 0.5138244032859802 & 0.5618598461151123 & 0.7044646143913269 & \unum{0.7451483607292175} & \bnum{0.7574920058250427} & 0.3271883726119995 & 0.3836562931537628 & 0.444290429353714 & 0.36568498611450195 & \bnum{0.294522225856781} & \unum{0.323761522769928}  \\
chandelier & 0.052221354097127914 & 0.03732798993587494 & 0.06470926105976105 & 0.03973579406738281 & \unum{0.03245861455798149} & \bnum{0.024761999025940895} & 0.5621330738067627 & 0.4942856431007385 & 0.4425349235534668 & 0.6843807101249695 & \bnum{0.7359568476676941} & \unum{0.7140561938285828} & 0.3769758343696594 & 0.39550238847732544 & 0.48848986625671387 & 0.36394554376602173 & \bnum{0.32033002376556396} & \unum{0.36038637161254883}  \\
cougar & \unum{0.03302055597305298} & 0.033861078321933746 & 0.04942876473069191 & 0.03424729034304619 & 0.04323956370353699 & \bnum{0.02890939451754093} & 0.5558632612228394 & 0.5288063287734985 & 0.4378119111061096 & \unum{0.6512407660484314} & \bnum{0.6661611795425415} & 0.6298693418502808 & \bnum{0.3803146183490753} & 0.4072285294532776 & 0.5306618213653564 & \unum{0.40613046288490295} & 0.4128974676132202 & 0.4812472462654114  \\
crab & 0.04263223707675934 & 0.04563877731561661 & 0.06415968388319016 & 0.03918036073446274 & \unum{0.03480346128344536} & \bnum{0.029526609927415848} & 0.5588220357894897 & 0.5227704048156738 & 0.4225541353225708 & 0.6728717684745789 & \bnum{0.697189450263977} & \unum{0.6809861063957214} & \bnum{0.37158578634262085} & 0.39589163661003113 & 0.5234422087669373 & 0.3980686366558075 & \unum{0.382877916097641} & 0.41129958629608154  \\
crayfish & 0.04438503831624985 & 0.07312123477458954 & 0.06005940958857536 & 0.057191409170627594 & \bnum{0.03713589161634445} & \unum{0.04409676045179367} & 0.6232045888900757 & 0.5022428035736084 & 0.5400248765945435 & 0.6823863387107849 & \unum{0.7163416743278503} & \bnum{0.7225345373153687} & \unum{0.3781009316444397} & 0.4031640291213989 & 0.48862141370773315 & 0.41002073884010315 & \bnum{0.3683026432991028} & 0.4157298803329468  \\
crocodile & 0.04206670820713043 & 0.03397906571626663 & 0.05668416991829872 & 0.03534894809126854 & \unum{0.030536629259586334} & \bnum{0.027442414313554764} & 0.5484054088592529 & 0.517947793006897 & 0.43166106939315796 & \unum{0.6708899736404419} & \bnum{0.6784414649009705} & 0.650530219078064 & \bnum{0.37755775451660156} & 0.4306820034980774 & 0.5302994847297668 & 0.414243221282959 & \unum{0.4095671772956848} & 0.497281014919281  \\
cup & 0.04589707404375076 & 0.05439053475856781 & 0.05748802796006203 & 0.052698321640491486 & \bnum{0.028152059763669968} & \unum{0.034471187740564346} & 0.6617227792739868 & 0.5439561605453491 & 0.5403267741203308 & 0.6705593466758728 & \bnum{0.7501673102378845} & \unum{0.7337238192558289} & 0.38303226232528687 & 0.41916194558143616 & 0.4545365571975708 & 0.4041695296764374 & \bnum{0.36731114983558655} & \unum{0.3731401562690735}  \\
dalmatian & 0.030669737607240677 & 0.035781677812337875 & 0.04221689701080322 & 0.029890961945056915 & \unum{0.025436732918024063} & \bnum{0.020089609548449516} & 0.6913222670555115 & 0.6173164248466492 & 0.5426518321037292 & 0.7030907869338989 & \bnum{0.7685672640800476} & \unum{0.7424918413162231} & \unum{0.3330126106739044} & 0.3348076045513153 & 0.45679157972335815 & 0.3598206639289856 & \bnum{0.313446044921875} & 0.3476613163948059  \\
dollar\_bill & \unum{0.037114858627319336} & 0.09194234013557434 & 0.04675807058811188 & 0.04949498176574707 & 0.04218906909227371 & \bnum{0.03539150208234787} & 0.5723087191581726 & 0.49481356143951416 & 0.4504109025001526 & 0.6691067218780518 & \bnum{0.7022643089294434} & \unum{0.6941999197006226} & 0.2951992154121399 & 0.34472793340682983 & 0.46250230073928833 & \unum{0.290574312210083} & \bnum{0.2861063778400421} & 0.3599131107330322  \\
dolphin & 0.04248194769024849 & \bnum{0.025111418217420578} & 0.0481855683028698 & 0.0398048534989357 & 0.04549071937799454 & \unum{0.03545327112078667} & 0.5973868370056152 & 0.5358885526657104 & 0.5102714896202087 & \unum{0.6577934622764587} & \bnum{0.6625553965568542} & 0.6470476984977722 & \bnum{0.38089317083358765} & \unum{0.39275842905044556} & 0.47762030363082886 & 0.42794620990753174 & 0.4120599031448364 & 0.4437705874443054  \\
dragonfly & 0.08123403787612915 & 0.06560979038476944 & 0.058299001306295395 & 0.06814686954021454 & \bnum{0.04568422585725784} & \unum{0.045815736055374146} & 0.6293262243270874 & 0.5123814940452576 & 0.5671529173851013 & 0.6701046824455261 & \bnum{0.7056453227996826} & \unum{0.7007688879966736} & \bnum{0.37772560119628906} & 0.41745859384536743 & 0.4860469102859497 & 0.4144159257411957 & \unum{0.3824652433395386} & 0.42867547273635864  \\
\midrule
\end{tabular}
\end{adjustbox}
\fi
\end{table*}

\twocolumn
\cleardoublepage

\cleardoublepage
\bibliographystyle{IEEEtran}

\begin{IEEEbiography}[{\includegraphics[width=1in,height=1.25in,clip,keepaspectratio]{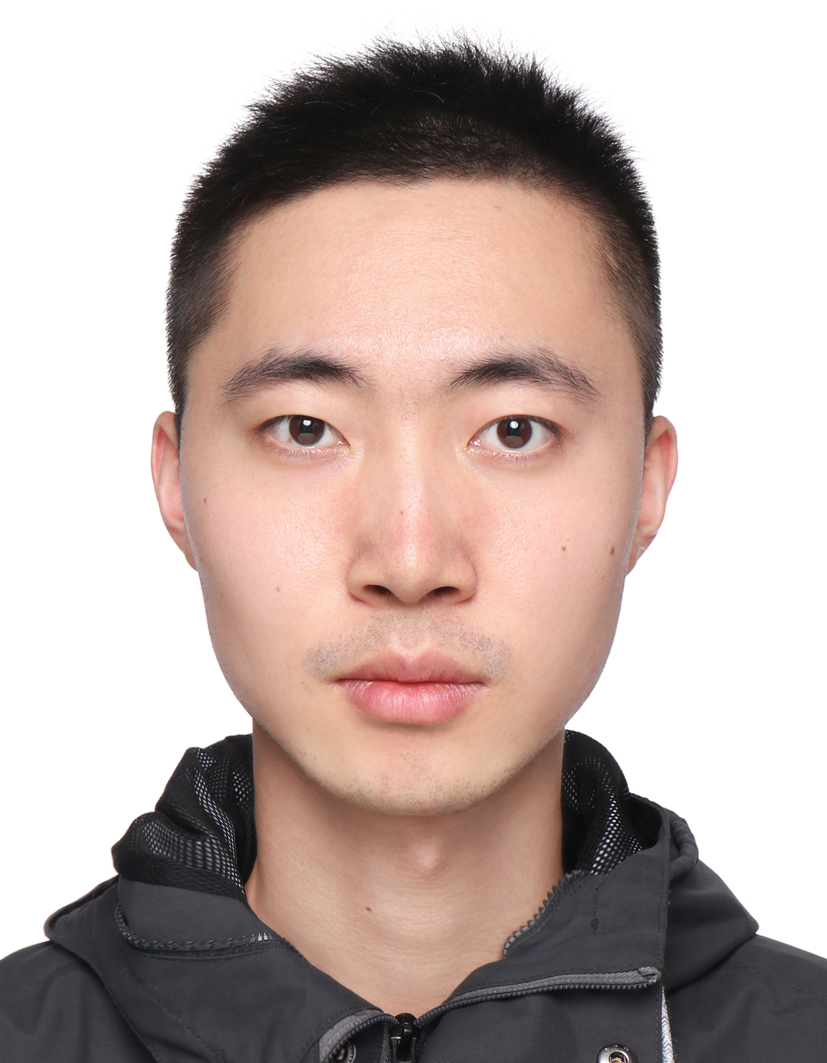}}]{Zelin Zhang} 
holds an Dual MS degree in Autonomous Systems from the Technische Universit\"at Berlin, Germany, and the KTH Royal Institute of Technology, Sweden (2022). 
Prior to that, Mr. Zhang earned a BS degree in Mechanical Design, Manufacturing and Automation from Harbin Institute of Technology, China (2018).
He has been a research assistant for Tencent Robotics X Lab (2019).
He is now a Computer Vision Engineer at the event camera company Prophesee in Paris, France.
\end{IEEEbiography}

\begin{IEEEbiography}[{\includegraphics[width=1in,height=1.25in,clip,keepaspectratio]{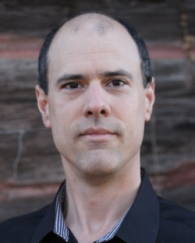}}]{Anthony J. Yezzi} 
holds the position of Julian Hightower Chair Professor within the School of Electrical and Computer Engineering at Georgia Institute of Technology where he directs the Laboratory for Computational Computer Vision. 
He has twenty five years of research experience in shape optimization via geometric partial differential equations. 
He obtained his Ph.D. in Electrical Engineering in December 1997 from the University of Minnesota with a minor in mathematics. 
After completing a postdoctoral research appointment at Massachusetts Institute of Technology, he joined the faculty at Georgia Tech in August 1999. 
Dr. Yezzi's research lies primarily within the fields of image processing and computer vision with a particular emphasis on medical imaging and 3D surface reconstruction.
His work spans a wide range of image processing and vision problems including image denoising, edge-detection, segmentation, shape analysis, multiview stereo reconstruction, visual tracking, and registration. 
Some central themes of his research include curve and surface evolution, differential geometry, partial differential equations, and shape optimization.
\end{IEEEbiography}

\begin{IEEEbiography}[{\includegraphics[width=1in,height=1.25in,clip,keepaspectratio]{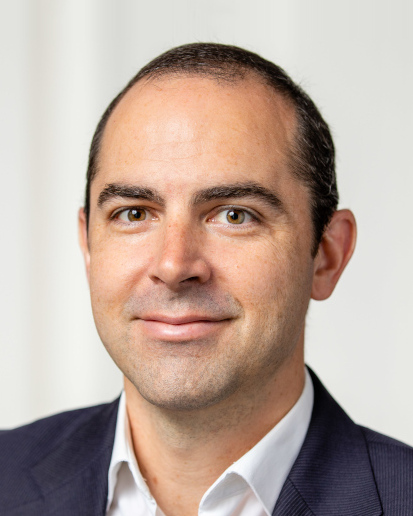}}]{Guillermo Gallego} (SM'19) 
is Associate Professor at Technische Universit\"at Berlin, in the Dept. of Electrical Engineering and Computer Science, and at the Einstein Center Digital Future, Berlin, where he leads the Robotic Interactive Perception Laboratory.
He is also a Principal Investigator at the Science of Intelligence Excellence Cluster, Berlin, Germany.
He received the PhD degree in Electrical and Computer Engineering from the Georgia Institute of Technology, USA, in 2011, supported by a Fulbright Scholarship.
From 2011 to 2014 he was a Marie Curie researcher with Universidad Politecnica de Madrid, Spain, and from 2014 to 2019 he was a postdoctoral researcher at the Robotics and Perception Group, University of Zurich and ETH Zurich, Switzerland.
Since 2022 he is also Co-Director of the HEIBRiDS interdisciplinary research school, Berlin.
His research interests include robotics, computer vision, signal processing, optimization and geometry. 
\end{IEEEbiography}

\typeout{get arXiv to do 4 passes: Label(s) may have changed. Rerun} %

\else 
\bibliographystyle{IEEEtran}
\bibliography{all,paper_ref}

\cleardoublepage

\fi

\end{document}